\documentclass[10pt,journal,compsoc]{IEEEtran}

\ifCLASSOPTIONcompsoc

  \usepackage[nocompress]{cite}
\else

  \usepackage{cite}
\fi

\ifCLASSINFOpdf
\else
\fi

\usepackage{tabu}
\usepackage{graphicx}
\usepackage{amsfonts}
\usepackage{amssymb}
\usepackage{amsmath}
\usepackage{subcaption}
\usepackage{threeparttable}
\usepackage{makecell}

\begin{document}

\title{Characterizing Types of Convolution in Deep Convolutional Recurrent Neural Networks for Robust Speech Emotion Recognition}

\author{Che-Wei~Huang,~\IEEEmembership{Student~Member,~IEEE,}
        and~Shrikanth~S~Narayanan,~\IEEEmembership{Fellow,~IEEE}
\IEEEcompsocitemizethanks{\IEEEcompsocthanksitem The authors are with the Signal Analysis and Interpretation Laboratory (SAIL), Department of Electrical Engineering and Signal and Image Processing Institute, University of Southern California, Los Angeles, CA 90089 USA\protect\\
E-mail: cheweihu@usc.edu; shri@sipi.usc.edu
\IEEEcompsocthanksitem This work is supported by NSF.
}%
}

\markboth{Submitted to The IEEE Transactions}
{Che-Wei Huang \MakeLowercase{\textit{et al.}}: Submission to the IEEE Transaction}

\IEEEtitleabstractindextext{%
\begin{abstract}
Deep convolutional neural networks are being actively investigated in a wide range of speech and audio processing applications including speech recognition, audio event detection and computational paralinguistics, owing to their ability to reduce factors of variations, for learning from speech. However, studies have suggested to favor a certain type of convolutional operations when building a deep convolutional neural network for speech applications although there has been promising results using different types of convolutional operations. In this work, we study four types of convolutional operations on different input features for speech emotion recognition under noisy and clean conditions in order to derive a comprehensive understanding. 
Since affective behavioral information has been shown to reflect temporally varying of mental state and convolutional operation are applied locally in time, all deep neural networks share a deep recurrent sub-network architecture for further temporal modeling.
We present detailed quantitative module-wise performance analysis to gain insights into information flows within the proposed architectures. In particular, we demonstrate the interplay of affective information and the other irrelevant information during the progression from one module to another. Finally we show that all of our deep neural networks provide state-of-the-art performance on the eNTERFACE'05 corpus. 
\end{abstract}

}

\maketitle

\IEEEdisplaynontitleabstractindextext

\IEEEpeerreviewmaketitle

\IEEEraisesectionheading{\section{Introduction}\label{sec:introduction}}

\IEEEPARstart{E}motion plays a fundamental role in our daily lives for effective communication, which underlies the abilities of humans to interact, collaborate, empathize and even compete with others. Researchers have been working on understanding human emotion, or in general human behaviors \cite{BSP13}, for years from both psychological and computational perspectives for several reasons including because it serves as a lens to observe the dynamics of one's internal mental state. Moreover, with the advent of artificially intelligent agents, it is hardly an overstatement to stress the importance of emotion recognition in supporting natural and engaging human-machine interaction. 

Human behavioral cues often mix and manifest multiple sources of information together. To robustly recover affective information from multiplexed behavioral cues renders emotion recognition a challenging task. For example, speech contains not only linguistic content of what is said but also attributes of the speaker such as identity, gender, age, speaking style, and language background as well as information about the environment and context. All of these factors are entangled and transmitted through a single channel during speech articulation. Speech emotion recognition, therefore, involves the inverse process of disentangling these signals and identifying affective information. 

A multitude of studies on the subject of emotion recognition have discovered a number of emotion-related parameters based on prior knowledge of psychology, speech science, vision science and through signal processing and machine learning approaches. The commonly used features include pitch, log-Mel filterbank energies (log-Mels) \cite{melscale37}, Mel-frequency cepstral coefficients (MFCCs) \cite{Bridle74} and perceptual linear prediction in the acoustic modality, and Haar, local binary pattern, histogram of oriented gradients and scale-invariant feature transform in visual modality. A variety of classifiers based on these features have been reported to perform well. In particular, an extensive feature set consisting of thousands of hand-engineered parameters has been recommended in the past few INTERSPEECH challenges \cite{CompareE13}, and in a recent meta research review article \cite{Eyben15}.

In addition to hand-crafted feature engineering, deep learning \cite{Rumelhart:1986:LIR, LeCun:1990, Hochreiter:1997:LSM:1246443.1246450} provides an alternative approach to formulate appropriate features for the task at hand. In the last few years, convolutional neural networks (CNNs) \cite{Alexnet12, ZeilerF14, VGG15, Googlenet15, Resnet16} have demonstrated outstanding performances in various applications including image recognition, object detection, and recently speech acoustic modeling. 
Compared to hand-crafted features, a CNN that learns from a large number of training samples via a deep architecture can capture a higher-level representation for the task-specific knowledge distilled from annotated data. In the area of speech emotion recognition, several researchers have investigated the effectiveness of CNNs into automatically learning affective information from signal data \cite{MaoDHZ14, Anand, Zhang:2016:MDC:2911996.2912051, MildeB15}.

Information encoded in speech signals is inherently sequential. Moreover, psychological studies have shown that affective information involves a slow temporal evolution of mental states \cite{Oatley96}. Based on this observation, previous studies have also investigated the use of architectures that explicitly model the temporal dynamics, such as hidden Markov models (HMM) \cite{Schuller03} or recurrent neural networks (RNN) \cite{Wollmer10, Metallinou2012, Lee15} for recognizing human emotion in speech.

Furthermore, there is a growing trend in combining CNN and RNN into one architecture and to train the entire model in an end-to-end fashion. The motivation behind a holistic training is derived from the need to avoid greedily enforcing the distribution of intermediate layers to approximate that of labels, which is believed to maximally exploit the advantage of deep learning over traditional learning methods and would lead to an improved performance. 
For example, Sainath et al. proposed an architecture, called the Convolutional Long Short-Term Memory Deep Neural Networks (CLDNN) model, made up of a few convolutional layers, long short-term memory gated (LSTM) RNN layers and fully connected (FC) layers in the respective order. They trained CLDNNs on the log-Mel filterbank energies \cite{SainathVSS15} and on the raw waveform speech signal \cite{SainathWSWV15} for speech recognition, and showed that both CLDNN models outperform CNN and LSTM alone or combined. Likewise, Huang et al. \cite{Huang17} and Lim et al. \cite{LimJL16} reported CLDNN-based speech emotion recognition experiments, on log-Mels and spectrograms respectively, using similar benchmark settings to highlight the superior performance resulting from an end-to-end training.  

In a recent work, Sainath et al. \cite{SainathLi16} observed that under a moderately noisy condition, the spectrally only convolutional operation degrades the performance. They hypothesized noise has made it difficult for local filters to learn translation invariance and thus the local decisions are prone to error. Our work is built upon this observation in order to quantitatively investigate whether different types of convolutional operations could show robustness to noise for speech emotion recognition.
 
In this work, we extend our previous work in \cite{Huang17} to characterize four types of convolutional operations in a CLDNN for speech emotion recognition. We use log-Mels and MFCCs as input to the proposed models depending on their spectral-temporal correlation. In particular, we compare spectral decorrelation power between one type of the convolutional operations and the discrete cosine transformation (DCT), under both clean and noisy conditions. In addition, we quantitatively and visually analyze modules in the proposed CLDNN-based models in order to gain insights into the information flows within the models. 

Our contributions are multi-fold. First of all, we consider all commonly used convolutional operations for offering a comprehensive understanding, including two types covered in \cite{Anand}. Second, unlike previous studies \cite{Anand, Keren16a} that increased training corpus size \textit{internally}, we perform data augmentation with a noise corpus. As a result, we evaluate the proposed models under both clean and noisy conditions to quantitatively measure the influence of noise on different types of convolutional operations. To the best of our knowledge, this is the first work to study noise influence on types of convolutional operations. Furthermore, we carry out module-wise evaluation and visualization to analyze the information flows of different factors encoded in speech and their interplay along the depth of an architecture. 

The outline of this paper is as follows. Section \ref{sec:related_work} reviews previous related work. Section \ref{sec:cldnn} presents the architecture of the proposed models and Section \ref{sec:baseline} describes three competitive baseline models. Section \ref{sec:data} introduces the corpus and data augmentation procedure. Section \ref{sec:exps} details the experimental settings and the results are interpreted in Section \ref{sec:exp_result}. Section \ref{sec:conclu} concludes this paper. 

\section{Related Work}
\label{sec:related_work} 
Before the present era of deep learning, speech emotion recognition systems prevalently relied on a two-stage training approach, where feature engineering and classifier training were performed separately. Commonly used hand-crafted features include pitch, MFCC, log-Mels and the recommended feature sets from the INTERSPEECH challenges. Support vector machine (SVM) and extreme learning machine (ELM) were two of the most competitive classifiers. For the ease to compare models, Eyben et al. \cite{Eyben15} summarized the performances by a SVM trained on the INTERSPEECH challenge feature sets over several public corpora. Yan et al. \cite{YanZXLLW16} recently proposed a sparse kernel reduced-rank regression (SKRRR) for bimodal emotion recognition from facial expressions and speech, which has achieved one of the state-of-the-art performances on the eNTERFACE'05 \cite{enterface05} corpus.

Han et al. \cite{Han14} employed a multilayer perceptron (MLP) to learn from spliced data frames and took statistics of aggregated frame posteriors as utterance-level features. An MLP-ELM supervised by these utterance features and the corresponding labels has been shown to outperform the MLP-SVM model. 

It has been known that emotion involves temporal variations of mental state. To exploit this fact, W{\"{o}}llmer et al. \cite{Wollmer10} and Metallinou et al. \cite{Metallinou2012} conducted experiments at the conversation-level to show that human emotion depends on the context of a long-term temporal relationship using HMM and Bi-directional LSTM (BLSTM). 
Lee et al. \cite{Lee15} posed speech emotion recognition at the utterance level as a sequence learning problem and trained an LSTM with a connectionist temporal classification objective to align voiced frames with emotion activation. 

Deep CNN models were initially applied to computer vision related tasks and have achieved many ground-breaking results \cite{Alexnet12, ZeilerF14, VGG15, Googlenet15, Resnet16}. Recently, researchers have started to consider their use in the acoustic domain, including speech recognition \cite{Abdel-Hamid:2014:CNN:2687092.2687099, Li13, SainathVSS15, SainathWSWV15, DChanL15}, audio event detection \cite{Hershey16, Takahashi17} and speech emotion recognition \cite{MaoDHZ14, Anand, Zhang17}. Abdel-Hamid et al. \cite{Abdel-Hamid:2014:CNN:2687092.2687099} concluded that one of the advantages in using CNNs to learn from less processed features such as raw waveforms, spectrograms and log-Mels is their ability to reduce spectral variation, including speaker and environmental variabilities; this capability is attributed to structures such as local connectivity, weight sharing, and pooling. 
When training a CNN model for speech emotion recognition, Mao et al. \cite{MaoDHZ14} proposed to learn the filters in a CNN on spectrally whitened spectrograms. The learning, however, is carried out by a sparse auto-encoder in an unsupervised fashion.
Anand et al. \cite{Anand} benchmarked two types of convolutional operations in their CNN-based speech emotion recognition systems: the spectral-temporally convolutional operation and the full-spectrum temporally convolutional operation (see Fig. \ref{fig:overview_more} for details). Their results showed the full-spectrum temporal convolution is more favorable for speech emotion recognition. They also reported the performance of an LSTM trained on the raw spectrograms.

Recently, Sainath et al. proposed the CLDNN architecture for speech recognition based on the log-Mels \cite{SainathVSS15} and the raw waveform signal \cite{SainathWSWV15}, in which both models have been shown to more competitive than a LSTM and a CNN model alone or combined. They also demonstrated that with a sufficient amount of training data (roughly $2,000$ hours), a CLDNN trained on the raw waveform signal can match the one trained on the log-Mels. Moreover, they found the raw waveform and the log-Mels in fact provide complementary information. Based on the CLDNN architecture, Trigeorgis et al. \cite{trigeorgis_icassp16} published a model using the raw waveform signal for continuous emotion tracking. Huang et al. \cite{Huang17} trained a CLDNN model on the log-Mels for speech emotion recognition and quantitatively analyzed the difference in spectrally decorrelating power between the discrete cosine transformation and the convolutional operation. Lim et al. \cite{LimJL16} repeated the comparison between CNN, LSTM and CLDNN for speech emotion recognition using spectrograms. Ma et al. \cite{Ma:2016:DED:2988257.2988267} applied the CLDNN architecture to classifying depression based on the log-Mels and spectrograms. They employed the full-spectrum temporally convolutional operation on the log-Mels but the temporally-only convolutional operation on the spectrograms.

On the multi-modal side, Zhang et al. \cite{Zhang:2016:MDC:2911996.2912051} fine-tuned the AlexNet on spectrograms and images, separately, for audio-visual emotion recognition but only applied time-averaging for temporal pooling. Tzirakis et al. \cite{Tzirakis17} extended the uni-modal work in \cite{trigeorgis_icassp16} to make use of visual cues. They fine-tuned the pre-trained ResNet model \cite{Resnet16} for facial expression recognition and then re-trained the concatenated bimodal network with the LSTM layers re-initialized again. 
 
Our work is similar to Anand et al. \cite{Anand} because we both report the benchmarking of convolutional types. However, in addition to the novelty aforementioned, we train our models in an end-to-end fashion on log-Mels and MFCCs depending on their locally spectral-temporal correlation. Moreover, we keep the testing partition speaker-independent of the training parition. Ma et al. \cite{Ma:2016:DED:2988257.2988267} also experimented with two types of convolutional operations but they applied them to different features. As a result, it is difficult to draw a fair conclusion from the comparison. This work is also similar to Trigeorgis et al. \cite{trigeorgis_icassp16}, Lim et al. \cite{LimJL16} and Huang et al. \cite{Huang17}, where all adopt the CLDNN architecture for speech emotion recognition/tracking but the underlying features and the intended goals are different.

\begin{figure*}[th]
  \centering
  \includegraphics[width=\textwidth,height=13cm]{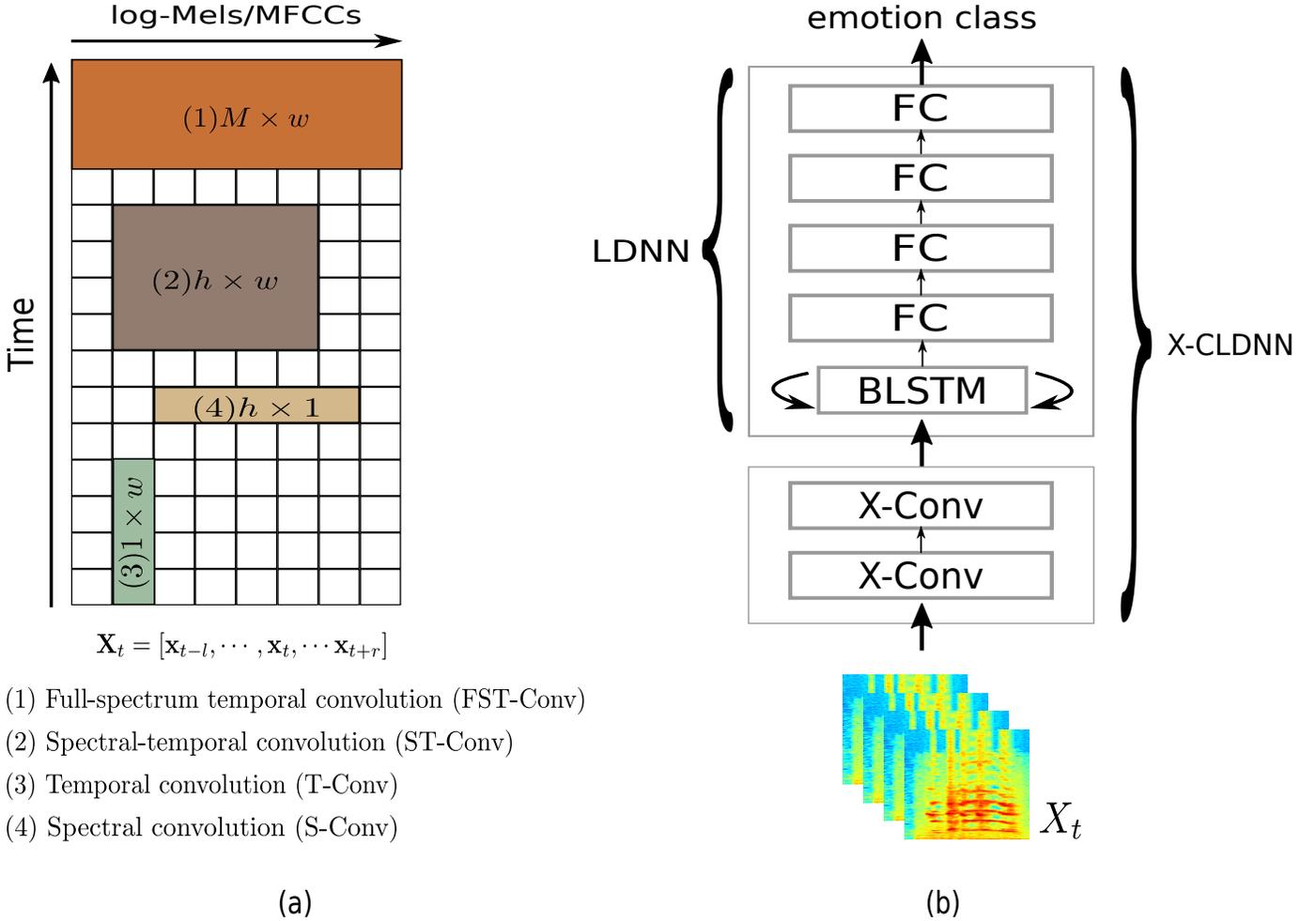}
  \caption{An overview of the proposed neural networks for speech emotion recognition. \textit{(a)} Four types of the convolutional operation over a given two-dimensional input $\mathbf{X}_t=[\mathbf{x}_{t-l},\cdots,\mathbf{x}_t,\cdots,\mathbf{x}_{t+r}]$ are defined, including the full-spectrum temporal convolution (FST-Conv), the spectral-temporal convolution (ST-Conv), the temporal only convolution (T-Conv), the spectral only convolution (S-Conv). The shape (height $h$ and width $w$ together) of a filter determines the type of a convolutional operation. Filters of shape $M\times w$ (FST-Conv) consider all $(M)$ frequency bands for $w$ frames per scan. Filters of shape $h\times w$ (ST-Conv) only process local spectral-temporal information. Filters of shape $1\times w$ (T-Conv) and of shape $h\times 1$ (S-conv) only observe local information along their designated direction, respectively. \textit{(b)} An LDNN model, consisting of a bi-directional long short-term memory (BLSTM) gated recurrent neural layer followed by four fully connected feed-forward neural layers (FC), serves to be the common sub-network architecture for each of the proposed models. Two X-Conv layers together with the LDNN sub-network forms a X-CLDNN model.
}
\label{fig:overview}
\end{figure*}

\section{Deep Convolutional Recurrent Models}
\label{sec:cldnn}
In this section we describe the proposed deep convolutional recurrent networks and details of structurally different convolutional operations on the log-Mels and the MFCCs. Fig. \ref{fig:overview} illustrates the overview of the models we design for speech emotion recognition. In Fig. \ref{fig:overview} (a), we define four types of convolutional operations depending on the shape of their feature maps. By dividing the convolutional operations into four types, we expect to understand their differences for a finer analysis after they have been optimized to learn from the spectral-temporal signals. In Fig. \ref{fig:overview} (b), we depict a deep recurrent neural network, called the LDNN model, as the common sub-network architecture for every model. Two convolutional layers together with the LDNN sub-network make up a CLDNN architecture. As a convolutional layer is applied locally in time, the LDNN model is supposed to model the long-term temporal relationship within an utterance. We only consider spectral-temporal features as input to a CLDNN model. Specifically, an emotional utterance $\mathbf{u}$ is represented by a sequence of spectral features $\left\{\mathbf{x}_t^{\mathbf{u}}\right\}$. These spectral features can either be the log-Mels or the MFCCs depending on the application scenario. All models are presented for a comprehensive study to understand the role a convolutional layer plays in learning the affective information in speech. Overall, we present eight models based on the combinations of the factors including the input features (the log-Mels or the MFCCs) and the type of convolutional operations (spectral only, temporal only,  spectral-temporal or full-spectrum temporal). In the following subsections, we give a brief review of the convolutional and recurrent neural layers and introduce corresponding notations.
 
\subsection{Types of Convolutional Operations}
\label{subsec:typesconv}

A convolutional neural layer $\textrm{Conv}$ that receives an input tensor $\mathbf{X}\in\mathbb{R}^{C\times H^0 \times W^0}$ consists of a convolutional function $\textbf{F}_\mathbf{\kappa}:\mathbb{R}^{C\times H^0 \times W^0}\mapsto\mathbb{R}^{K\times H^1 \times W^1}$, an activation function $\sigma$ and an optional pooling function $\mathbf{F}_\mathbf{\pi}:\mathbb{R}^{K\times H^1 \times W^1}\mapsto\mathbb{R}^{K\times H^2 \times W^2}$. 

The convolutional function $\textbf{F}_\mathbf{\kappa}$ is defined by $K$ feature maps $(\mathbf{h}_k, \mathbf{b}_k)\in\mathbb{R}^{C\times h\times w}\times \mathbb{R}^{H^1 \times W^1}$ of shape $h\times w$, where the $kij$th component of $\mathbf{F}_{\mathbf{\kappa}}(\mathbf{X})$ is given as
\begin{eqnarray}
\label{eq:conv}
\mathbf{F}_\mathbf{\kappa}(\mathbf{X})_{kij} 
&\triangleq&
\mathbf{h}_k
*
\mathbf{X}_{ij}^{hw}
+
\mathbf{b}_{kij}
\nonumber\\
&=&
\sum_{c=0}^{C-1}
\sum_{\mu=0}^{h-1}
\sum_{\nu=0}^{w-1}
\mathbf{X}_{ij}[c,\mu,\nu]
\mathbf{h}_k[c,\mu,\nu]
+
\mathbf{b}_{kij},\quad
\end{eqnarray}
in which $\mathbf{X}_{ij}[c,\mu,\nu] = \mathbf{X}[c, i\cdot s_\kappa+\mu,j \cdot t_\kappa+\nu]$ and $s_\kappa$, $t_\kappa$ are the strides, i.e. the amount of shift, of the filters in the convolutional operation in their respective directions.

Likewise, it is straight-forward to formulate the pooling function $\mathbf{F}_\mathbf{\pi}$ acting on an input $\mathbf{Y}\in\mathbb{R}^{K\times H^1 \times W^1}$ through a filter of shape $m\times n$ by the component-wise definition:
\begin{eqnarray}
\label{eq:pool}
\mathbf{F}_{\mathbf{\pi}}
(\mathbf{Y})_{kij}
&\triangleq&
\pi
\left(
\mathbf{Y}_{kij}^{mn}
\right),
\end{eqnarray}
where $\mathbf{Y}_{kij}^{mn}\in\mathbb{R}^{m\times n}$ is a sub-tensor of $\mathbf{Y}$ lying on the $k$th slice of $\mathbf{Y}$ with its first entry aligned to $\mathbf{Y}[k,i\cdot s_\pi,j\cdot t_\pi]$, and $\pi$ is the pooling operation, usually the max or the mean functions. Similarly, $s_\pi$ and $t_\pi$ are the strides of the filters in the pooling operations in their respective directions.

Typical choices of the activation functions include the sigmoid function $\sigma(x)=\frac{1}{1+\exp(-x)}$, the hyperbolic tangent function $\sigma(x)=\tanh(x)$ and the rectified linear unit (ReLU) $\sigma(x)=\textrm{max}(0,x)$. 

 Concisely, a convolutional neural layer can be summarized as a function composition 
\begin{equation}
\textrm{Conv} \triangleq \mathbf{F}_\mathbf{\pi}\circ (\sigma \odot \mathbf{F}_\mathbf{\kappa}),
\end{equation} 
where $\circ$ and $\odot$ denotes function composition and element-wise application.

In this work, we concentrate entirely on the convolution function $\mathbf{F}_\mathbf{\kappa}$ and adjust the pooling function $\mathbf{F}_\mathbf{\pi}$ accordingly. In particular, we are interested in the relationship between the acoustic emotional pattern learnt by the model and the shape of the filter $\mathbf{h}_k$ in feature maps. To this end, we divide the shapes of the filters $\mathbf{h}_k$ into four categories to highlight their structural differences: the full-spectrum temporally (FST-Conv), the spectral-temporally (ST-Conv), the temporally only (T-Conv) and the spectrally only (S-Conv) convolutional operations. In what follows, we mathematically define each category. 

\paragraph{FST-Conv}
First of all, we consider filters of shape $M\times w$ for $w\geq2$, where $M$ denotes the number of spectral bands and $w$ specifies the width on the temporal axis. Since this type of filters covers the entire spectrum, they convolve with the input tensor only in the temporal direction and as a result the pooling function can only perform temporal pooling. This type convolves with global spectral information and models across neighboring frames. 

\paragraph{ST-Conv}
A ST-Conv layer contains filters of shape $h\times w$, where $2\leq h \leq M-1$ and $w\geq2$. This type of filters observes local spectral-temporal information at a time and is free to convolve with the input tensor in both directions. Accordingly, the pooling function also operates on the convolved tensor through a two-dimensional filter.

\paragraph{T-Conv}
A T-Conv layer is similar to a FST-Conv layer except that the filters in a T-Conv layer has a shape of $1\times w$ for $w\geq2$. These filters convolve with the input tensor along the temporal direction from one frequency band to another and ignore spectrally contextual information. The pooling function acts on the convolved tensor along the temporal direction correspondingly.  

\paragraph{S-Conv}
A spectrally only convolutional neural layer consists of filters of shape $h\times 1$, where $h\geq1$ and the pooling function down-samples the convolved tensor along the spectral direction. Note that the S-Conv type is closely related to the traditional signal processing techniques; for example, DCT transformation from log-Mels to MFCCs belongs to this category when $h=M$ except that the filters in DCT are mathematically pre-defined; see Sec. \ref{subsec:baseline_ldnn-mfccs} for more details.

For each type of the convolutional operations, we employ a stride of $1$.  Since our focus is on the convolutional operations, we employ a fixed pooling size of $3$ and a fixed stride of $2$ in their respective direction(s) of convolution. Table \ref{tab:conv_archs} summarizes the parameters for all Conv layers.

\begin{table}[ht]
\caption{A summary of the parameters for each model architecture. $M$ denotes the spectral dimensionality and var stands for variable parameters for tuning. The dash symbol indicates the situation where the parameter tuning is not applicable.}
\centering
\begin{tabular}{c c c c c c c c c c c } 
\Xhline{4\arrayrulewidth}
 & $h$ & $w$ & $m$ & $n$ & $s_\kappa$ & $t_\kappa$ & $s_\pi$ & $t_\pi$\\
\Xhline{2\arrayrulewidth}
\multicolumn{1}{ c }{\textrm{S-Conv}} & var & $1$ & $3$ & $1$ & $1$ & $1$ & $2$ & $1$\\ 
\Xhline{2\arrayrulewidth}
\multicolumn{1}{ c }{\textrm{T-Conv}} & $1$ & var & $1$ & $3$ & $1$ &  $1$ & $1$ & $2$\\ 
\Xhline{2\arrayrulewidth}
\multicolumn{1}{ c }{\textrm{ST-Conv}} & var & var & $3$ & $3$ & $1$ & $1$ &  $2$ & $2$\\
\Xhline{2\arrayrulewidth}
\multicolumn{1}{ c }{\textrm{FST-Conv}} & $M$ & var & $1$ & $3$ & -- & $1$ & -- & $2$\\
\Xhline{4\arrayrulewidth}
\end{tabular}
\label{tab:conv_archs}
\end{table}

\subsection{Deep Recurrent Neural Network}
\label{subsec:drnn}
Suppose the input is a sequence of vectors $\{\mathbf{x}_t\}$. The Elman type simple recurrent neural network RNN \cite{Elman90findingstructure} is defined through the following equations:
\begin{eqnarray}
\mathbf{h}_t 
&=& 
\sigma_\textbf{h}
\left(
\mathbf{U}^{\textbf{hx}}\mathbf{x}_t
+
\mathbf{U}^{\textbf{hh}}\mathbf{h}_{t-1}
+
\mathbf{u}_\textbf{h}
\right)\label{eq:rnn_h}
\\
\mathbf{y}_t 
&=& 
\sigma_\textbf{y}
\left(
\mathbf{U}^{\textbf{yh}}\mathbf{h}_t
+
\mathbf{u}_\textbf{y}
\right),
\end{eqnarray}
where $\mathbf{h}_t$ as an non-linear recurrent transformation of all past history $\{\mathbf{x}_s\}_{s=1}^t$ represents the system memory at time $t$, $(\mathbf{U}^{\textbf{ba}}, \mathbf{u}_b)$ is an affine mapping from a space of type $\textbf{a}$ to one of type $\textbf{b}$, and $\sigma_\textbf{c}$ is the activation function for type $\mathbf{c}$. Here $\mathbf{x}$, $\mathbf{h}$ and $\mathbf{y}$ denote the input, hidden and output vectors, respectively. However, training a simple RNN with the back-propagation algorithm may cause the issues of gradient vanishing or explosion. Although heuristic techniques such as gradient clipping can alleviate the issue of gradient explosion, the gradient vanishing problem is mitigated by an enhanced architecture: the LSTM architecture \cite{Hochreiter:1997:LSM:1246443.1246450}.

An LSTM is able to decide when to read from the input, to forget the memory or to write an output by controlling a gating mechanism. By definition an LSTM learns the following internal controlling functions:
\begin{eqnarray}
\mathbf{i}_t
&=&
\sigma_\mathbf{i}
\left(
\mathbf{U}^{\mathbf{ix}}\mathbf{x}_t
+
\mathbf{U}^{\mathbf{is}}\mathbf{s}_{t-1}
+
\mathbf{u}_\mathbf{i}
\right)\\
\mathbf{f}_t
&=&
\sigma_\mathbf{f}
\left(
\mathbf{U}^{\mathbf{fx}}\mathbf{x}_t
+
\mathbf{U}^{\mathbf{fs}}\mathbf{s}_{t-1}
+
\mathbf{u}_\mathbf{f}
\right)\\
\mathbf{o}_t
&=&
\sigma_\mathbf{o}
\left(
\mathbf{U}^{\mathbf{ox}}\mathbf{x}_t
+
\mathbf{U}^{\mathbf{os}}\mathbf{s}_{t-1}
+
\mathbf{u}_\mathbf{o}
\right)\\
\mathbf{g}_t
&=&
\tanh
\left(
\mathbf{U}^{\mathbf{gx}}\mathbf{x}_t
+
\mathbf{U}^{\mathbf{gs}}\mathbf{s}_{t-1}
+
\mathbf{u}_\mathbf{g}
\right)\\
\mathbf{c}_t
&=&
\mathbf{c}_{t-1}\odot\mathbf{f}_t+\mathbf{g}_t\odot\mathbf{i}_t
\label{eq:lstm_c}
\\
\mathbf{s}_t
&=&
\tanh
\left(
\mathbf{c}_t
\right)
\odot
\mathbf{o}_t
\end{eqnarray}
where $\mathbf{i}$, $\mathbf{f}$, $\mathbf{o}$, $\mathbf{g}$, $\mathbf{c}$ and $\mathbf{s}$ represent the input, forget, output, gate, cell and output vectors, respectively. In particular, the change from non-linear multiplicative recurrence in Eq. (\ref{eq:rnn_h}) to linear additive recurrence in Eq. (\ref{eq:lstm_c}) theoretically prevents gradients from vanishing during back-propagating the error through time.
Moreover, studies have found that a BLSTM layer can further improve upon a unidirectional LSTM in applications such as speech recognition \cite{Graves:2005:BLN:1986079.1986220}, translation \cite{Sundermeyer14translationmodeling} and emotion recognition \cite{Wollmer10,Metallinou2012}  
as it fuses information from the past and the future.

Suppose an $\textrm{LSTM}:\mathbb{R}^{D_1\times T}\mapsto\mathbb{R}^{D_2\times T}$ takes in a sequence $\{\mathbf{x}_t\}_{t=1}^T$ and returns $\{\mathbf{y}_t^f\}_{t=1}^T$, and another $\textrm{LSTM}:\mathbb{R}^{D_1\times T}\mapsto\mathbb{R}^{D_2\times T}$ takes in a reversed sequence $\{\mathbf{x}_{T+1-t}\}_{t=1}^T$ and returns $\{\mathbf{y}_t^b\}_{t=1}^T$. A $\textrm{BLSTM}:\mathbb{R}^{D_1\times T} \times \mathbb{R}^{D_1\times T} \mapsto\mathbb{R}^{(2*D_2)\times T}$, which is made of two LSTMs, runs on two sequences $\{\mathbf{x}_t\}_{t=1}^T$ and $\{\mathbf{x}_{T+1-t}\}_{t=1}^T$ and gives another sequence $\{\mathbf{z}_t\}_{t=1}^T$, where $\mathbf{z}_t=[\mathbf{y}_t^f;\mathbf{y}_t^b]$ is the concatenation of $\mathbf{y}_t^f$ and $\mathbf{y}_t^b$.

\subsection{CLDNN-based Models}
\label{subsec:cldnn}
Before defining a variety of CLDNN-based models, we introduce a shared sub-network architecture among them. The sub-network contains one BLSTM layer followed by four fully connected feed-forward layers. Each direction of the BLSTM layer has $128$ cells so the BLSTM outputs a sequence of vectors in $\mathbb{R}^{256}$. We take a mean pooling over the output of the BLSTM layer to obtain the utterance representation $\mathbf{c}$ rather than using the output vector at the last time step. A dropout mechanism \cite{Srivastava:2014} of probability $0.2$ is fixed and applied to the representation $\mathbf{c}$ to regularize the learning process. These four FC layers have their own size of $128,32,32,N$, respectively, where $N$ denotes the number of emotion classes, in which the first three FC layers are activated by the ReLU and the last one by the softmax function for classification. This architecture based on (B)LSTM and FC layers is conveniently called an LDNN model \cite{SainathVSS15}. Note that we employ a BLSTM layer instead of an LSTM layer as in \cite{SainathVSS15} because it has been shown that the ability of a BLSTM to integrate future information into representation learning is beneficial to emotion recognition. 

In the bottom of the LDNN sub-network, there are two Conv layers. Each Conv layer has $32$ feature maps and each of them is activated by the ReLU. Formally, we define X-CLDNN model to be an LDNN sub-network architecture specified above on top of two X-Conv layers, where $\textrm{X}\in\{\textrm{S}, \textrm{T}, \textrm{ST}, \textrm{FST}\}$.

A Conv layer is often said to be local because its feature maps when being computed at a local region on the input tensor depend only on the entries that the feature maps currently overlap with. As a result, we expect the input tensor to preserve locality in both spectral and temporal directions in general. However, due to the aforementioned structural differences, it is reasonable to relax this expectation a little bit accordingly. For example, a ST-Conv certainly requires its input tensor to maintain spectral-temporal correlation locally while a (FS)T-Conv and a S-Conv only need such locality preservation in the temporal or spectral direction, respectively. Taking this issue into consideration, in this work, we apply all four types of the Conv layer to the log-Mels and denote the corresponding CLDNN-based models as X-CLDNN (log-Mels) for $\textrm{X}\in\{\textrm{S}, \textrm{T}, \textrm{ST}, \textrm{FST}\}$. On the other hand, because the discrete cosine transformation decorrelates the spectral energies, the MFCCs may not maintain locality in the spectral domain. Therefore, we apply only temporal convolutional operations to the MFCCs and denote these CLDNN-based models as X-CLDNN (MFCCs) for $\textrm{X}\in\{\textrm{T}, \textrm{FST}\}$.

\section{Baseline Models}
\label{sec:baseline}
We evaluate our CLDNN-based models for understanding the convolutional operations by comparing with three baseline models on a speech emotion recognition task. First of the baseline models uses the low-level descriptors and their statistical functionals within an utterance to train a support vector machine. The other two of the baseline models are based on the BLSTM recurrent neural networks and take the log-Mels and the MFCCs features as its input, respectively.

\subsection{Support vector machine with the Low-Level Descriptors and Their Statistical Functionals}
\label{subsec:baseline_svm}

Many speech scientific studies have empirically found emotion correlating parameters, 
also known as the low-level descriptors (LLDs), along different aspects of phonation and articulation in speech, such as speech rate in the time domain, fundamental frequency or formant frequency in the frequency domain, intensity or energy in the amplitude domain, or relative energy in different frequency bands in the spectral energy domain. Furthermore, statistical functionals of an entire emotional utterance are derived from the LLDs to obtain global information, complementary to local information captured by frame-level LLDs. Popular selections of these parameters for developing machine learning algorithms in practical applications often amount to several thousands of features. For example, in the INTERSPEECH 2013 computational paralinguistics challenge, the recommended feature set contains $6,373$ parameters of the LLDs and statistical functionals altogether \cite{CompareE13}. Fortunately, researchers have identified the support vector machine as one of the most effective machine learners for using these hand-crafted high-dimensional features \cite{Eyben15}.

To make our work comparable to the published results, we set up the first baseline model similar to the evaluation experiments conducted in \cite{Eyben15}. We use the openSMILE toolkit \cite{Eyben:2010:OMV:1873951.1874246} to extract the acoustic feature sets for INTERSPEECH Challenges from 2009 to 2013 
, including Emotion Challenge (EC, $384$ parameters), Paralinguistic Challenge (PC, $1582$ parameters), Speaker State Challenge (SSC, $4368$ parameters), Speaker Trait Challenge (STC, $5757$ parameters) and Computational Paralinguistic ChallengE (ComParE, $6373$ parameters). On each of these feature sets, we train a SVM for speech emotion recognition. 

\subsection{LDNN with the log-Mels}
\label{subsec:baseline_ldnn-logmels}
As suggested by previous studies \cite{Schuller03,Wollmer10, Metallinou2012, Lee15}, explicit temporal modeling is beneficial for speech emotion recognition, in which a recurrent neural network is a better choice than a hidden Markov model for its outstanding ability to model longer-term temporal relationship. Meanwhile, in order to build a competitive as well as compatible baseline model with respect to the CLDNN-based model, we take the LDNN architecture defined in Sec. \ref{subsec:cldnn} as our second baseline model. In particular, we use the log-Mels as the input to the LDNN model as the "raw" feature set without temporal or spectral convolutional operations. We denote this model as the LDNN (log-Mels).

\subsection{LDNN with the MFCCs}
\label{subsec:baseline_ldnn-mfccs}
MFCCs are related to log-Mels via a mathematical construct: the discrete cosine transformation (DCT). Specifically, the relationship is defined as the following:
\begin{equation}
\label{eq:mfccs}
\textrm{MFCC}[k] = \sum_{m=0}^{M-1} \textrm{log-Mel}[m]\cos\left(\frac{k\pi}{M}\left(m+\frac{1}{2}\right)\right),
\end{equation}
where MFCC[$k$] and log-Mel[$m$] are the $k$th and the $m$th coefficients of MFCCs and log-Mels, respectively, and $M$ is the number of the Mel-scaled filter banks. 

We can easily convert Eq. (\ref{eq:mfccs}) into a convolutional operation along the spectral direction, in which scenario all feature maps are thus tensors of shape $M\times1$. For the $k$th feature map $\mathbf{h}_k$, its $m$th component
\begin{equation}
\mathbf{h}_k[m] = \cos\left(\frac{k\pi}{M}\left(m+\frac{1}{2}\right)\right)
\end{equation}
is pre-defined mathematically based on the prior knowledge of signal processing, rather than task-specifically learnt from training samples. With this development, Eq. (\ref{eq:mfccs}) can be succinctly summarized as 
\begin{equation}
\textrm{MFCC} \triangleq \textrm{DCT-Conv}\left(\textrm{log-Mel}\right),
\end{equation}
where $\textrm{DCT-Conv}$ represents the mathematically pre-defined spectrally only convolutional layer transforming log-Mels into MFCCs. Note that the properties of a conventional convolutional layer, such as the pooling function and the non-linear activation function are missing in this special configuration of a convolutional layer. In fact, there is no convolutional operation per se. Nevertheless, the purpose for this identification of DCT as a convolutional operation is to encapsulate this spectral modeling into the language of convolutional operations, to help us focus on the difference among various convolutional operations and mostly to contrast DCT with the S-Conv layer. 

Our third baseline model is an LDNN model which takes the MFCCs as its input. Similarly, we denote this model as the LDNN (MFCCs). By comparing the the performances of the LDNN (MFCCs) and the S-CLDNN (log-Mels), we are able to quantitatively demonstrate the advantages of the S-Conv layer over the DCT-CNN layer.

\section{Databases Description}
\label{sec:data}
\subsection{The Clean Set}
\label{subsec:cleanset}
We use the eNTERFACE'05 emotion database \cite{enterface05}, which is a publicly-available multi-modal corpus of elicited emotional utterances, to evaluate the performance of our proposed models. Although the entire database contains speech, facial expression and text, in this work we only conduct experiments on the audio modality. 
This database includes $42$ subjects from $14$ various countries, in which $34$ of them were male and $8$ were female. 
Each subject was asked to listen carefully to $6$ short stories, and each of them was designed to elicit a particular emotion from among the $6$ archetypal emotions defined by Ekman et al. \cite{ekman1969pan}. The subjects then reacted to each of the scenarios to express their emotion according to a proposed script in English. Each subject was asked to speak five utterances per emotion class for $6$ emotion classes (anger, disgust, fear, happiness, sadness, and surprise). For each recorded emotional utterance, there is one corresponding global label describing the affective information conveyed by the whole utterance. The resulting corpus, however, is slightly unbalanced in the emotion class distribution because the subject $23$ has only two utterances portraying happiness, so the total number of emotional utterances in this corpus is $1,257$. We call the set of these $1,257$ utterances the clean set. The average length of utterances is around $2.78$ seconds, and the total duration of the clean set amounts to roughly $0.97$ hour.
We believe it is the moderate number of speakers and a variety of their cultural backgrounds that render it one of the most popular corpora for benchmarking speech emotion recognition models. 

\subsection{The Noisy Set}
\label{subsec:noisyset}
Deep neural networks have a well-known reputation of being data-hungry. Despite the aforementioned diversity, it is not data-efficient enough to train a deep neural network as big as a CLDNN on the clean set alone for it would potentially incur a high risk of over-fitting. Various techniques have been proposed to implicitly or explicitly regularize the training process of deep neural networks in order to prevent over-fitting as well as to improve the generalization performance, such as dropout \cite{Srivastava:2014}, early-stopping \cite{Maclaurin16}, data augmentation \cite{Keren16a}, transfer learning \cite{MildeB15} and the recent group convolution approach \cite{Cohen:20165,Dieleman:2016}. In addition to the dropout mechanism and the early-stopping strategy, we also adopt the data augmentation approach to artificially increase the number of our data samples for the purpose of implicit regularization. To be precise, we aggressively mix samples from the clean set with samples from another publicly-available database, called the MUSAN corpus \cite{musan2015}, for a few randomly chosen levels of signal-to-noise ratio (SNR). 

The MUSAN corpus consists of three portions: music, speech and noise. As speech and music may inherently convey affective information, mixing samples from these two portions with clean emotional utterances would unnecessarily complicate the learning process and would possibly result in a suboptimal system due to a mixture of inconsistent emotion types. Therefore, to avoid adding confounding factors to clean emotional utterances, we only use the noise portion in the MUSAN corpus for data augmentation. The noise portion contains $929$ samples of assorted noise types, including technical noises, such as Dual-tone multi-frequency (DTMF) tones, dialtones, fax machine noises, and ambient sounds, such as car idling, thunder, wind, footsteps, paper rustling, rain, animal noises, and so on so forth. The total duration of the noise portion is about $6$ hours. We generate artificially corrupted data based on the clean set using the following recipe. For each clean utterance, $20$ noise samples are uniformly selected from the noise portion and $3$ levels of the SNR are uniformly chosen from the interval $[-10,15]$. Mixing the clean utterance with the $60$ combinations of the $20$ noise samples and $3$ SNR levels augments the clean set by a factor of $61$. Note that randomly selecting samples from the noise portion gives an advantage over simply using a fixed subset of the noise portion. Due to the stochasticity, the probability of choosing the same set of $20$ noise samples is on the order of one out of $7\times 10^{40}\sim C^{929}_{20}$, which is almost impossible. By carefully eliminating potential artificial patterns, we hope the deep neural networks could concisely capture the true underlying acoustic emotion prototypes. 

We call this set of the resulting $75,420$ noisy utterances the noisy set, as opposed to the clean set defined above. The total duration of the noisy set is about $58.25$ hours. In the following sections, when referring to the clean condition, we mean the experiments are conducted on the clean set; on the other hand, when referring to the noisy condition, we mean they are conducted on the union of the noisy and the clean sets. Moreover, we further randomly divide the set of subjects into training, validation and testing (TVT) partitions under the percentage constraint of $70$:$10$:2$0$, respectively, for experimental convenience. Partitioning the subject set, instead of the utterance set, allows us to maintain speaker independence across all experiments.

\section{Speech Emotion Recognition Experiments}
\label{sec:exps}
In this section, we evaluate the proposed models with the following experiments:
\begin{enumerate}
\item Baseline models
\begin{enumerate}
\item SVM with openSMILE features
\item LDNN (MFCCs)
\item LDNN (log-Mels)
\end{enumerate}
\item CLDNN-based models
\begin{enumerate}
\item T-CLDNN (MFCCs)
\item FST-CLDNN (MFCCs)
\item T-CLDNN (log-Mels)
\item S-CLDNN (log-Mels)
\item ST-CLDNN (log-Mels)
\item FST-CLDNN (log-Mels)
\end{enumerate}
\end{enumerate}
The purposes of these experiments are multi-fold. The comparison between the baseline models and the CLDNN-based models aims to demonstrate the effectiveness of the convolutional operations in learning the affective information. Within the category of CLDNN-based models, the goal is to quantify the difference between types of convolutional operations.

\label{sec:serexp}
\subsection{SVM with openSMILE features}
\label{subsec:serexp_svm}
For the first set of baseline experiments, we employ two evaluation strategies. In the first one, we perform a leave-one-subject-out (LOSO) cross validation. Since we train our deep neural network models using the TVT partitions, the second strategy evaluates the performances of SVM classifiers on the TVT partitions for a fair comparison. In addition, we also take the regular pre-processing procedures, including speaker standardization for removing speaker characteristics and class weighting for slight class imbalance. We conduct the baseline experiments using SVM classifiers trained on the acoustic feature sets in the past INTERSPEECH challenges. The SVM classifiers are trained on these hand-crafted high-dimensional features using the Scikit-Learn machine learning toolkit \cite{scikit-learn} with linear, polynomial and radial basis function (RBF) kernels. All of SVM experiments are conducted under the clean condition.  

\subsection{CLDNN-based Models with the MFCCs and the log-Mels}
\label{subsec:serexp_ldnn_cldnn}
To begin with, we extract the log-Mels and the MFCCs using the KALDI toolkit \cite{Povey11thekaldi} with a window size of $25$ ms and a window shift of $10$ ms. In both cases, the number of Mel-frequency filterbanks is chosen to be $40$. It has been shown \cite{Eyben15} that due to the strong energy compaction property of the discrete cosine transformation, the lower order MFCCs are more important for affective and paralinguistic analysis, while the higher order MFCCs are more related to the phonetic content understanding. In fact, the INTERSPEECH challenges feature sets contain the first $12$-$14$ orders of MFCCs; however, the Geneva Minimalistic Acoustic Parameter Set (GeMAPS) \cite{Eyben15} recommends the use of only the first $4$ orders of the MFCCs. In this work, we keep the conventional first $13$ coefficients when computing the MFCCs. After feature extraction, we splice the raw log-Mels and raw MFCCs with a context of $10$ frames in the left and $5$ frames in the right. At this point, each spliced log-Mel or spliced MFCC $\mathbf{x}_t$ lives in $\mathbb{R}^{40 \times 16}$ or $\mathbb{R}^{13 \times 16}$, respectively. An emotional utterance is now represented as a sequence of spliced spectral vectors $\{\mathbf{x}_t\}$. We train the LDNN (log-Mels) and the LDNN (MFCCs) as depicted in Fig. \ref{fig:overview} with their corresponding inputs. 

\begin{table}[ht]
\caption{A summary of the ranges for parameter tuning on each type of the convolutional layers, where $M$ denotes the spectral dimensionality and the subscripts of $h$ and $w$ correspond to the first and the second convolutional layers, respectively.}
\centering
\begin{tabular}{c c c c c c c c c c } 
\Xhline{4\arrayrulewidth}
 & $h_1$ & $w_1$ & $h_2$ & $w_2$  \\
\Xhline{2\arrayrulewidth}
\multicolumn{1}{ c }{\textrm{T-Conv}} & 1 & $3$:$8$ & 1 & $2$:$3$   \\ 
\Xhline{2\arrayrulewidth}
\multicolumn{1}{ c }{\textrm{S-Conv}} & $4$:$9$ & 1 & $3$:$4$ & 1   \\ 
\Xhline{2\arrayrulewidth}
\multicolumn{1}{ c }{\textrm{ST-Conv}} & $4$:$9$ & $3$:$8$ & $3$:$4$ & $2$:$3$   \\
\Xhline{2\arrayrulewidth}
\multicolumn{1}{ c }{\textrm{FST-Conv}} & $M$ & $3$:$8$ & 1 & $2$:$3$  \\
\Xhline{4\arrayrulewidth}
\end{tabular}
\label{tab:conv_tune}
\end{table}

In order to accommodate the inputs to various CLDNN models in Fig. \ref{fig:overview}, we further reshape each $\mathbf{x}_t$ to a matrix $\mathbf{X}_t$ with the shape of $40 \times 16$ or $13\times 16$. We train the X-CLDNN (log-Mels) and X-CLDNN (MFCCs) on the emotional utterances $\{\mathbf{X}_t^\mathbf{u}\}$ for each training utterance $\mathbf{u}$ and for $\textrm{X}\in\{\textrm{S}, \textrm{T}, \textrm{ST}, \textrm{FST}\}$. The ranges of the tunable parameters for the convolutional layers are summarized in Table \ref{tab:conv_tune}, where as shown we focus mostly on the first Conv layer. We exhaust all of the parameter combinations for the S-Conv, T-Conv and FST-Conv types when tuning the architectural parameters. Note that, however, the search space of the optimal parameter set for the ST-Conv is rather huge. Therefore, instead of exploring all of the combinations aimlessly, we limit our attention to the combinations of top $k$ parameters from the S-Conv and T-Conv. 

We use the Keras library \cite{chollet2015keras} on top of the Theano \cite{Theano16} backend to specify the network architectures and execute the learning processes on an NVIDIA K40 Kepler GPU. The weights of all deep neural network models are learnt by minimizing the cross-entropy objective through the Adam method \cite{KingmaB14} to adjust the parameters in the stochastic optimization with an initial learning rate being $0.001$. The size of mini-batch is fixed to $10$ due to the capacity of the GPU memory as well as the pursuit for a better generalizing power \cite{Keskar17}. An early-stopping strategy \cite{Maclaurin16} with the patience of $3$ epochs is employed to avoid over-training. We train all deep neural network models with the emotional utterances in the training partition under the noisy condition; we perform parameter tuning on the validation partition, and the most competitive model on the validation partition under the noisy (clean) condition is tested under the noisy (clean) condition, respectively.   

\section{Experimental Results}
We present our experimental results for speech emotion recognition in this section. Even though the class imbalance in the corpus is insignificant, throughout the entire section, we use the un-weighted accuracy (UA) as the performance metric to avoid being biased to the larger classes. 
\label{sec:exp_result}
\subsection{SVM with openSMILE features}
\label{subsec:result_svm}
Table \ref{tab:result_svm} summarizes the results of using SVM classifiers to identify the emotion class
of an emotional utterance with one of the 6 archetypal emotions. Based on the LOSO evaluation strategy, a SVM with the STC feature set gives the best baseline performance, while under the TVT evaluation strategy, a SVM with the ComParE feature set stands out among other feature sets. It is clear from these results that a SVM learns better from higher-dimensional feature sets such as the ComParE and the STC sets, which is also a consistent phenomenon observed in \cite{Eyben15}. Yan et al. \cite{YanZXLLW16} recently published a baseline result on the eNTERFACE'05 corpus using the PC feature set. They trained a SVM classifier on the PC feature set with a speaker-dependent five-fold cross validation evaluation strategy as one of their baseline models. Their baseline work is comparable to ours, and is included in the Table \ref{tab:result_svm} as well. 

\begin{table}[ht]
\caption{The SVM baseline performance (UA (\%)) based on the leave-one-subject-out (LOSO) cross validation and on the training-validation-testing (TVT) partitions using the acoustic feature sets from past INTERSPEECH challenges.}
\centering
\begin{threeparttable}
\begin{tabular}{c c c c c c c c} 
\Xhline{4\arrayrulewidth}
 & \textbf{EC} & \textbf{PC} & \textbf{SSC} & \textbf{STC} & \textbf{ComParE}  \\
\Xhline{2\arrayrulewidth}
\multicolumn{1}{ c }{\textbf{LOSO}} & \textbf{66.61} & \textbf{73.87} & \textbf{79.19} & \textbf{81.18}  & \textbf{80.45} \\ 
\Xhline{2\arrayrulewidth}
\multicolumn{1}{ c }{\textbf{TVT}} & \textbf{70.83} & \textbf{71.66} & \textbf{77.92} & \textbf{80.00}  & \textbf{80.83} \\ 
\Xhline{2\arrayrulewidth}
\multicolumn{1}{ c }{\textbf{Yan et al. \cite{YanZXLLW16}}} & -- & \textbf{74.21} & -- & --  & -- \\
\Xhline{4\arrayrulewidth}
\end{tabular}
\begin{tablenotes}\footnotesize
\item[*] Emotion Challenge (EC), Paralinguistic Challenge (PC), Speaker State Challenge (SSC), Speaker Trait Challenge (STC), Computational Paralinguistic ChallengE (ComParE)
\end{tablenotes}
\end{threeparttable}
\label{tab:result_svm}
\end{table}

\begin{table}[ht]
\caption{The performances (UA (\%)) of the optimal SVM model, the LDNN-based models and the CLDNN-based models. The sparse kernel reduced rank regression (SKRRR) \cite{YanZXLLW16} is one of the state-of-the-art models on the eNTERFACE'05 corpus.}
\centering
\begin{tabular}{ c c c c c c c c } 
\Xhline{4\arrayrulewidth}
\multicolumn{1}{ c } {\textbf{Model (features)}} & \textbf{noisy} & \textbf{clean} \\
\Xhline{4\arrayrulewidth}
\multicolumn{1}{ c } {\textbf{SVM (ComParE)}} & -- & \textbf{80.83}\\ 
\Xhline{2\arrayrulewidth}
\multicolumn{1}{ c } {\textbf{SKRRR \cite{YanZXLLW16}}} & -- & \textbf{87.46}\\ 
\Xhline{2\arrayrulewidth}
\multicolumn{1}{ c } {\textbf{LDNN (MFCCs)}} & \textbf{75.51} & \textbf{88.33}\\ 
\Xhline{2\arrayrulewidth}
\multicolumn{1}{ c } {\textbf{LDNN (log-Mels)}} & \textbf{78.87} & \textbf{90.42}\\ 
\Xhline{4\arrayrulewidth}
\multicolumn{1}{ c } {\textbf{T-CLDNN (MFCCs)}} & \textbf{83.44} & \textbf{87.92}\\ 
\Xhline{2\arrayrulewidth}
\multicolumn{1}{ c } {\textbf{FST-CLDNN (MFCCs)}} & \textbf{84.45} & \textbf{92.92}\\ 
\Xhline{2\arrayrulewidth}
\multicolumn{1}{ c } {\textbf{T-CLDNN (log-Mels)}} & \textbf{84.23} & \textbf{92.92}\\ 
\Xhline{2\arrayrulewidth}
\multicolumn{1}{ c } {\textbf{S-CLDNN (log-Mels)}} & \textbf{82.73} & \textbf{91.67}\\ 
\Xhline{2\arrayrulewidth}
\multicolumn{1}{ c } {\textbf{ST-CLDNN (log-Mels)}} & \textbf{84.26} & \textbf{93.75}\\ 
\Xhline{2\arrayrulewidth}
\multicolumn{1}{ c } {\textbf{FST-CLDNN (log-Mels)}} & \textbf{86.21} & \textbf{94.58}\\ 
\Xhline{4\arrayrulewidth}
\end{tabular}
\label{tab:result_cldnns}
\end{table}

\subsection{LDNN with the MFCCs and the log-Mels}
\label{subsec:result_ldnn_mfccs_logmels}
We present the results of the LDNN-based models in Table \ref{tab:result_cldnns}. Under the noisy condition, the LDNN (MFCCs) and the LDNN (log-Mels) models are able to accurately classify $75.51\%$ and $78.87\%$ of the testing samples, respectively. Under the clean condition, they give a performance of $88.33\%$ and $90.42\%$, respectively. One can easily observe that there is a gap of $3.36\%$ and $2.09\%$, respectively, between LDNN (MFCCs) and LDNN (log-Mels) under each condition. Since MFCCs are DCT transformed log-Mels, it implies that DCT may have removed a certain amount of affective information when transforming the log-Mels into the MFCCs. The widened gap under the noisy condition also suggests MFCCs are more sensitive to noise compared to log-Mels, which renders learning from MFCCs a more challenging task. Nevertheless, both LDNN models achieve promising results comparable to that by one of the state-of-the-art models on the eNTERFACE'05 corpus, the sparse kernel reduced rank regression (SKRRR) \cite{YanZXLLW16}.

\subsection{CLDNN with the MFCCs and the log-Mels}
\label{subsec:result_cldnn_mfccs_logmels}
Finally, Table \ref{tab:result_cldnns} also presents the effectiveness of the CLDNN-based models for classifying emotional utterances into one of the 6 archetypal emotions. First of all, notice that with the CNN layers all CLDNN-based models improve upon their LDNN-based  counterparts under both noisy and clean conditions, except that the T-CLDNN (MFCCs) results in a slightly inferior performance under the clean condition. 
Since MFCCs are rather sensitive to noise, it is likely that the T-Conv layers are mainly optimized to reduce prominent variations due to the artificial noise while neglecting other subtle factors of variation such as speaker or gender. 
Yet, the result from the FST-CLDNN (MFCCs) also suggests that the MFCCs still contain a reasonable amount of affective information which is learnable by a suitable architecture. 

Among the X-CLDNN (log-Mels) models, the order of performances from high to low is the FST-CLDNN (log-Mels), the ST-CLDNN (log-Mels), the T-CLDNN (log-Mels) and the S-CLDNN (log-Mels). The fact that the FST-Conv outperforms the ST-Conv is consistent with the conclusion from \cite{Anand} under the clean condition. However, the margin is not as significant when there is an LDNN sub-network to help with temporal modeling. It has been reported that the S-Conv layer in a S-CLDNN (log-Mels) would degrade the performance for speech recognition under a moderately noisy condition \cite{SainathLi16}. The authors attributed this deterioration to the noise-enhanced difficulty for local filters  of small sizes to make decision when learning to capture translational invariance. This attribution seems valid when we contrast the FST-Conv with the other three types. Actually, if we take a closer look, we can easily discover that there is a varying degree of enhanced difficulty to the type of convolutional operations, in which the S-Conv suffers from noise the most, followed by the T-Conv and the ST-Conv to a roughly equivalent degree and finally the FST-Conv the least. Even though we validate on the clean validating partition for selecting the model to be tested on the clean testing partition, the performances under the clean condition demonstrate a similar trend influenced by noise since we carried out the training process under the noisy condition. 

One of our goals is to benchmark the strength of the S-Conv and the discrete cosine transformation for spectral modeling. Specifically, the fair comparison should be between the LDNN (MFCCs) and the S-CLDNN (log-Mels) where the DCT-CNN and the S-Conv layers, respectively, act on the spliced log-Mels along the spectral direction, and both of them have an LDNN sub-network for further temporal modeling. Despite the negative impact on the S-Conv layer by noise, it is interesting to observe a stark performance gap between them under the noisy condition. Even under the clean condition, the S-CLDNN (log-Mels) still has a leading margin by more than $3\%$. 
Due to its task independence, DCT is not particularly designed to decorrelate the affective information from the other factors. 
Moreover, since the DCT-CNN layer is shallow and structurally simple, the S-Conv layer has an advantage over DCT as it is deeper and thus better at disentangling the underlying factors of variations \cite{PascanuGCB14,Glorot11,Goodfellow09}. This strength is manifested the most especially when it comes to the noise-related factors.  
Given that the MFCCs still carry a reasonable amount of affective information, these significant differences in performance between the S-Conv and DCT can be best explained by the inability of DCT to adequately disentangle the affective information from other irrelevant factors of variations. 

Last but not the least, we notice that temporally convolutional operations and temporally recurrent operations are learning complimentary information. For instance, the LDNN (log-Mels) models the evolution of affective information through temporal recurrence alone, while the FST-CLDNN (log-Mels) does so by fitting itself to the dynamics via temporal convolution and then temporal recurrence, which improves upon the LDNN (log-Mels) and results in a more competitive system.

\subsection{Finer hyper-parameter search on the spectral axis}
\label{subsec:finesearch}
\begin{figure}[th]
  \centering
  \includegraphics[width=0.45\textwidth,height=4cm]{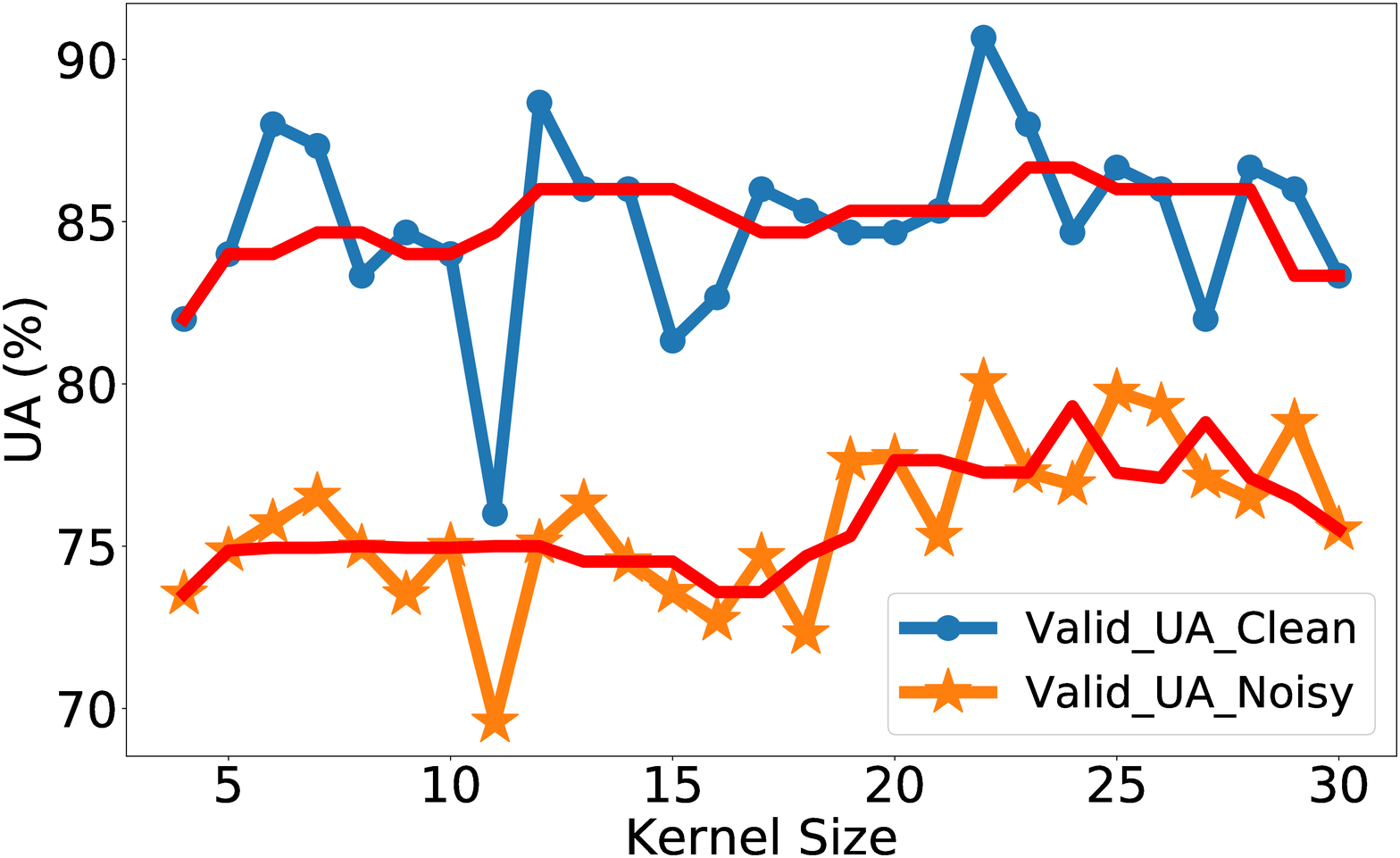}
  \caption{Unweighted accuracy, UA (\%), of the S-CLDNN (log-Mels) model on the validation partition under noisy and clean conditions with respect to different kernel sizes $h_1$ in the first convolutional layer. The curves in red are median filtered UAs.}
\label{fig:speconly_kernel_size}
\end{figure}
We have seen the negative effect of noise on the S-Conv for speech emotion recognition in Table \ref{tab:result_cldnns} and speech recognition in \cite{SainathLi16}. The authors hypothesized that noise has increased the difficulty for these local filters to correctly capture translational invariance. On the other hand, the performance shown by the FST-CLDNN (log-Mels) model suggests that global information over the entire spectrum helps to learn a better representation. To gain more insight into how convolving with more spectral information contributes to affective learning, we further conduct an extensive search on the spectral axis for the optimal kernel size in the first convolutional layer of the S-CLDNN (log-Mels) model, i.e. $h_1$ in Table \ref{tab:conv_tune}. For the search, we fix $h_2=3$ and the pooling hyper-parameters the same as in Sec \ref{subsec:typesconv}. We iterate the filter height $h_1$ through all possible sizes from $4$ to $30$ (to allow pooling and convolution in the second layer). 

Fig. \ref{fig:speconly_kernel_size} depicts the validation UA under clean and noisy conditions with respect to different kernel size $h_1$. Although highly fluctuating possibly due to the influence of noise, the accuracy is indeed improving along with a larger kernel size until it peaks at $h_1=22$ for both conditions, and increasing the kernel size larger than $22$ does not result in any further improvement. Second, from the median filtered curves, the S-Conv is able to benefit more under the noisy condition from having a larger kernel size, specifically $h_1>18$ in Fig. \ref{fig:speconly_kernel_size}, which suggests a phase transition from small to large filters; however, such a pattern is not equally significant when under the clean condition as the curve is relatively flat. Third, when $h_1=22$, the respective test UA are $85.87\%$ and $95.42\%$ under the noisy and clean conditions. Despite the outstanding performance under the clean condition, when compared with the FST-CLDNN (log-Mels) model, these results further highlight the influence of noise on the S-Conv operation as well as the robustness of two-dimensional filters to noise \cite{DChanL15} even though it has convolved with the optimal amount of spectral information.

Overall, this extended set of experiments demonstrate one of the advantages of convolving with more spectral information, emphasizing on the ability to counter the negative effect due to noise in learning. Since S-Conv shows two characterizations with small and large filter sizes and to convolve with more spectral information is one of the characteristics of the FST-Conv in addition to being two-dimensional, therefore, we will continue to refer to S-Conv as defined in Table \ref{tab:conv_tune} for consistency in this work.

\subsection{Module-wise evaluations}
\label{subsec:modulewise}
We have so far analyzed the proposed models from an end-to-end perspective and observed interesting phenomena. Although this kind of external analysis has distilled certain working knowledge, what we are equally interested in is the internal mechanism within these models. Along these lines, a key step is to track the flow of relevant information using techniques such as information regularizer \cite{Huang16} or layer-wise evaluation  \cite{alain16,Huang17}. In this work, we take the second approach due to its simplicity. To make it clear, we only evaluate the intermediate representations at the module level, where by module we mean the CNN module (two Conv layers), the BLSTM module (a BLSTM layer) and the multi-layer perceptron (MLP) module (four FC layers) that make up a CLDNN model.

To begin with, we take the trained CLDNN-based model as the feature extractors and the activated responses of each layer as the discriminative features. For each CLDNN model, we only keep the extraction from the output layer of each module. In addition, the raw spectral-temporal features are presented to serve as the lower bound. A mean pooling over the temporal direction is applied to the raw features, the output of the CNN module and the output of the BLSTM module to form an utterance representation for each of them. In order to quantify the improvement of the representations for speech emotion recognition achieved by each module, we train a SVM classifier on the utterance representation from the output of each module as well as the raw features. The experiment setting is similar to the SVM baseline, where only the clean set is used and the evaluation is based on the TVT strategy. 

\subsubsection{Quantitative Analysis}
\label{subsubsec:quant_anal}
\begin{table}[ht]
\caption{The performances (UA (\%)) of a SVM classifier trained on the spliced log-Mels, the spliced MFCCs and the output of each module from all CLDNN-based models under the clean condition.}
\centering
\begin{tabular}{c c c c c c c c} 
\Xhline{4\arrayrulewidth}
 {\textbf{Model (features)}} & \textbf{Raw} & \textbf{CNN} & \textbf{BLSTM} & \textbf{MLP}   \\
\Xhline{4\arrayrulewidth}
\multicolumn{1}{ c }{\textbf{T-CLDNN (MFCCs)}} & \textbf{23.75} & \textbf{52.50} & \textbf{88.75} & \textbf{88.75}  \\ 
\Xhline{2\arrayrulewidth}
\multicolumn{1}{ c }{\textbf{FST-CLDNN (MFCCs)}} & \textbf{23.75} & \textbf{56.25} & \textbf{88.75} & \textbf{92.50}  \\ 
\Xhline{2\arrayrulewidth}
\multicolumn{1}{ c }{\textbf{T-CLDNN (log-Mels)}} & \textbf{27.92} & \textbf{59.17} & \textbf{93.33} & \textbf{93.33}  \\ 
\Xhline{2\arrayrulewidth}
\multicolumn{1}{ c }{\textbf{S-CLDNN (log-Mels)}} & \textbf{27.92} & \textbf{45.83} & \textbf{88.33} & \textbf{91.67}  \\ 
\Xhline{2\arrayrulewidth}
\multicolumn{1}{ c }{\textbf{ST-CLDNN (log-Mels)}} & \textbf{27.92} & \textbf{55.83} & \textbf{89.17} & \textbf{93.75}  \\ 
\Xhline{2\arrayrulewidth}
\multicolumn{1}{ c }{\textbf{FST-CLDNN (log-Mels)}} & \textbf{27.92} & \textbf{54.17} & \textbf{89.17} & \textbf{94.58}  \\ 
\Xhline{4\arrayrulewidth}
\end{tabular}
\label{tab:modulewise}
\end{table}

\begin{figure*}[th]
\centering
  \begin{subfigure}[b]{0.33\linewidth}
    \centering
    \includegraphics[width=1\textwidth]{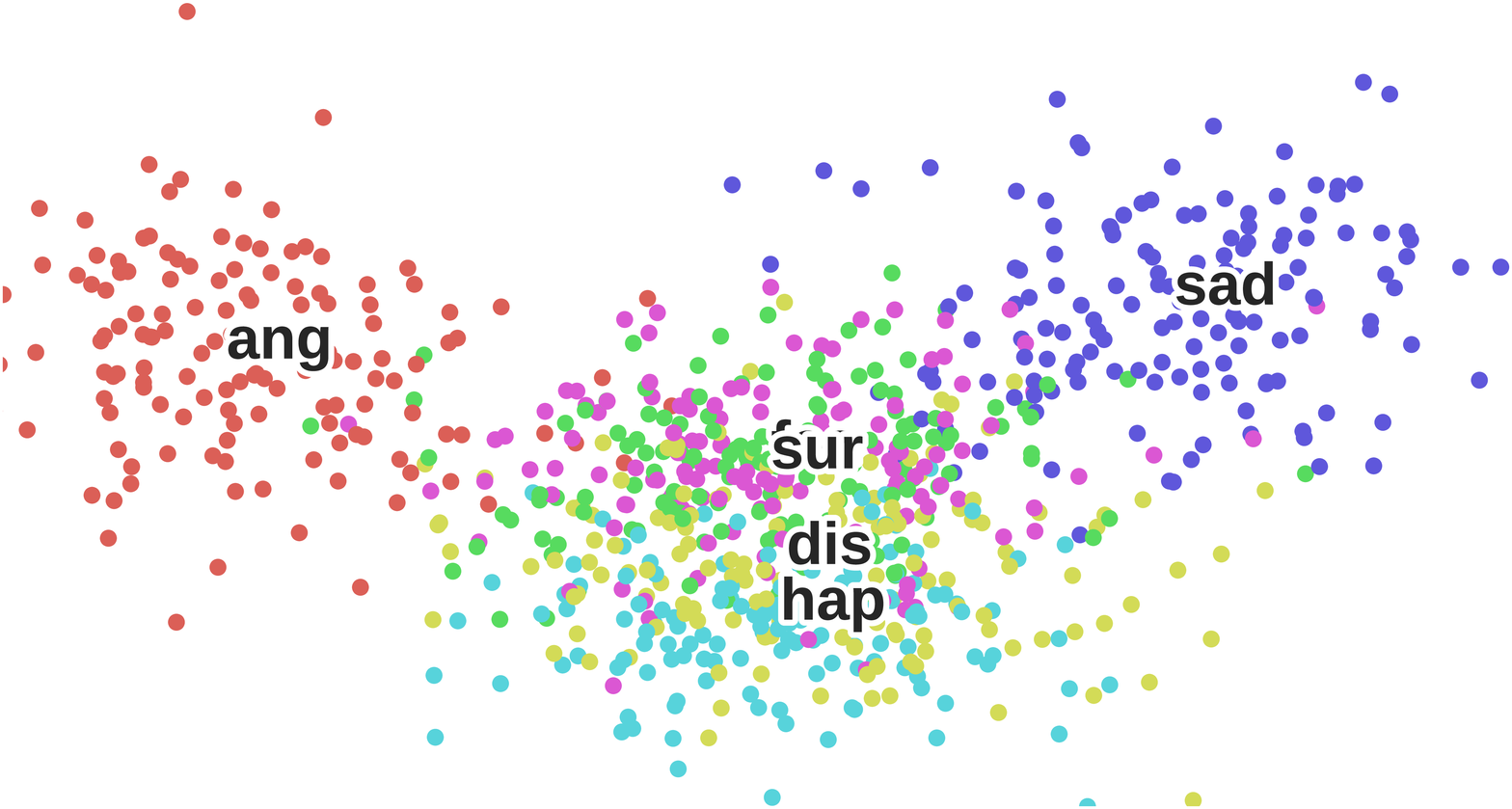}
    \caption{$2.506$}
    \label{fig:tcldnn_logmels_act_emo}
  \end{subfigure}%
  \begin{subfigure}[b]{.33\linewidth}
    \centering
    \includegraphics[width=1\textwidth]{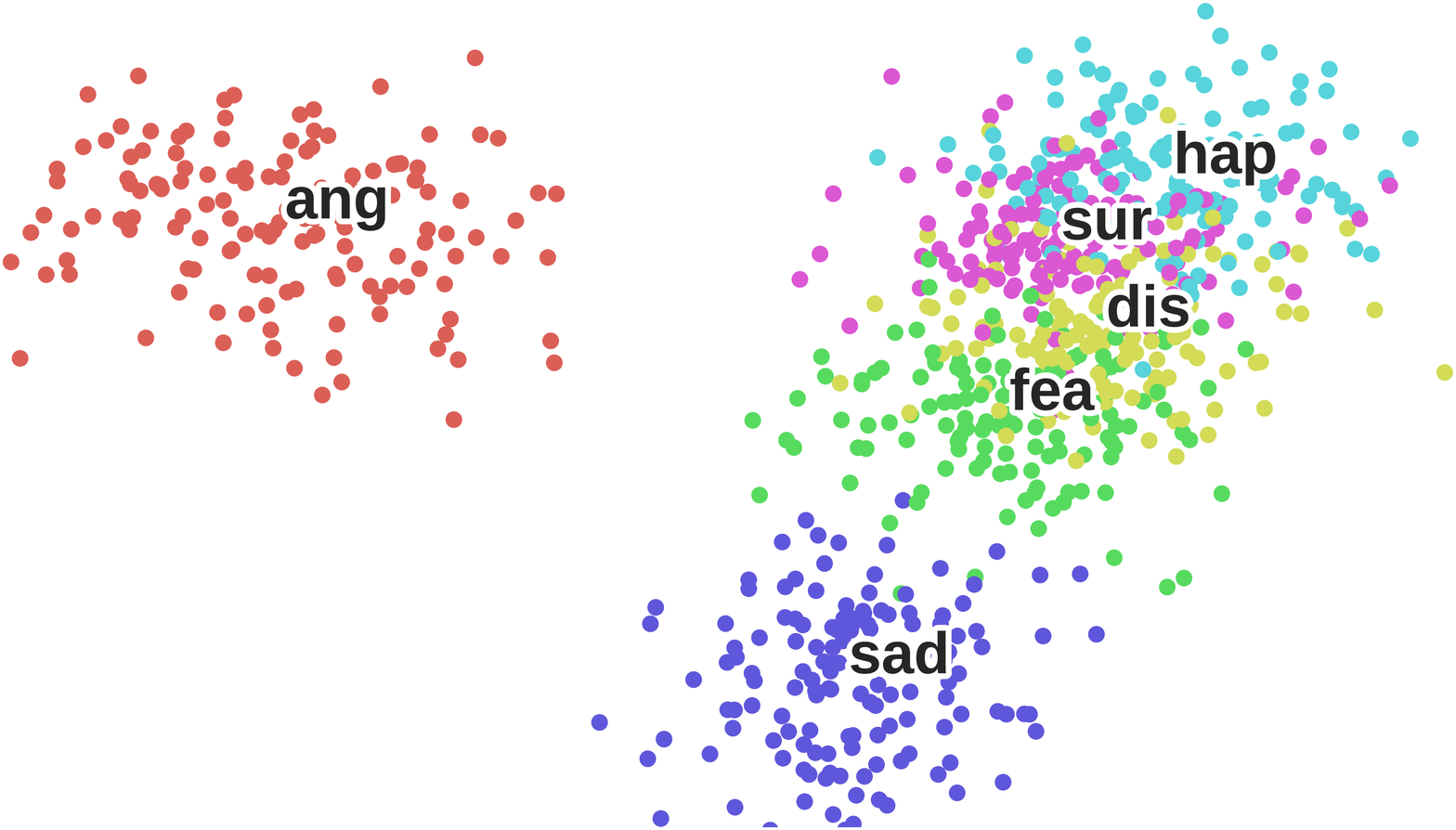}
    \caption{$1.148$}
    \label{fig:tcldnn_logmels_tpool_emo}
  \end{subfigure}%
  \begin{subfigure}[b]{.33\linewidth}
    \centering
    \includegraphics[width=1\textwidth]{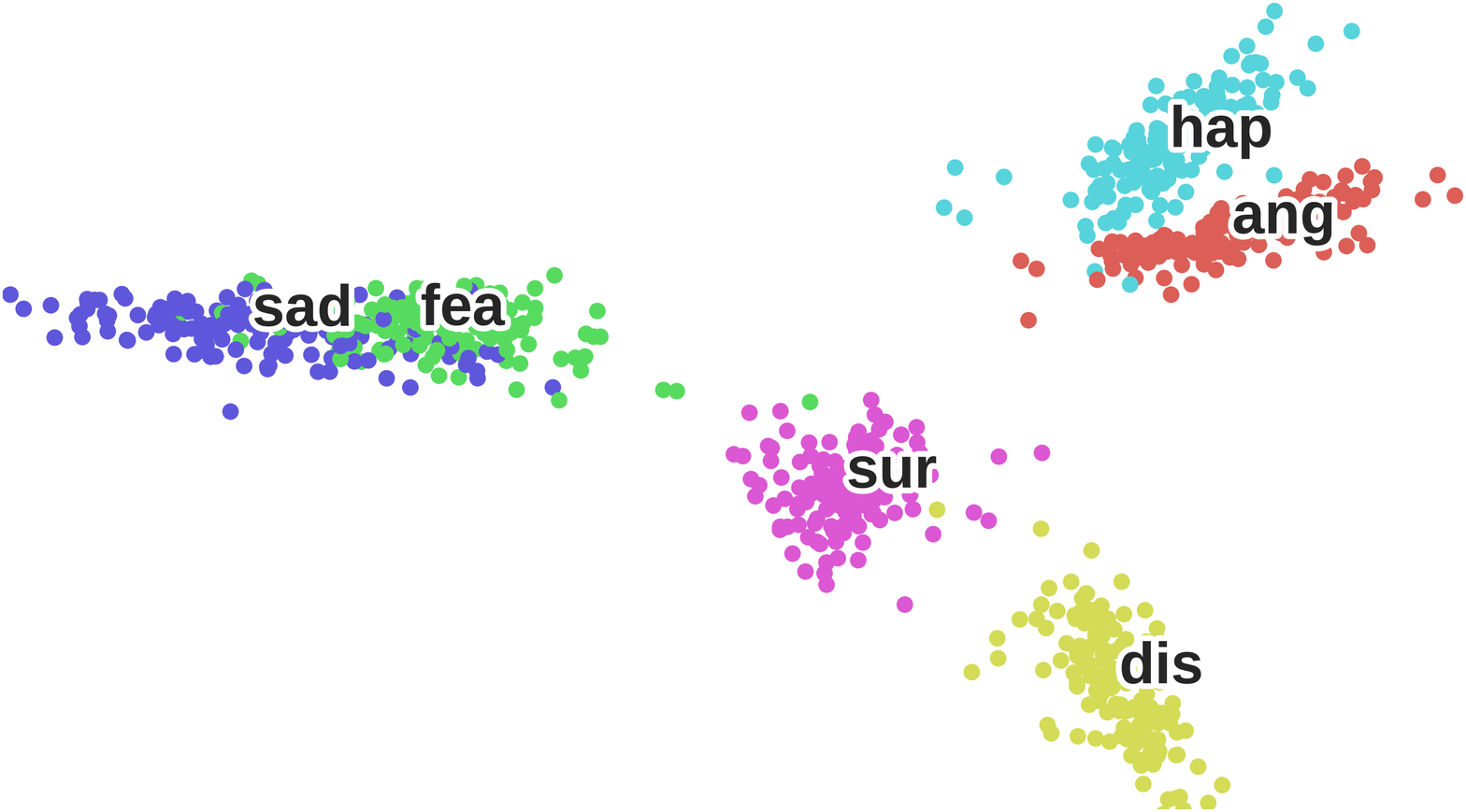}
    \caption{$0.263$}
    \label{fig:tcldnn_logmels_lastfc_emo}
  \end{subfigure}\\%
  
  \begin{subfigure}[b]{0.33\linewidth}
    \centering
    \includegraphics[width=1\textwidth]{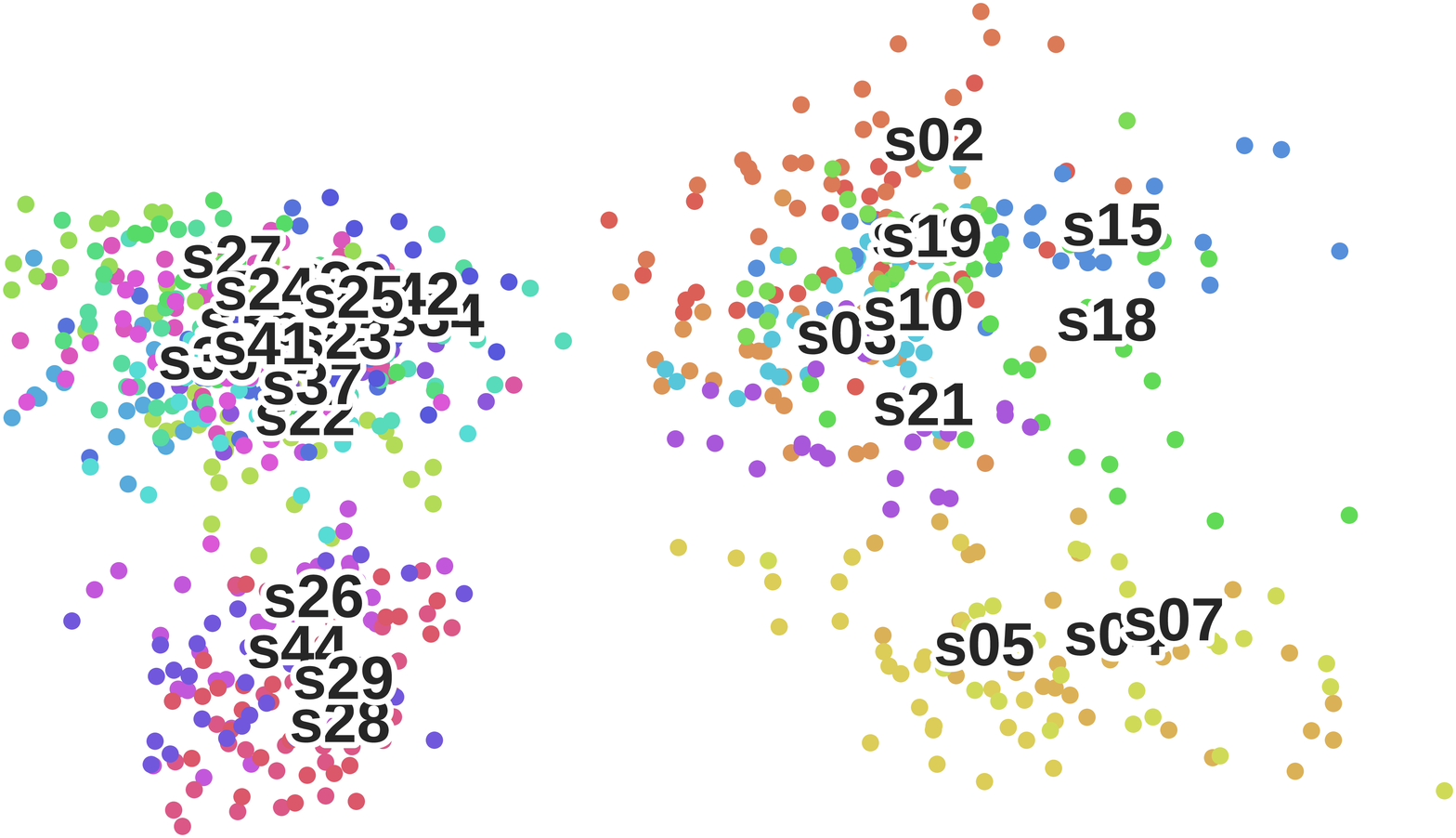}
    \caption{$1.298$}
    \label{fig:tcldnn_logmels_act_spkr}
  \end{subfigure}%
  \begin{subfigure}[b]{.33\linewidth}
    \centering
    \includegraphics[width=1\textwidth]{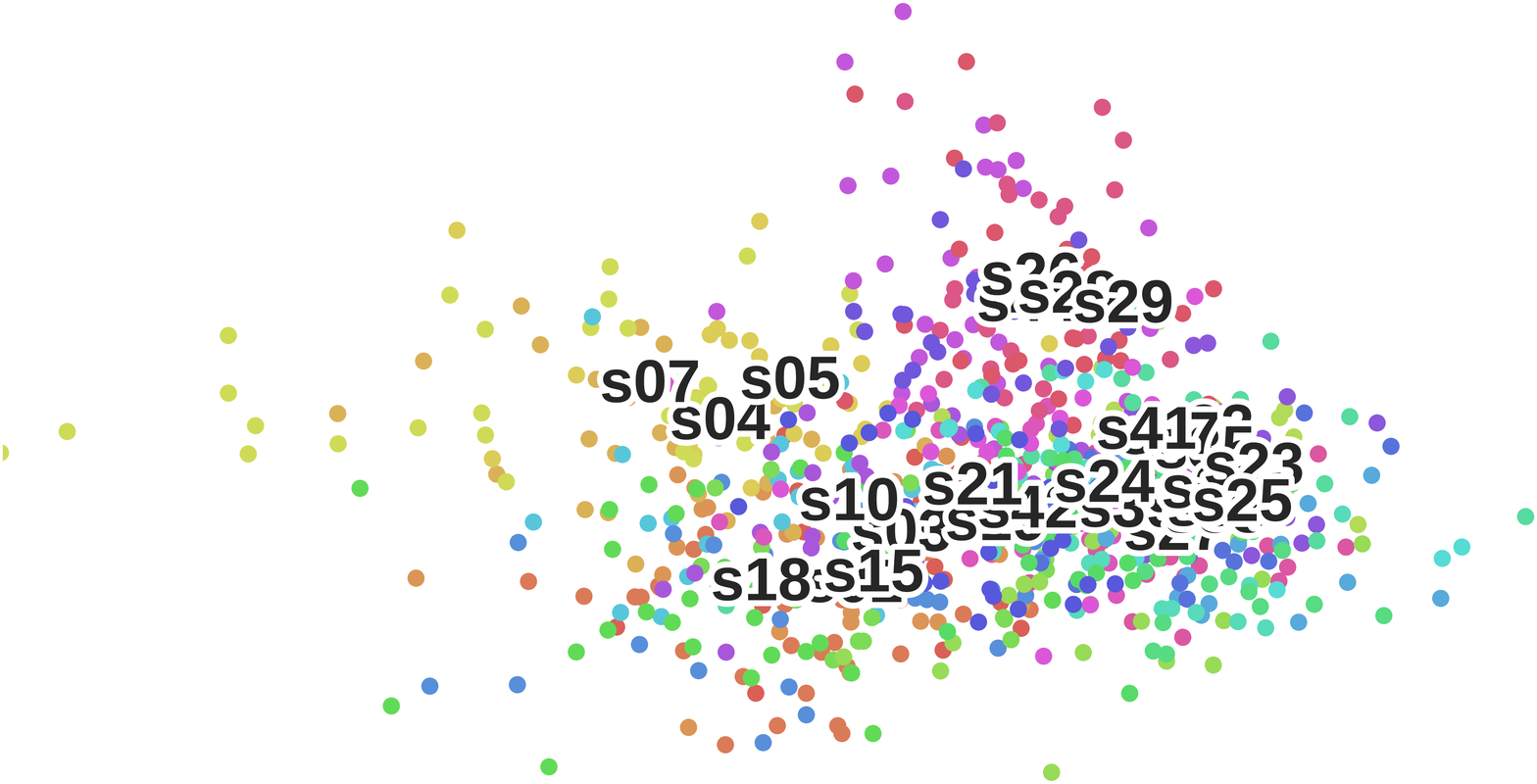}
    \caption{$2.800$}
    \label{fig:tcldnn_logmels_tpool_spkr}
  \end{subfigure}%
  \begin{subfigure}[b]{.33\linewidth}
    \centering
    \includegraphics[width=1\textwidth]{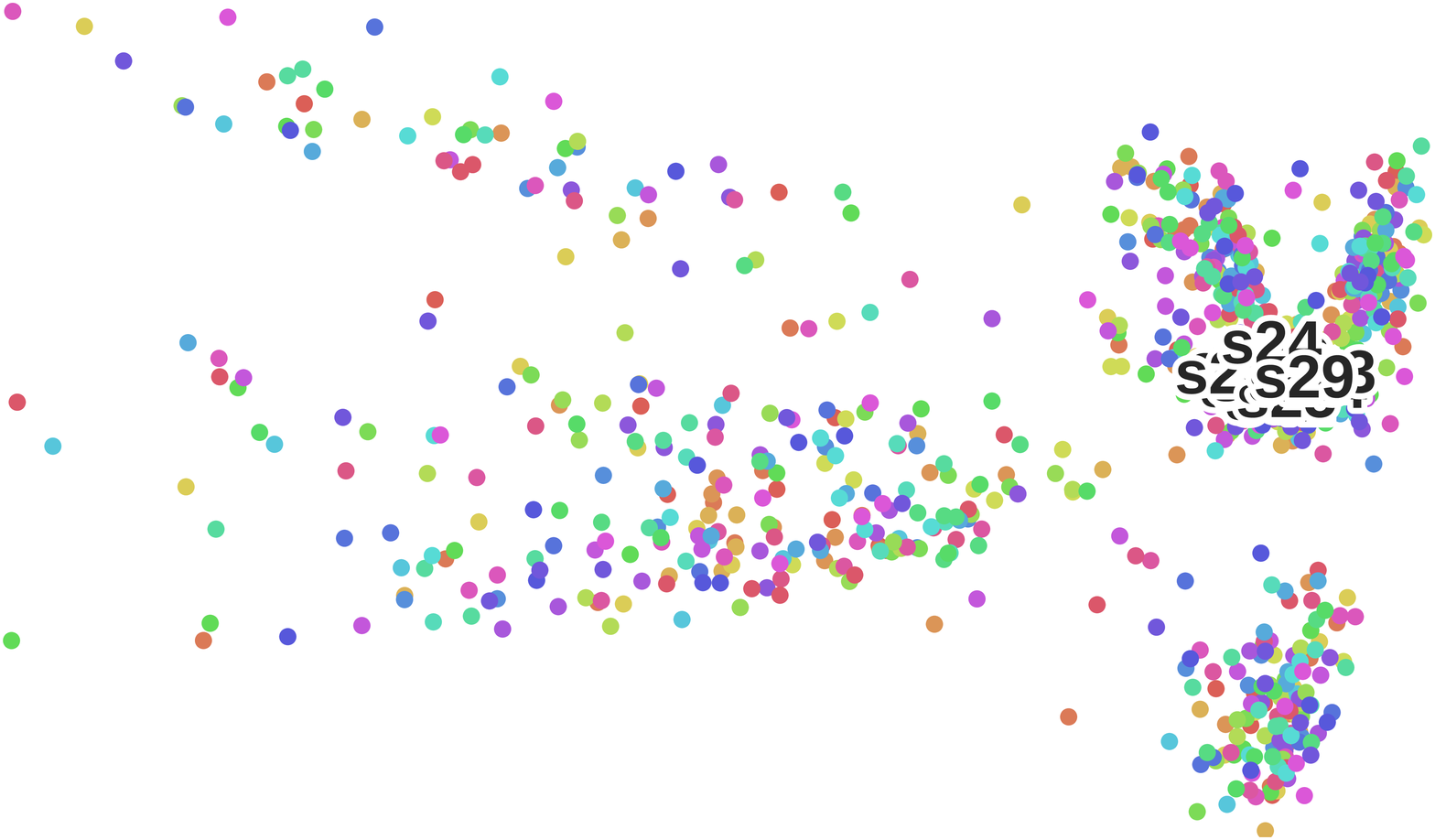}
    \caption{$7.762$}
    \label{fig:tcldnn_logmels_lastfc_spkr}
  \end{subfigure}\\%
  
  \begin{subfigure}[b]{0.33\linewidth}
    \centering
    \includegraphics[width=1\textwidth]{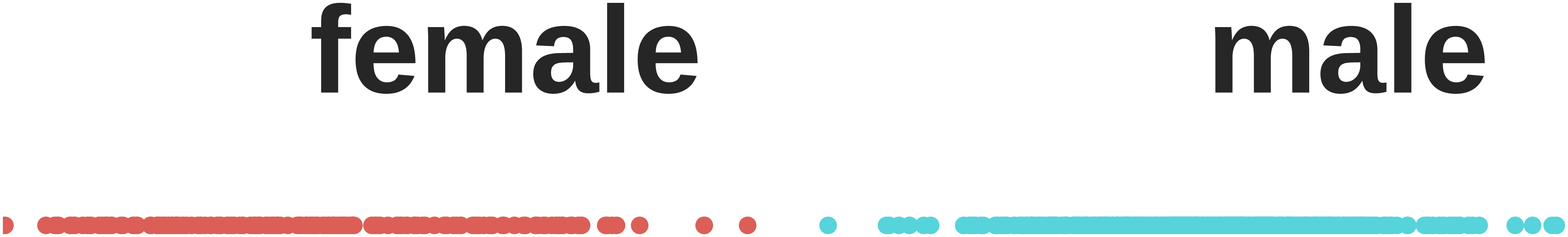}
    \caption{$2.988$}
    \label{fig:tcldnn_logmels_act_gender}
  \end{subfigure}%
  \begin{subfigure}[b]{.33\linewidth}
    \centering
    \includegraphics[width=1\textwidth]{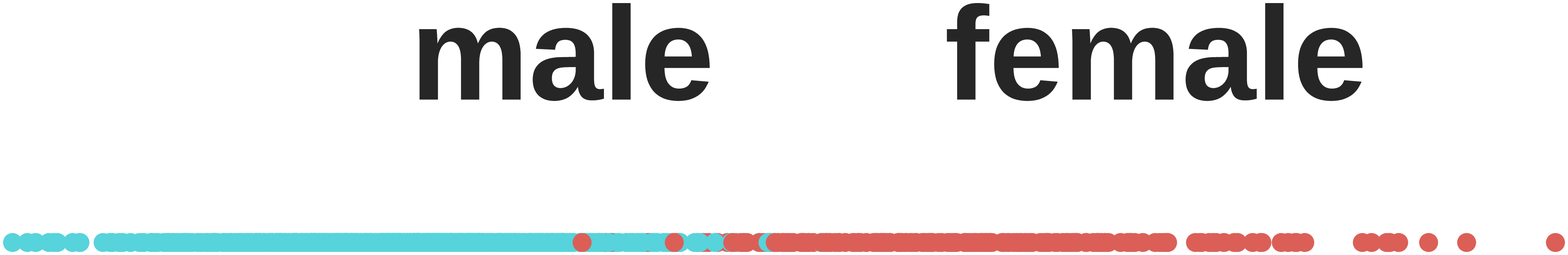}
    \caption{$5.292$}\label{fig:tcldnn_logmels_tpool_gender}
  \end{subfigure}%
  \begin{subfigure}[b]{.33\linewidth}
    \centering
    \includegraphics[width=1\textwidth]{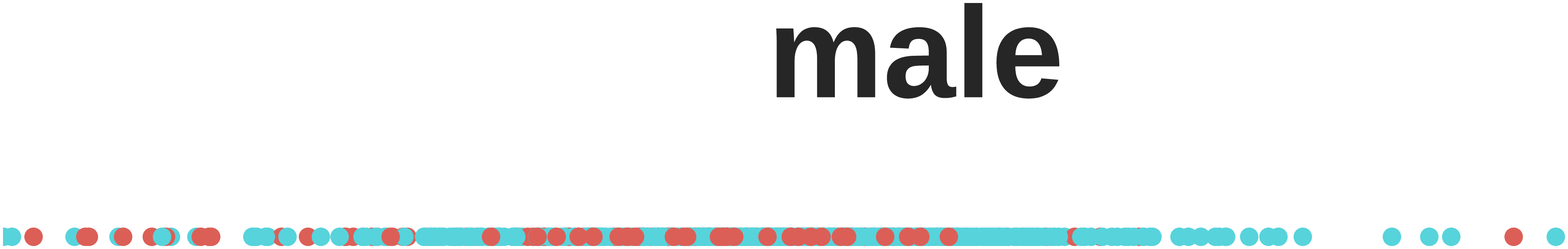}
    \caption{$17.292$}\label{fig:tcldnn_logmels_lastfc_gender}
  \end{subfigure}\\%
  \caption{The visualization for the modules in the T-CLDNN (log-Mels). The first, second and third rows correspond to the affective, speaker and gender information, while the first, second and third columns denote the output of the CNN, the BLSTM and the MLP modules, respectively. In each subplot, every dot indicates an utterance, where utterances within the same class are painted with the same color and their centers of classes are marked with according labels such as \textit{hap}, \textit{s07} and \textit{female}. The title of each subplot is the $\mathbf{\rho}$ value, i.e. the quality measure of a clustering, for the distributions in the subplot.}
 \label{fig:tcldnn_logmels}
\end{figure*}

Table \ref{tab:modulewise} summarizes the results of the module-wise evaluation. As shown in the second column, even though the training and the testing are carried out under the clean condition, the discrete cosine transformation degrades the performance once again. Nevertheless, most of the CNN modules have helped to lift the discriminative power to around $55\%$ regardless of the raw features except for a particularly under-performing model, the S-CLDNN (log-Mels), which based on the previous analysis is known to suffer from noise drastically. One can easily observe that each type of the Conv layers is learning a different representation and hence results in different levels of discriminative power. 

It is interesting to note that the SVMs trained on the activations of the CNN module in the \{T,ST\}-CLDNN (log-Mels) give a better accuracy than that based on the FST-CLDNN (log-Mels), but from a holistic perspective the FST-Conv based system is the most robust one. This may reflect one of the biggest advantages of the end-to-end training approach over the traditional layer-wise approach, which works on feature engineering and classifier training separately; i.e. a greedy layer-wise training that forces the distribution of an intermediate layer to prematurely approximate the distribution of the label is likely to result in a suboptimal system.  

Going deeper into the networks, we can see most of the BLSTM modules have further improved the discriminative power to the level of $88$-$89\%$ except for the T-CLDNN (log-Mels). In fact, as we take a closer look at the T-CLDNN (MFCCs) and the T-CLDNN (log-Mels), we find that they both attain one of their optimal forms of affective representation at the output of the BLSTM module. Instead of implying their MLP modules have done nothing based on the constant performance, it may suggest that their MLP modules are integrating out irrelevant information while maintaining the optimal representation. Finally, in the other CLDNN models, the MLP modules further refine the representation to make the prediction an easier task. To sum up, in terms of the UA, the contributions from the CNN module, the BLSTM module and the MLP module are $27.43\pm 5.18\%$, $35.63\pm 3.61\%$ and $2.85\pm 2.32\%$, respectively. 

\subsubsection{Visualization}
\label{subsubsec:visual}

In addition to the quantitative analysis of each module, we also present the visualization of the representations to gain intuition toward the internal working mechanism. In order to demonstrate the interplay between the modules and the other irrelevant information, we take into consideration two other types of information which along with the affective information are embedded in the original utterances at the same time; that is, the gender and speaker information. For every representation extracted from each module, we assign three labels to it, including the gender of the speaker (\textit{female, male}), the serial number of the speaker (\textit{sN}, where $1\leq \textit{N} \leq 42$) and the emotional class (\textit{ang} for \textit{anger}, \textit{dis} for \textit{disgust}, \textit{fea} for \textit{fear}, \textit{hap} for \textit{happiness}, \textit{sad} for \textit{sadness}, and \textit{sur} for \textit{surprise}). On the clean training partition, a linear discriminant analysis (LDA) is applied to the representations and projects them onto the space spanned by the first $\min(N-1,2)$ components, where $N$ stands for the number of classes. Each LDA is carried out with respect to these three labels separately and a class prior is employed to match the number of samples in each class. Moreover, we also compute the intra-cluster and inter-cluster inertia \cite{Lebart79,Lamirel2012} on the extracted representations for each label using the following definitions
\begin{eqnarray}
\mathbf{\omega}_i 
&=& 
\frac{1}{|C|}\sum_{c\in C} \frac{1}{|c|} \sum_{s\in c} 
\lVert \mathbf{p}_c-\mathbf{p}_s \rVert^2, 
\\
\mathbf{\omega}_o 
&=& 
\frac{1}{|C|^2-|C|}\sum_{c\in C} 
\sum_{c^\prime\in C-{c}}
\lVert \mathbf{p}_{c^\prime}-\mathbf{p}_c \rVert^2,
\end{eqnarray}
where $C$ is the training set, $c$ is the subset of $C$ containing only a specific class, $\mathbf{p}_c$ is the arithmetic center of $c$, $\mathbf{p}_s$ is the member of $c$ and $\lVert\cdot\rVert$ is the Euclidean distance. Note that the vectors $\mathbf{p}$ are the original representations rather than the LDA projections. One may expect to have a small intra-cluster inertia and a large inter-cluster inertia when assessing the quality of a good clustering; in other words, the following ratio measures the quality of a clustering

\begin{equation}
\mathbf{\rho} = \mathbf{\omega}_i / \mathbf{\omega}_o, 
\end{equation}
where the smaller the value of $\mathbf{\rho}$ the better a clustering.

Fig. \ref{fig:tcldnn_logmels} shows an example of the visualization for the modules in the T-CLDNN (log-Mels). The first, second and third rows correspond to the affective, speaker and gender information, while the first, second and third columns denote the output of the CNN, the BLSTM and the MLP modules, respectively. In each subplot, every dot indicates an utterance, where utterances within the same class are painted with the same color and their centers of classes are marked with according labels such as \textit{hap}, \textit{s07} and \textit{female}. The title of each subplot is the $\mathbf{\rho}$ value for the distributions in the subplot. 

Based on the visualization or the $\mathbf{\rho}$ values in the first row, it is clear that the CLDNN model is gradually learning to discriminate different affective patterns. Out of the six emotion classes, anger consistently seems to be the most prominent class across different architectures, and sadness is ranked the second. The progressively improving separability in the first row confirms our quantitative analysis results as well.

On the other hand, the speaker and the gender information are rather salient at the beginning of the architecture. As the forward propagation proceeds, these two types of information are getting filtered out incrementally. Note that the LDA projections on the second and the third rows are computed on the raw extracted representations with their respective labels, i.e. speaker and gender labels, and yet the deteriorating separability is apparently evident from the scatter plots and the increasing trend of $\mathbf{\rho}$. 

Based on the results of the quantitative analysis, we thought it is the BLSTM module that discards most amount of irrelevant information compared to the other modules. However, contrary to our initial expectation, it is the MLP module that excessively degrades the separability among speaker or gender classes. For instance, even at the output of the BLSTM module, the model still keeps a fair amount of gender information (Fig. \ref{fig:tcldnn_logmels_tpool_gender}) but at the output of the MLP module the centers of the male and the female utterances are practically overlapping each other (Fig. \ref{fig:tcldnn_logmels_lastfc_gender}). 
Previous studies have shown that the higher-level representation of a deep neural networks could better disentangle the underlying factors of variations embedded in the input signals \cite{PascanuGCB14,Glorot11,Goodfellow09}.
This visualization suggests that the CNN and the BLSTM modules are mostly playing a role to lift the input tensor into a high-dimensional manifold, a role similar to the kernel method, for disentangling the affective factor from the others, and consequently the MLP module is mainly responsible for integrating out the other factors of variations in order to optimize the corresponding objective function. 
In addition, this observation also vividly explains the working mechanism of multi-tasking learning that learns multiple related tasks jointly by sharing a common sub-network in the front, and of transfer learning approach that freezes the underlying layers in a pre-trained model and re-learns or fine-tunes the top few, often fully-connected, layers.

The progression from the second column to the third column corroborates our working hypothesis in the quantitative analysis about the T-CLDNN models as well. Instead of doing nothing, the MLP module in the T-CLDNN model is refining the representations while keeping the affective information. 

For the visualization of all CLDNN-based models, please refer to Supplemental Materials.
\label{sec:disc}
\section{Conclusion}
\label{sec:conclu}
We report the benchmarking of four types of convolutional operations in deep convolutional recurrent neural networks for speech emotion recognition, including the spectrally only, the temporally only, the spectral-temporally, and the full-spectrum temporal convolutional operations. We found these types suffer from noise to a varying degree, in which noise negatively influences the S-Conv the most, followed by the T-Conv and the ST-Conv, and the FST-Conv the least. Under both conditions, the FST-Conv outperforms all of the other three types, and one of the state-of-the-art models under the clean condition. A set of extended experiments further shows that insufficient amount of spectral information is the major reason that leads to the negative influence of noise on the S-Conv. However, without temporal convolution, the S-Conv with larger filters is still not as robust to noise as the FST-Conv.  

Even though the S-Conv is the weakest type, the comparison between the S-CLDNN (log-Mels) and the LDNN (MFCCs) shows a significant performance gap between them, which can mostly be attributed to the difference between the S-Conv and the discrete cosine transformation. On the other hand, the FST-CLDNN (MFCCs) is still able to achieve a reasonably good accuracy. These two experiments suggest that although DCT may discard certain amount of affective information, the loss does not entirely account for the performance gap. However, we may link the mediocre performance of the LDNN (MFCCs) to the inability of DCT to adequately disentangle the affective information from other correlated irrelevant factors of variations such as speaker and gender differences and those caused by noise. Based on previous studies of deep neural networks, it is likely the shallow and structurally simple architecture of the DCT-DNN and its task-independent nature leads to such incapability of DCT. 

Meanwhile, we also found that the temporal convolution and the temporal recurrence are able to learn complementary information, and the combination of both results in a robust model such as the FST-CLDNN. Nevertheless, we only consider the architecture of a CNN module followed by a BLSTM module. It would be interesting to see if an architecture of a BLSTM module followed by a CNN module would make any difference.

In order to understand the internal mechanism within a CLDNN model, we quantitatively analyzed the module-wise discriminative power by training a SVM on the extracted activations from the output of modules. The reported accuracy can be viewed as an approximated measure of quality in the sense of readiness to exploit the affective information. From the results in Table \ref{tab:modulewise}, we found the CNN module, the BLSTM module and the MLP module contribute a refinement of $27.43\pm 5.18\%$, $35.63\pm 3.61\%$ and $2.85\pm 2.32\%$ to the quality, respectively. This ranking is not surprising as studies from psychology \cite{Oatley96} or computational paralinguistics \cite{Schuller03,Wollmer10, Metallinou2012, Lee15} all point out emotion is characterized by temporally dependent dynamics. Nevertheless, our findings have shown that the CNN module is capable of significantly enhancing the separability for emotional classes compared to raw features, particularly when under a noisy condition. 

In addition, we visualize three types of information along the depth of the proposed models, including the affective, speaker and gender information. From the visualization, we observe that the model is progressively learning to discriminate different emotional patterns, in which anger and sadness are two of the most prominent emotional classes across all models. What's more interesting is that other irrelevant factors of variations are integrated out at a varying rate from one module to another. Specifically, the CNN and the BLSTM modules still keep a moderate portion of the gender and speaker information but in the end the MLP module refines the learnt representations by drastically reducing other type of variations. 

\bibliographystyle{IEEEtran}
\bibliography{mybib}

\appendices
\section{Visualization of All Models}
\label{app:visual}
\begin{figure*}[hb]
\centering
  \begin{subfigure}[b]{0.33\linewidth}
    \centering
    \includegraphics[width=1\textwidth]{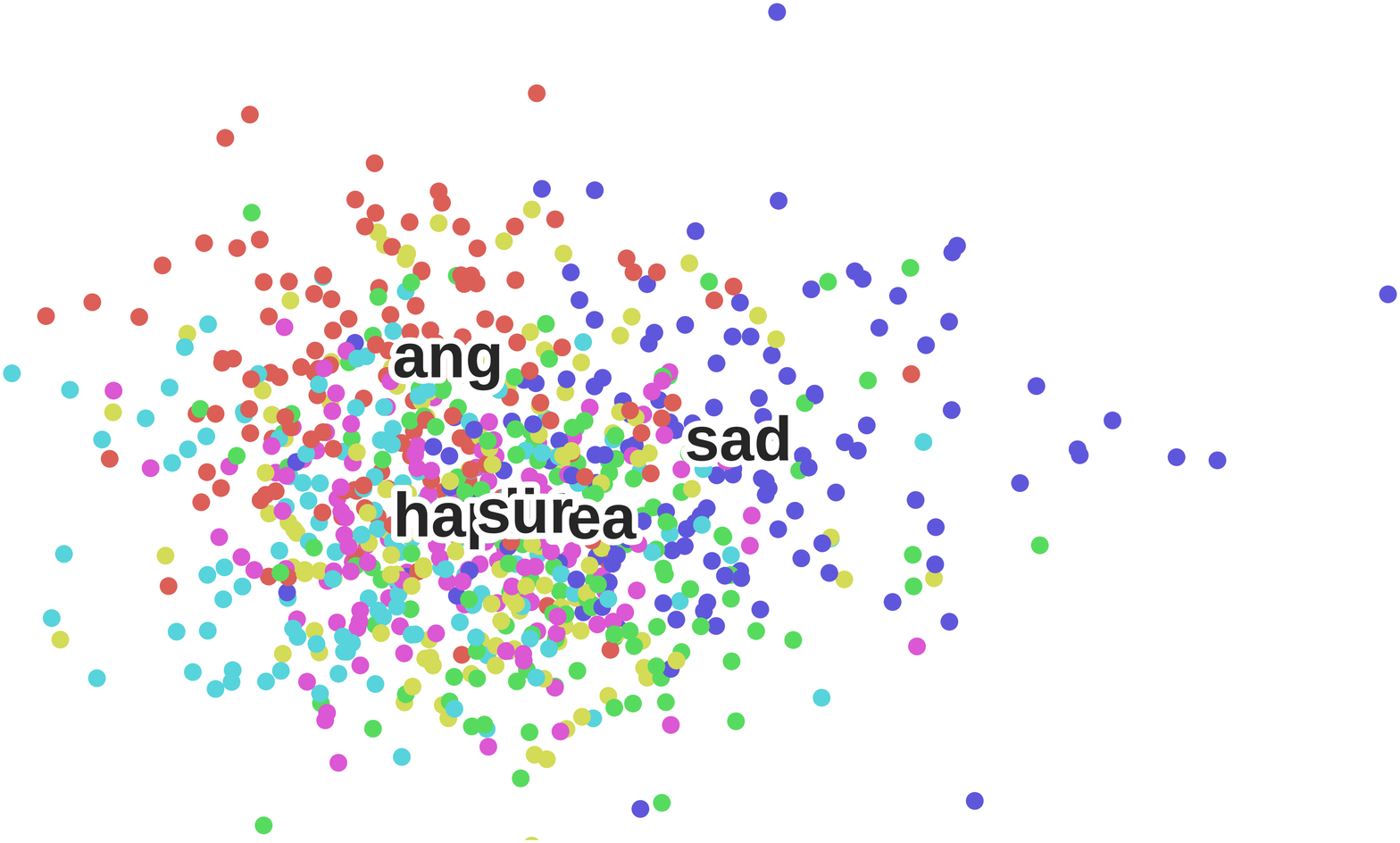}
    \caption{$2.821$}\label{fig:fcldnn_logmels_act_emo}
  \end{subfigure}%
  \begin{subfigure}[b]{.33\linewidth}
    \centering
    \includegraphics[width=1\textwidth]{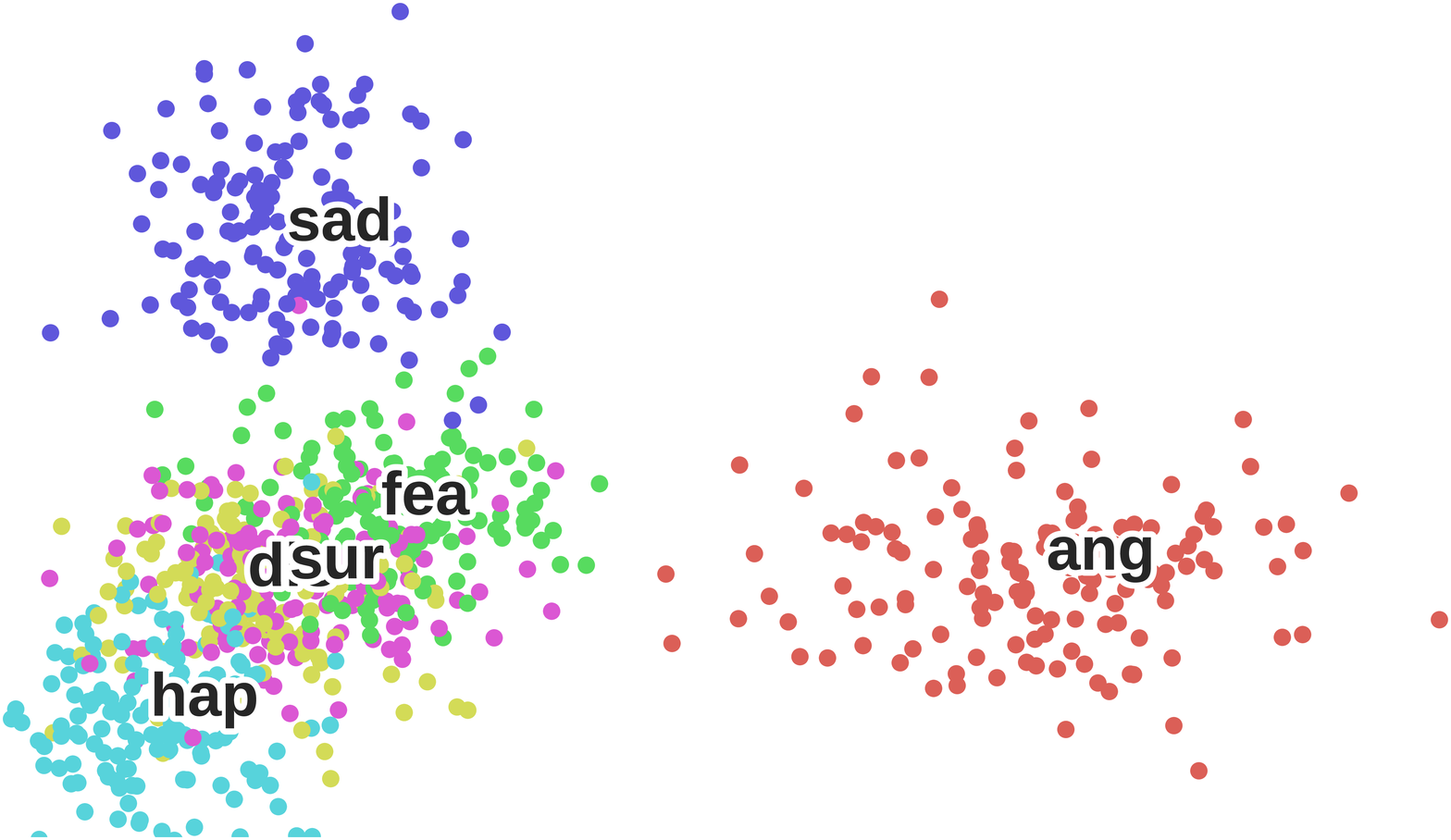}
    \caption{$1.193$}\label{fig:fcldnn_logmels_tpool_emo}
  \end{subfigure}%
  \begin{subfigure}[b]{.33\linewidth}
    \centering
    \includegraphics[width=1\textwidth]{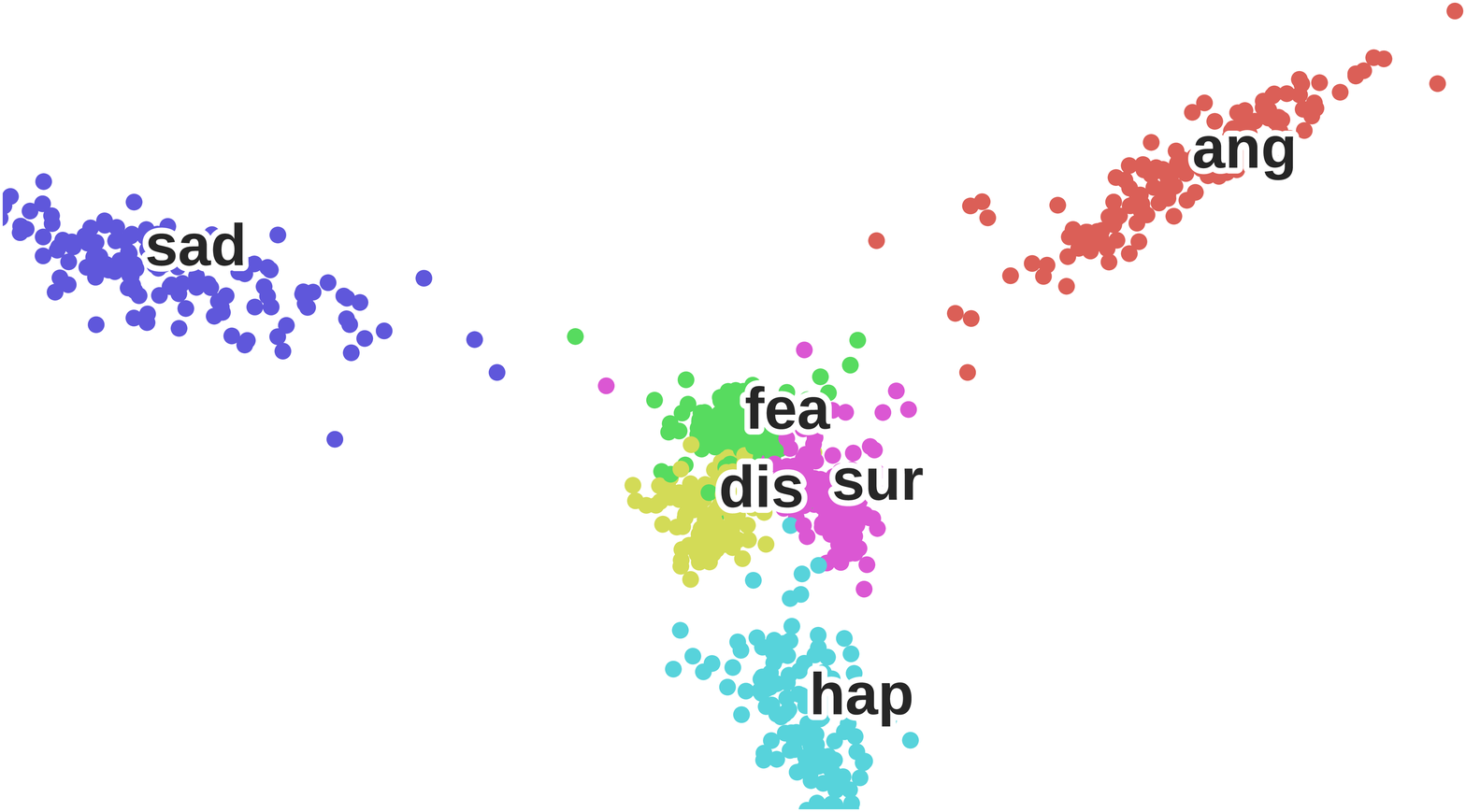}
    \caption{$0.314$}\label{fig:fcldnn_logmels_lastfc_emo}
  \end{subfigure}\\%
  
  \begin{subfigure}[b]{0.33\linewidth}
    \centering
    \includegraphics[width=1\textwidth]{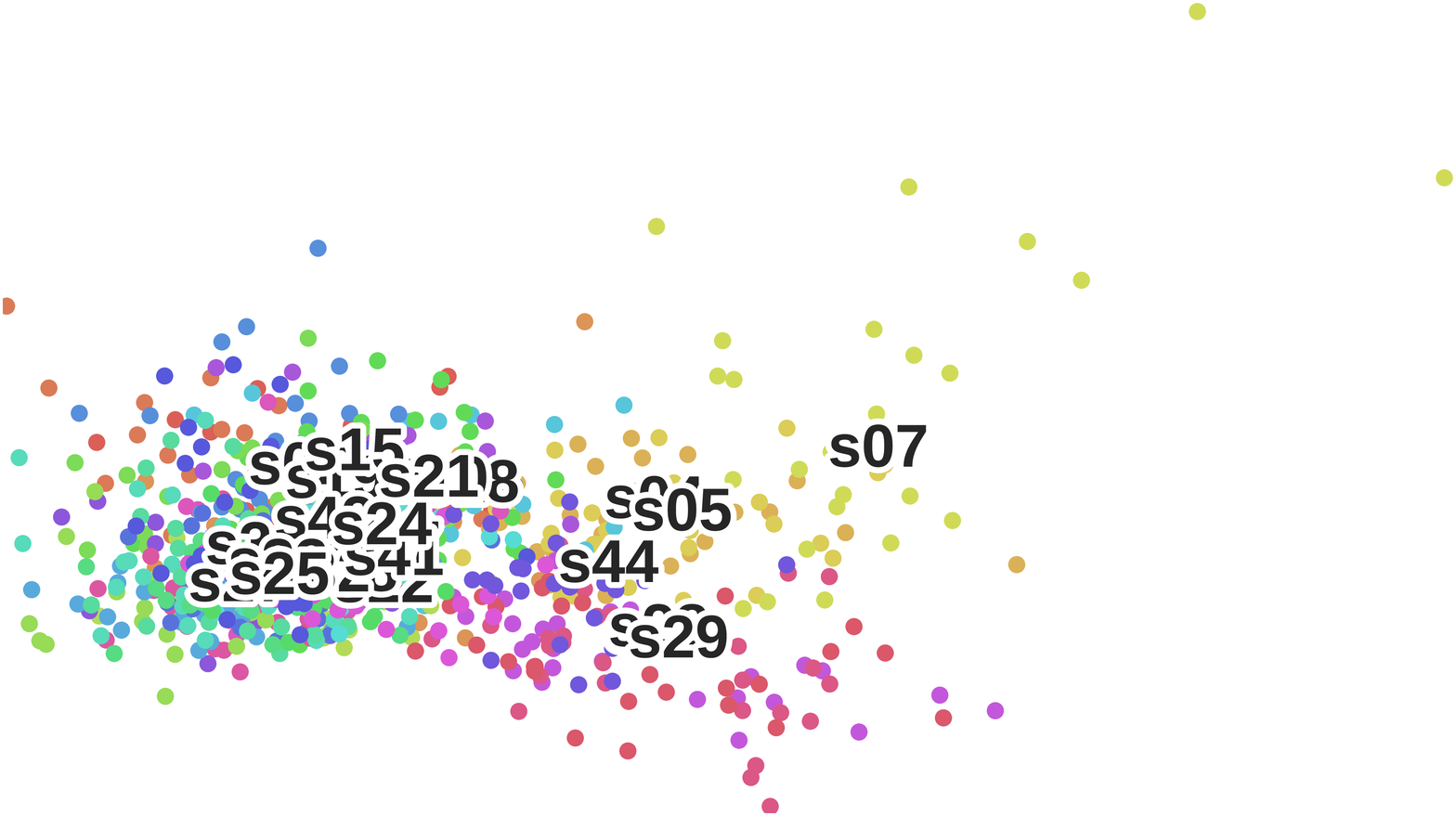}
    \caption{$1.278$}\label{fig:fcldnn_logmels_act_spkr}
  \end{subfigure}%
  \begin{subfigure}[b]{.33\linewidth}
    \centering
    \includegraphics[width=1\textwidth]{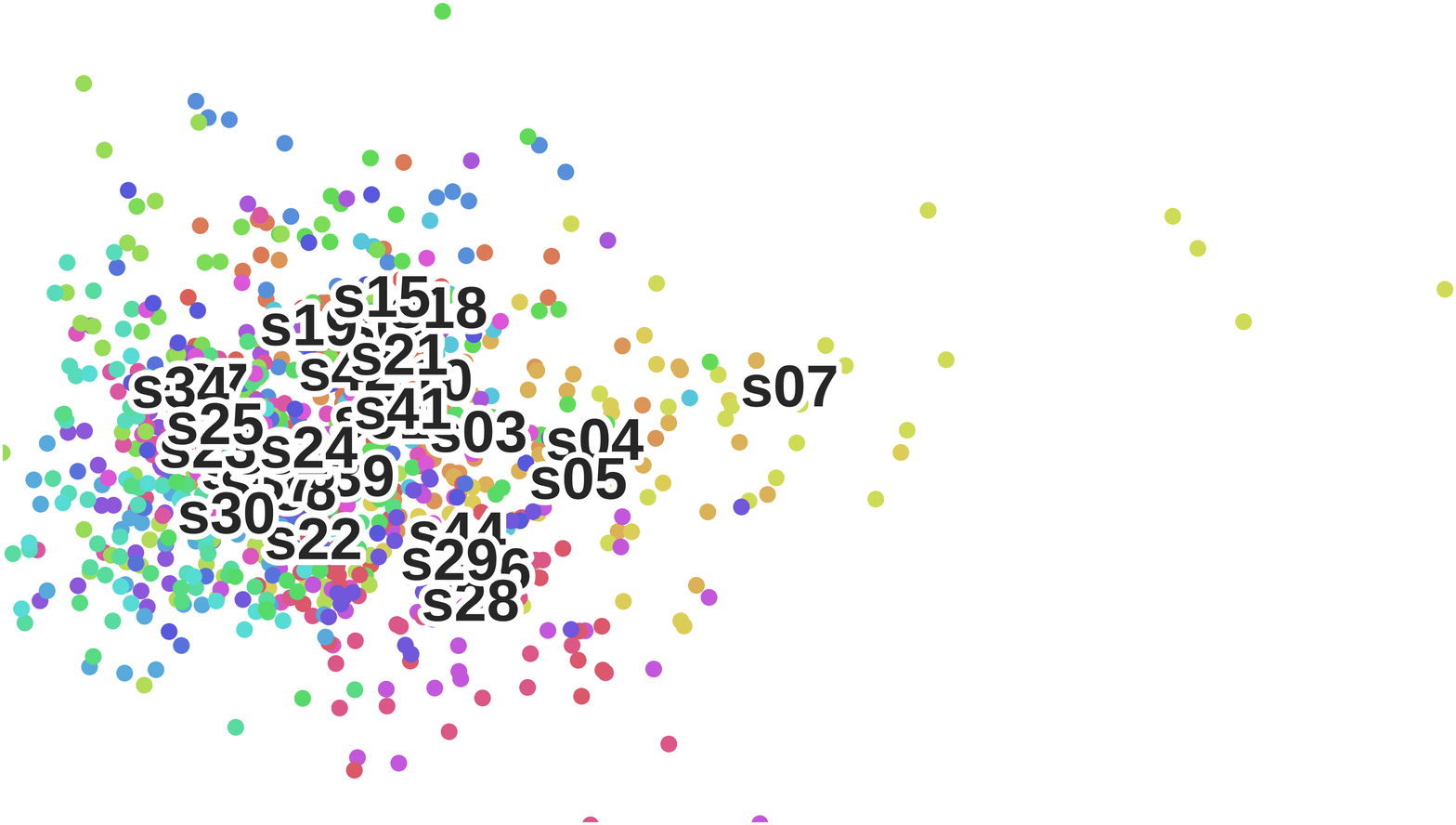}
    \caption{$2.857$}\label{fig:fcldnn_logmels_tpool_spkr}
  \end{subfigure}%
  \begin{subfigure}[b]{.33\linewidth}
    \centering
    \includegraphics[width=1\textwidth]{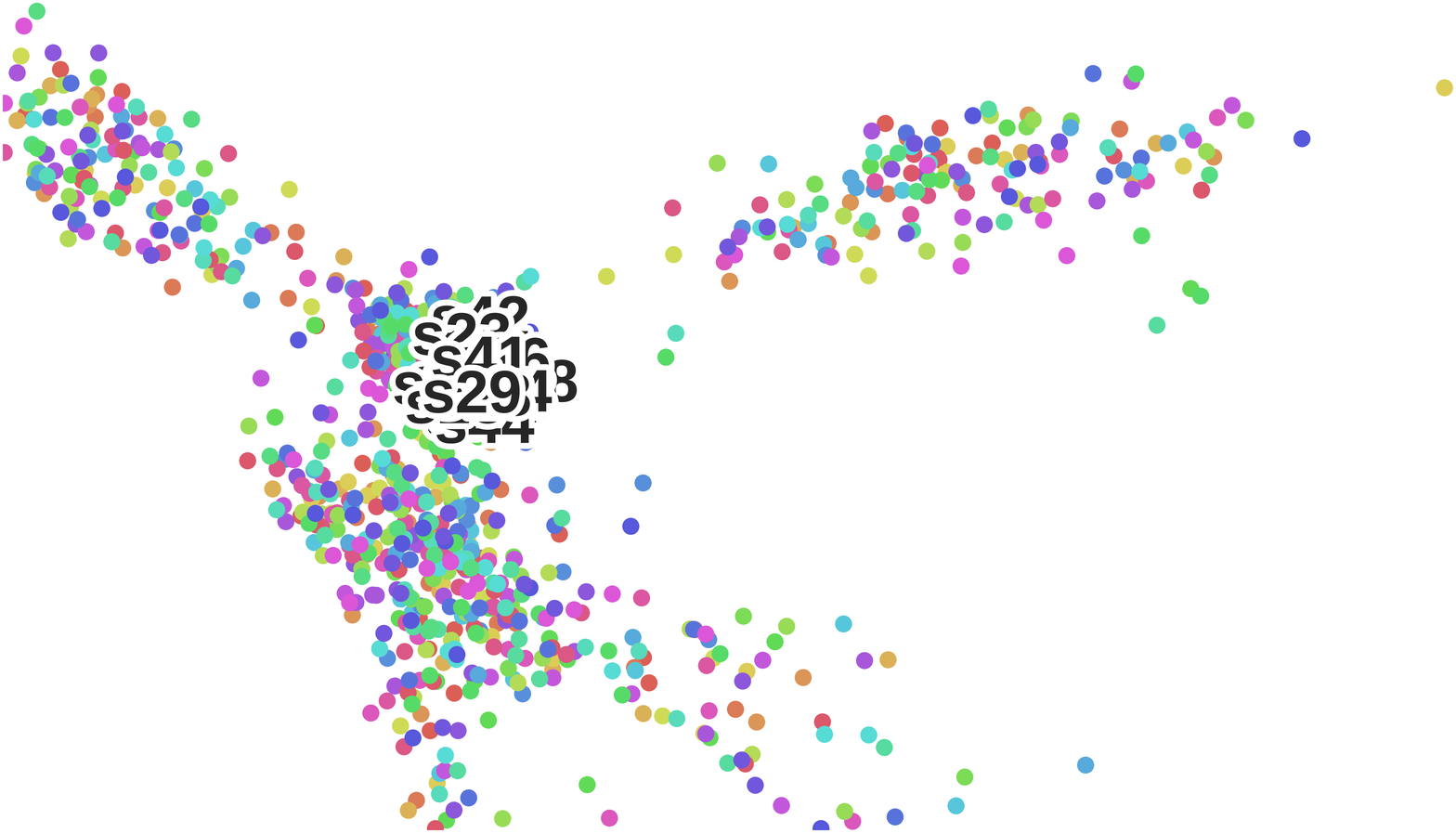}
    \caption{$7.376$}\label{fig:fcldnn_logmels_lastfc_spkr}
  \end{subfigure}\\%
  
  \begin{subfigure}[b]{0.33\linewidth}
    \centering
    \includegraphics[width=1\textwidth]{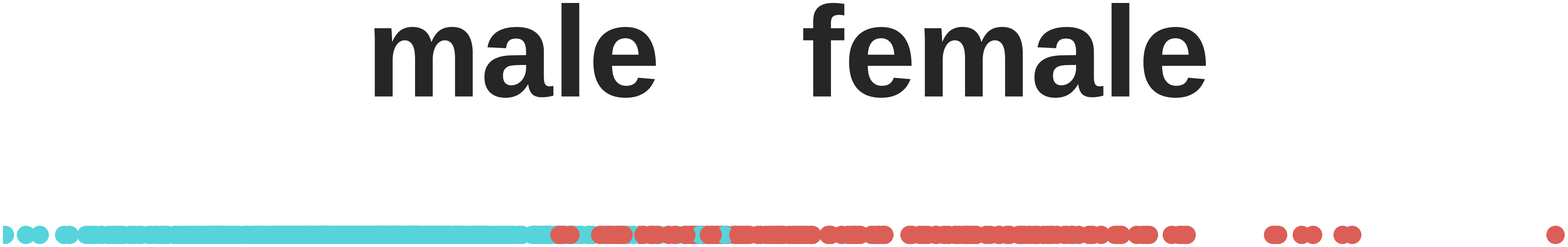}
    \caption{$1.777$}\label{fig:fcldnn_logmels_act_gender}
  \end{subfigure}%
  \begin{subfigure}[b]{.33\linewidth}
    \centering
    \includegraphics[width=1\textwidth]{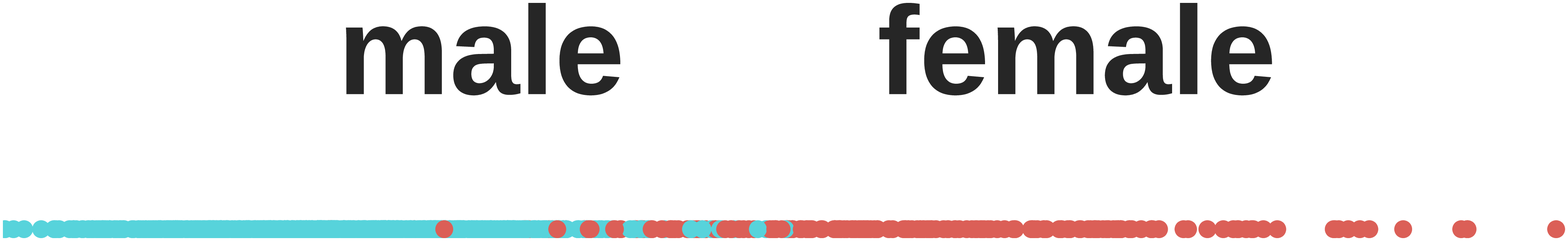}
    \caption{$4.303$}\label{fig:fcldnn_logmels_tpool_gender}
  \end{subfigure}%
  \begin{subfigure}[b]{.33\linewidth}
    \centering
    \includegraphics[width=1\textwidth]{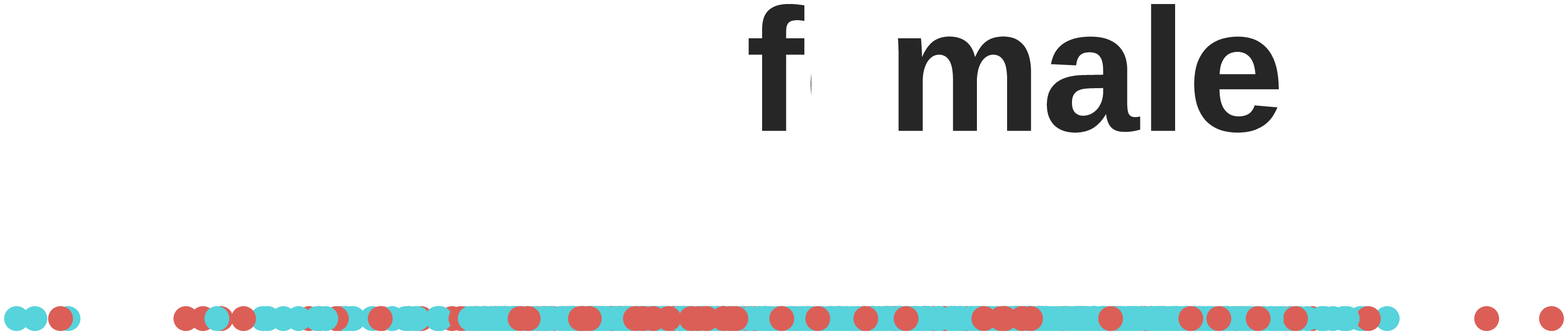}
    \caption{$19.403$}\label{fig:fcldnn_logmels_lastfc_gender}
  \end{subfigure}\\%
  \caption{The visualization for the modules in the S-CLDNN (log-Mels). The first, second and third rows correspond to the affective, speaker and gender information, while the first, second and third columns denote the output of the CNN, the BLSTM and the MLP modules, respectively. In each subplot, every dot indicates an utterance, where utterances within the same class are painted with the same color and their centers of classes are marked with according labels such as \textit{hap}, \textit{s07} and \textit{female}. The title of each subplot is the $\mathbf{\rho}$ value, i.e. the quality measure of a clustering, for the distributions in the subplot.}
  \label{fig:fcldnn_logmels}
\end{figure*}

\begin{figure*}[hb]
\centering
  \begin{subfigure}[b]{0.33\linewidth}
    \centering
    \includegraphics[width=1\textwidth]{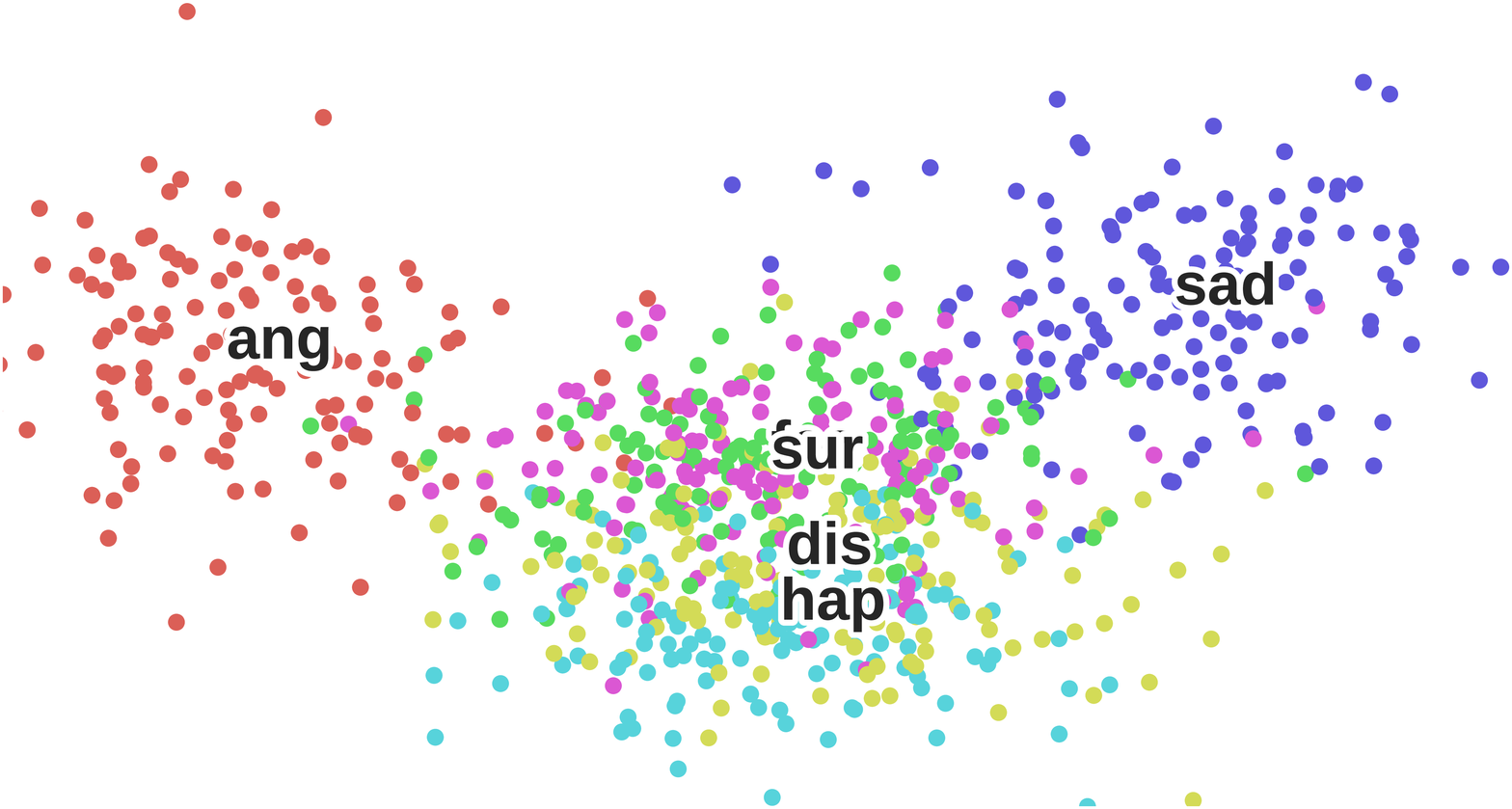}
    \caption{$2.506$}
  \end{subfigure}%
  \begin{subfigure}[b]{.33\linewidth}
    \centering
    \includegraphics[width=1\textwidth]{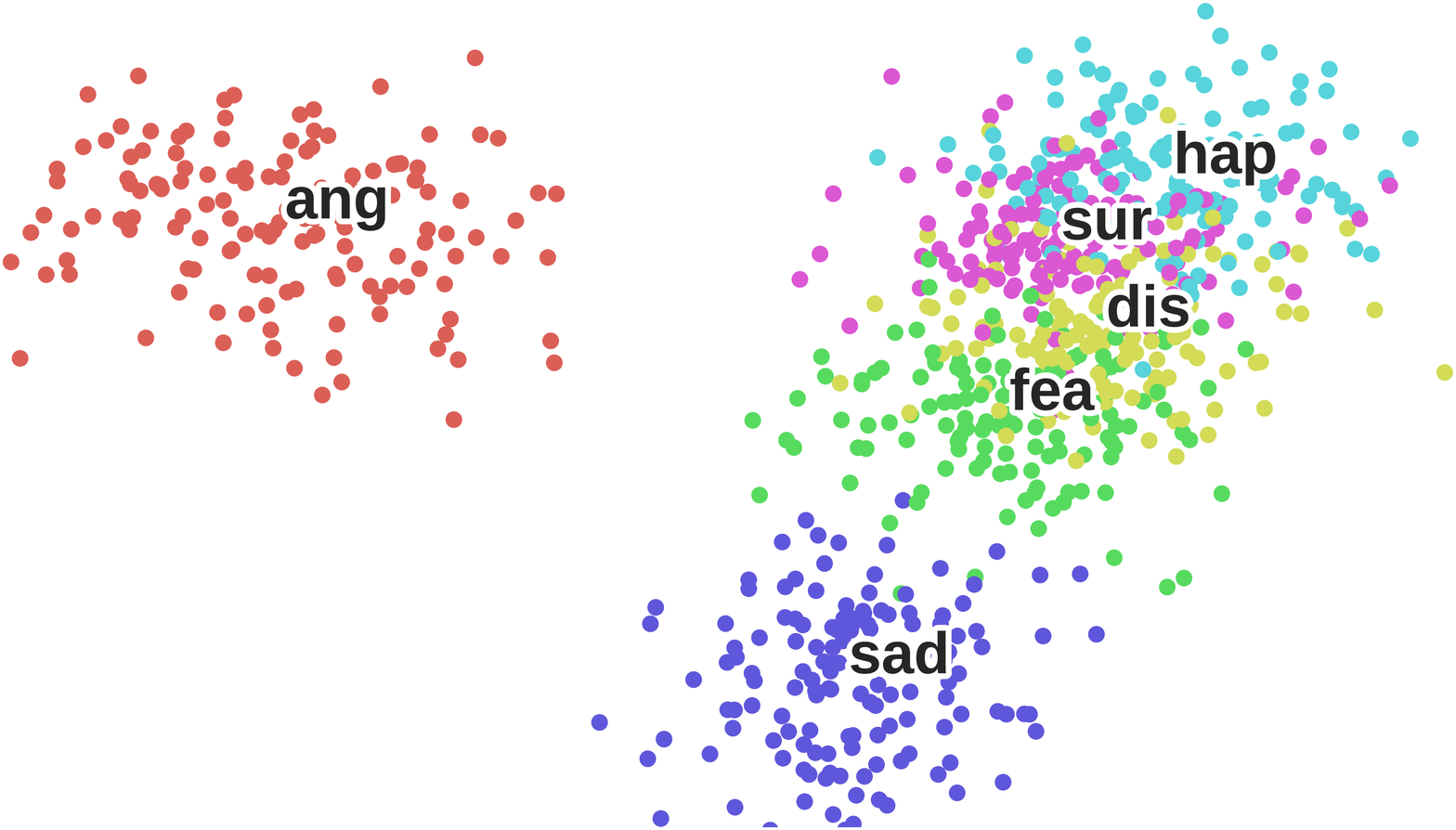}
    \caption{$1.148$}
  \end{subfigure}%
  \begin{subfigure}[b]{.33\linewidth}
    \centering
    \includegraphics[width=1\textwidth]{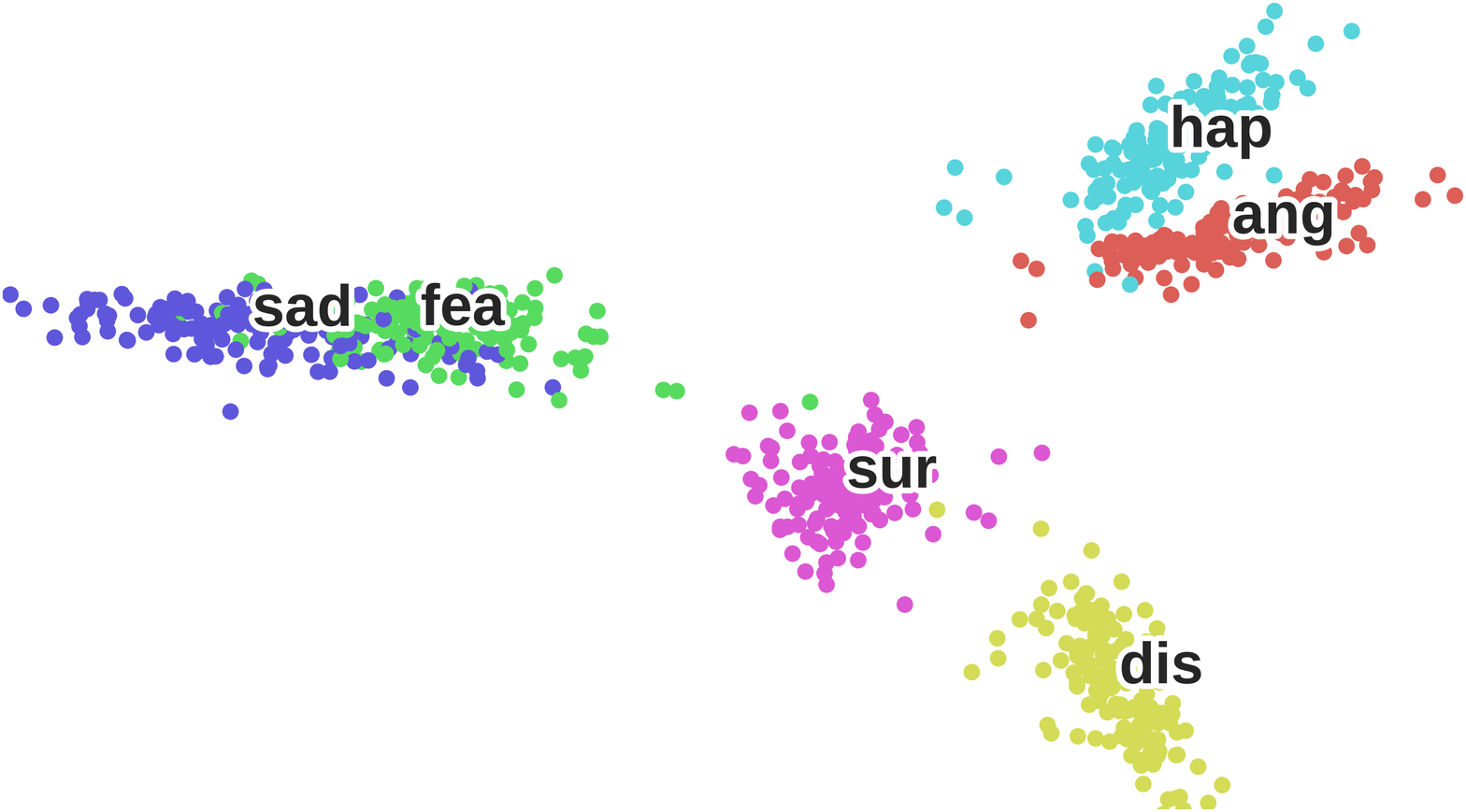}
    \caption{$0.263$}
  \end{subfigure}\\%
  
  \begin{subfigure}[b]{0.33\linewidth}
    \centering
    \includegraphics[width=1\textwidth]{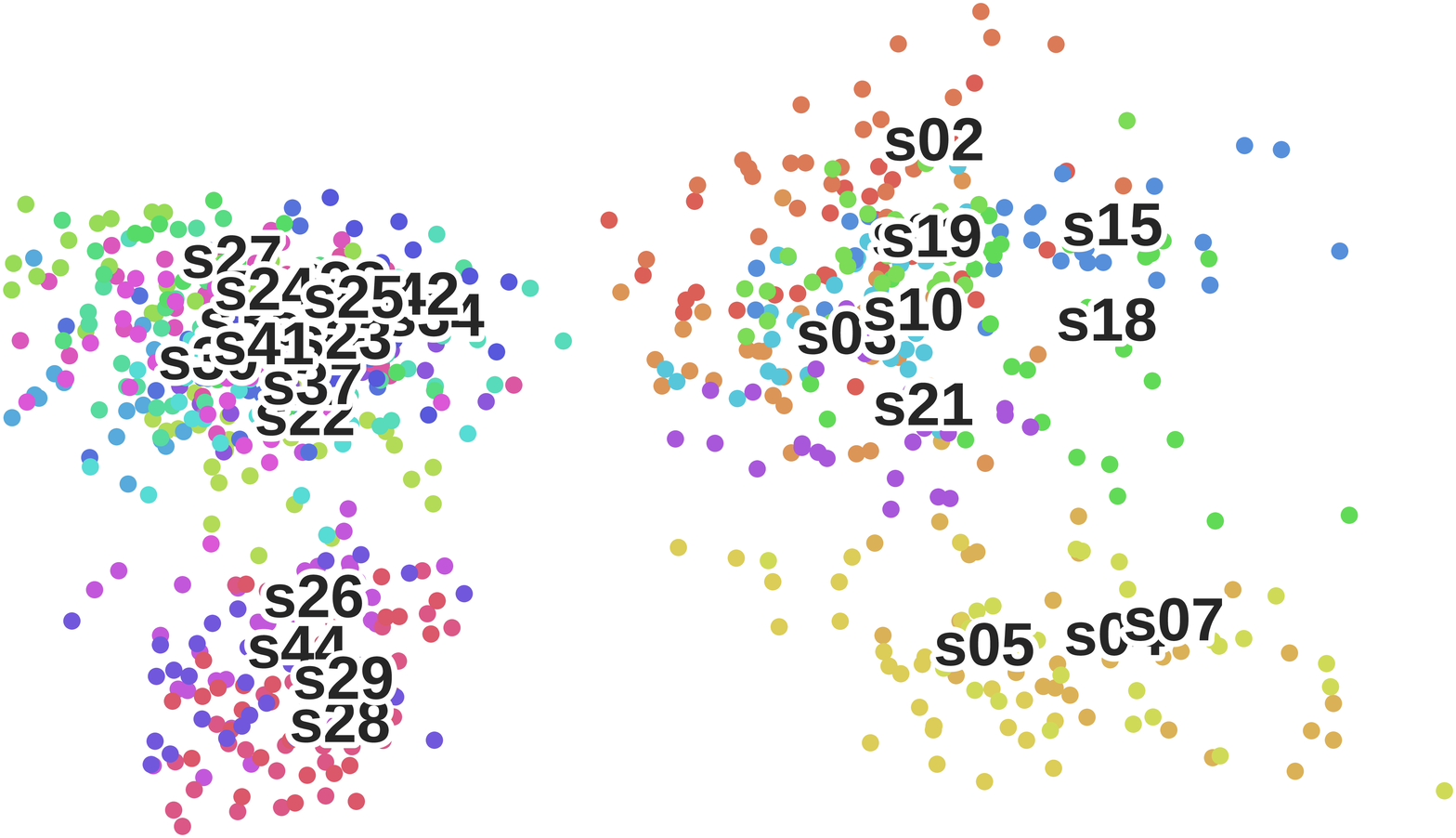}
    \caption{$1.298$}
  \end{subfigure}%
  \begin{subfigure}[b]{.33\linewidth}
    \centering
    \includegraphics[width=1\textwidth]{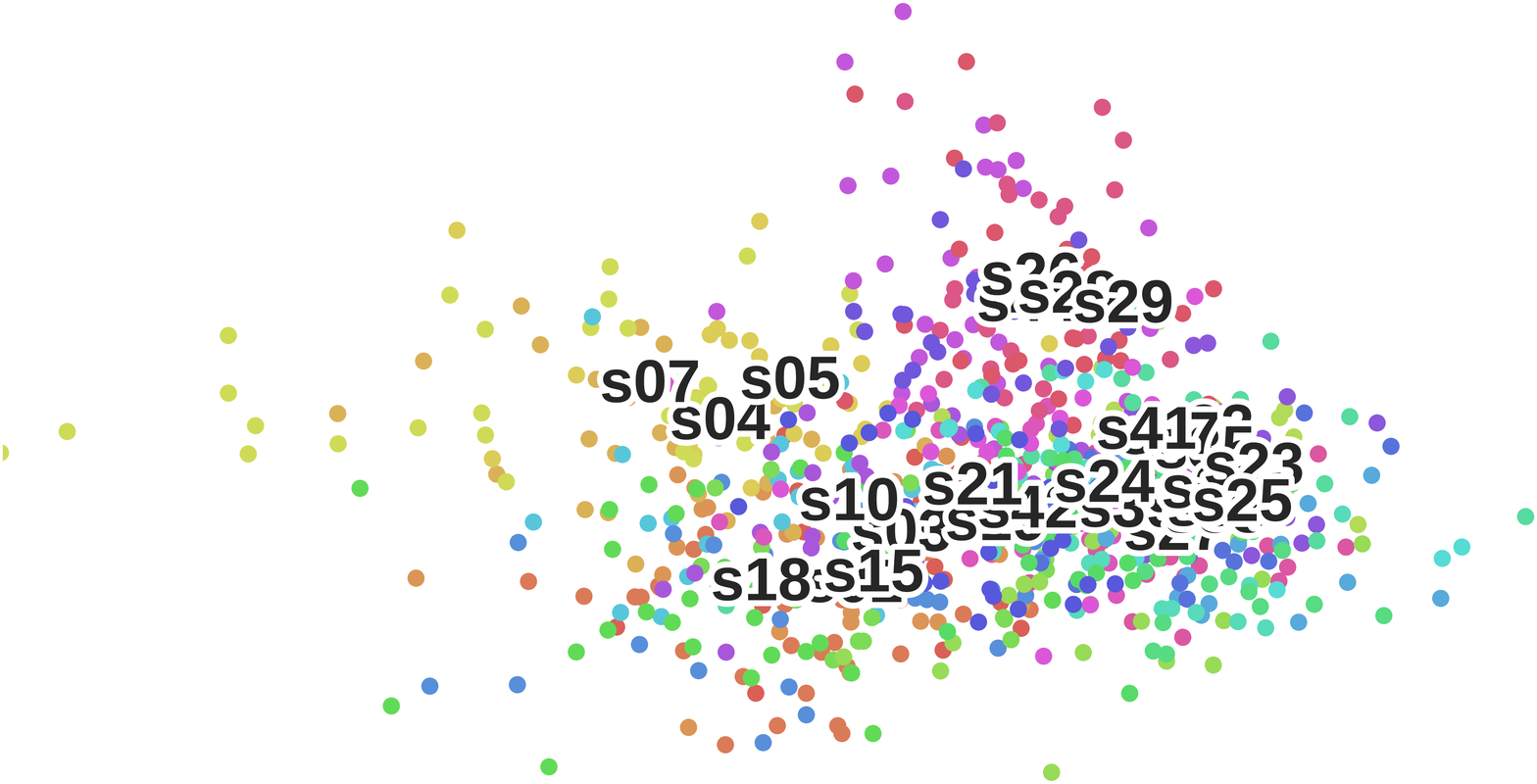}
    \caption{$2.800$}
  \end{subfigure}%
  \begin{subfigure}[b]{.33\linewidth}
    \centering
    \includegraphics[width=1\textwidth]{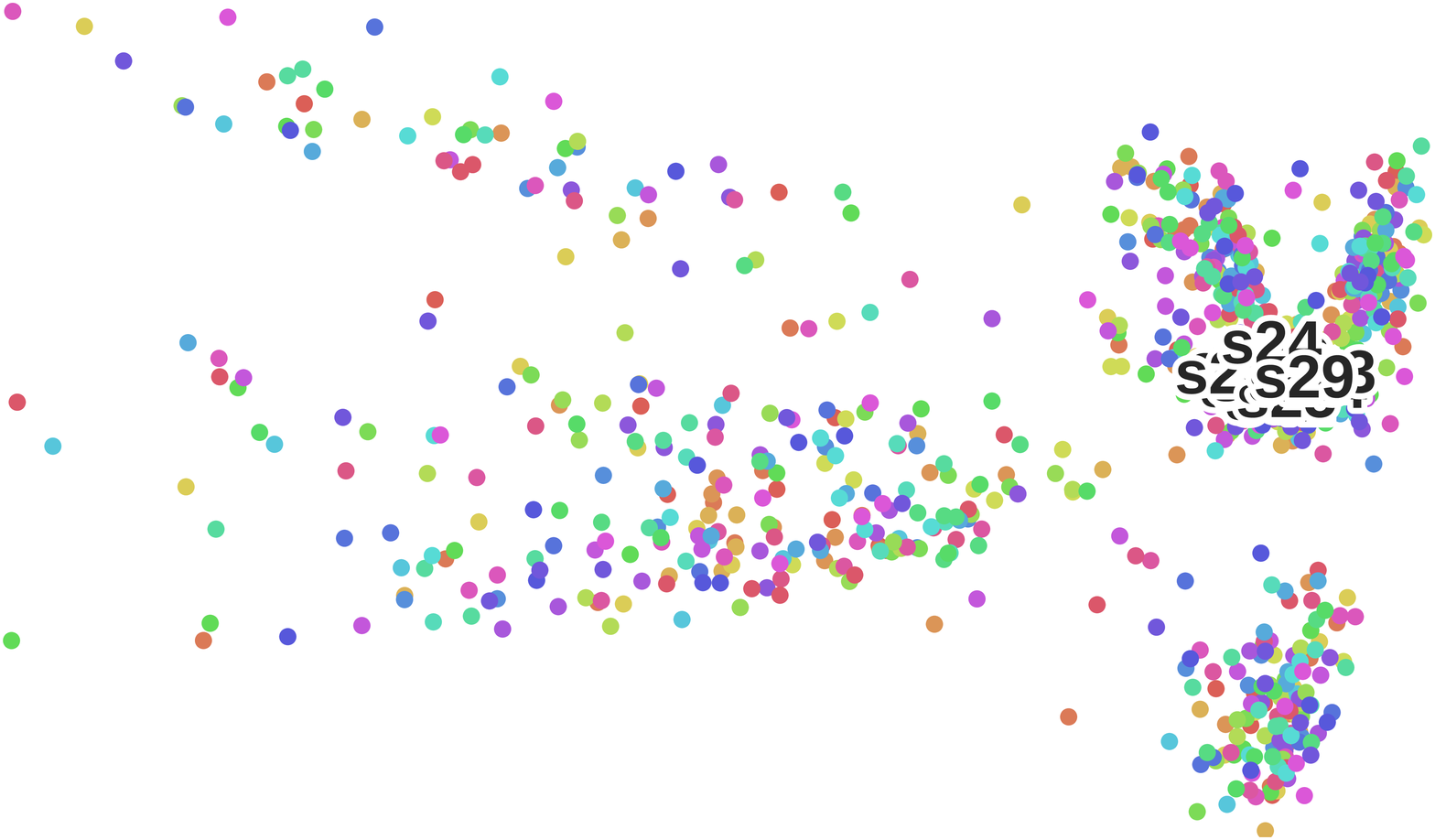}
    \caption{$7.762$}
  \end{subfigure}\\%
  
  \begin{subfigure}[b]{0.33\linewidth}
    \centering
    \includegraphics[width=1\textwidth]{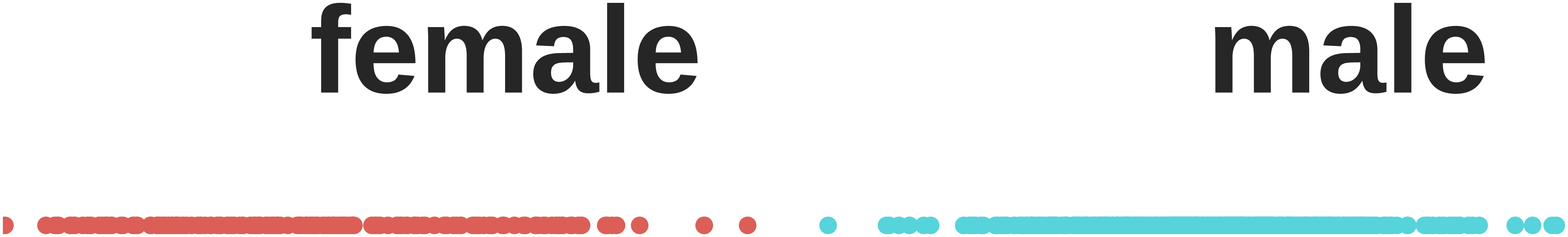}
    \caption{$2.988$}
  \end{subfigure}%
  \begin{subfigure}[b]{.33\linewidth}
    \centering
    \includegraphics[width=1\textwidth]{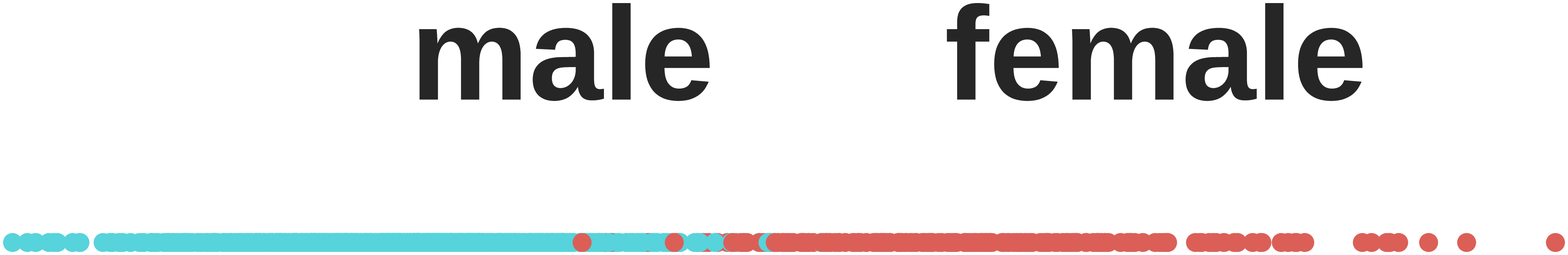}
    \caption{$5.292$}
  \end{subfigure}%
  \begin{subfigure}[b]{.33\linewidth}
    \centering
    \includegraphics[width=1\textwidth]{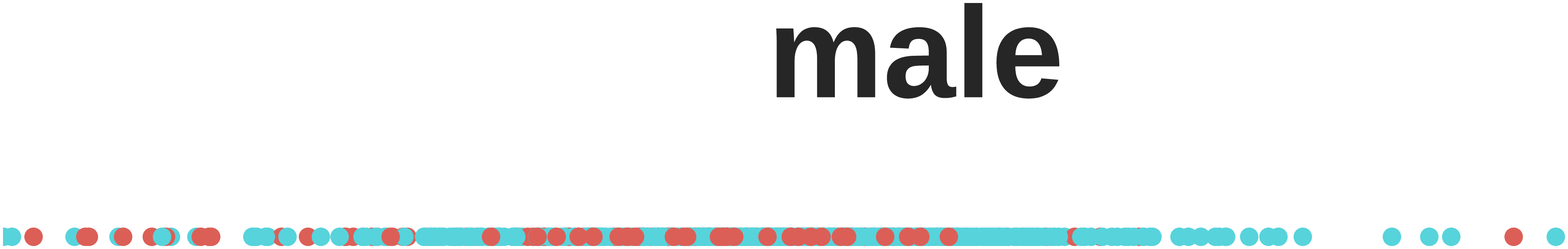}
    \caption{$17.292$}
  \end{subfigure}\\%
  \caption{The visualization for the modules in the T-CLDNN (log-Mels). The first, second and third rows correspond to the affective, speaker and gender information, while the first, second and third columns denote the output of the CNN, the BLSTM and the MLP modules, respectively. In each subplot, every dot indicates an utterance, where utterances within the same class are painted with the same color and their centers of classes are marked with according labels such as \textit{hap}, \textit{s07} and \textit{female}. The title of each subplot is the $\mathbf{\rho}$ value, i.e. the quality measure of a clustering, for the distributions in the subplot.}
\end{figure*}

\begin{figure*}[hb]
\centering
  \begin{subfigure}[b]{0.33\linewidth}
    \centering
    \includegraphics[width=1\textwidth]{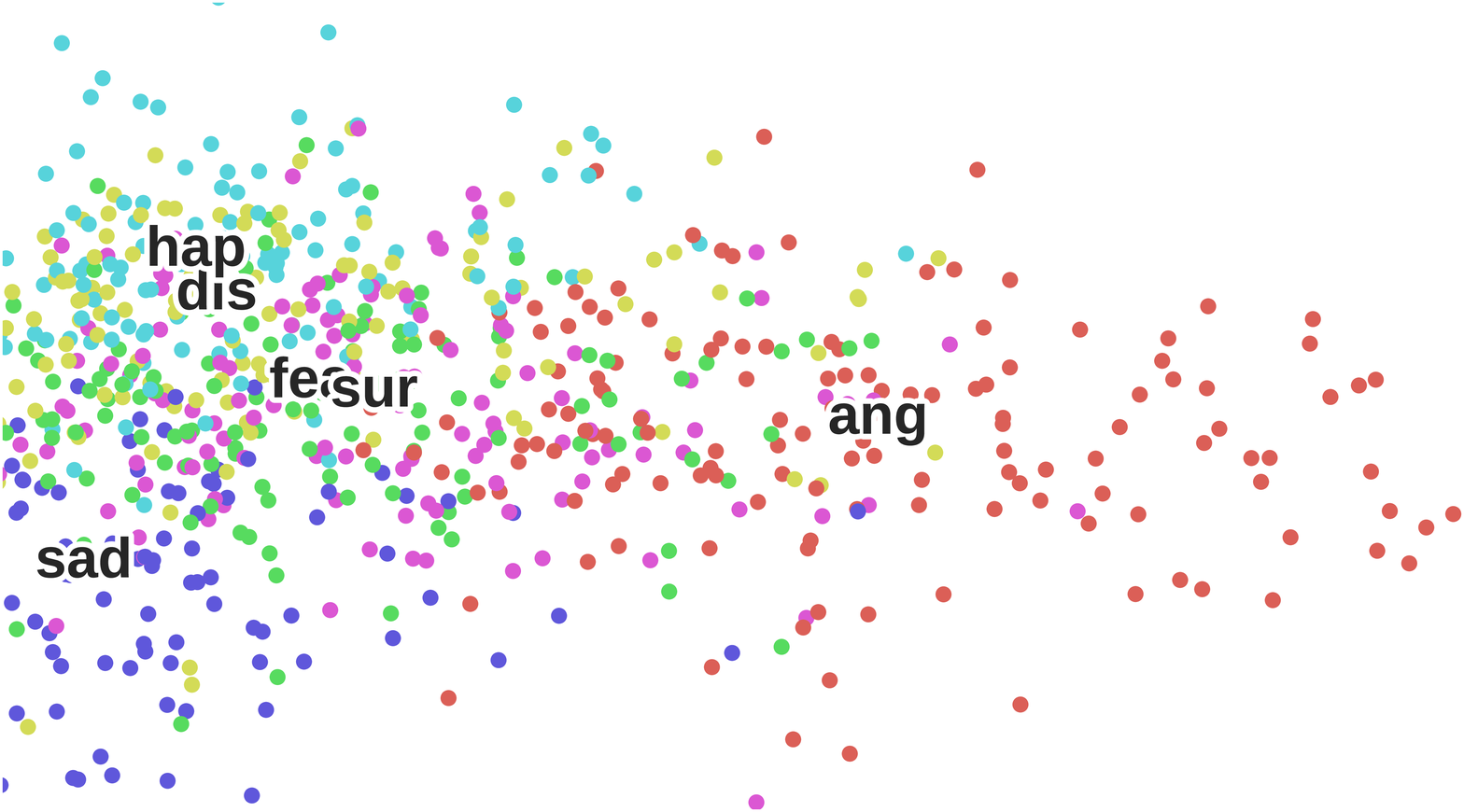}
    \caption{$2.655$}\label{fig:ftcldnn_logmels_act_emo}
  \end{subfigure}%
  \begin{subfigure}[b]{.33\linewidth}
    \centering
    \includegraphics[width=1\textwidth]{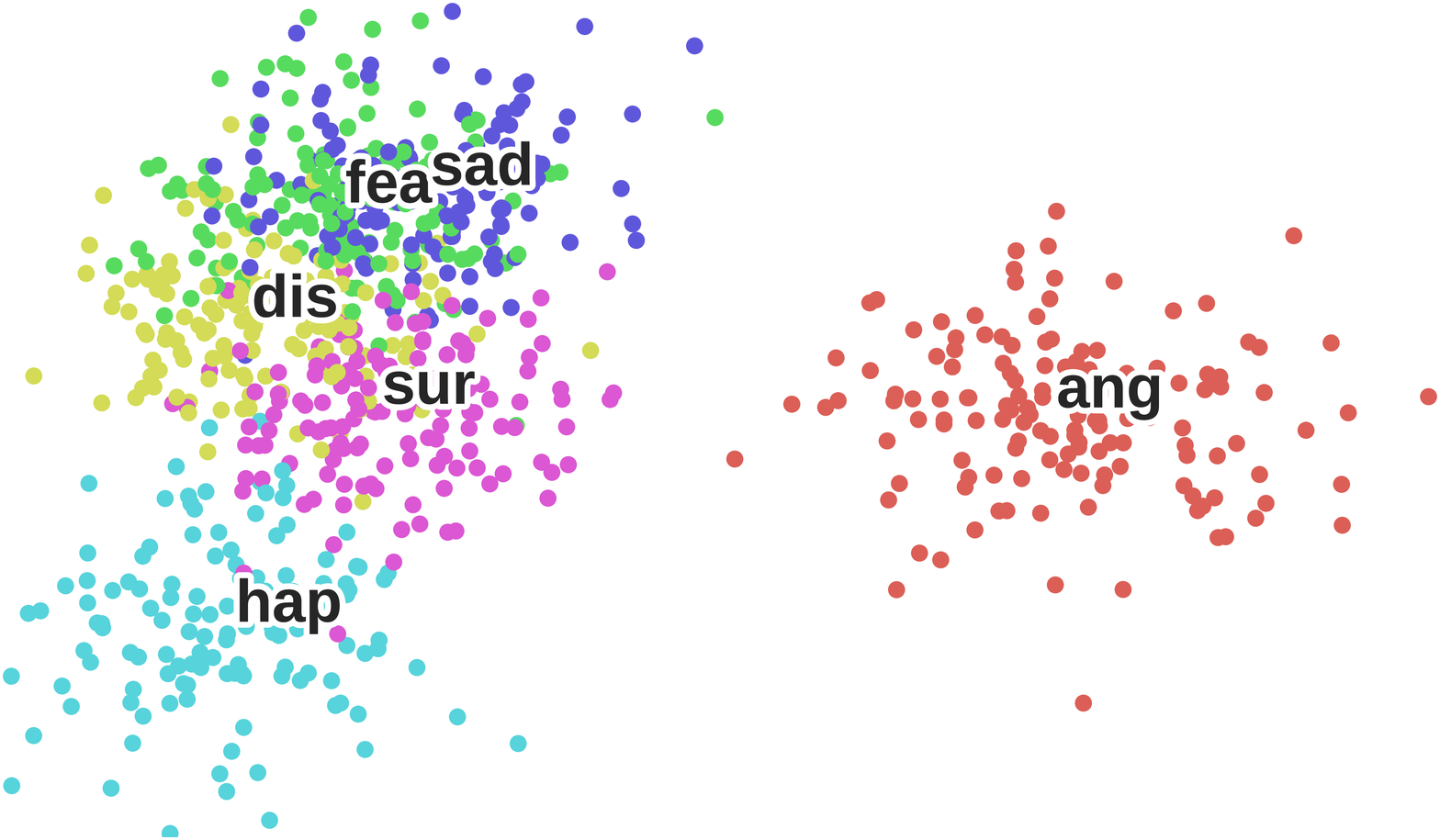}
    \caption{$1.329$}\label{fig:ftcldnn_logmels_tpool_emo}
  \end{subfigure}%
  \begin{subfigure}[b]{.33\linewidth}
    \centering
    \includegraphics[width=1\textwidth]{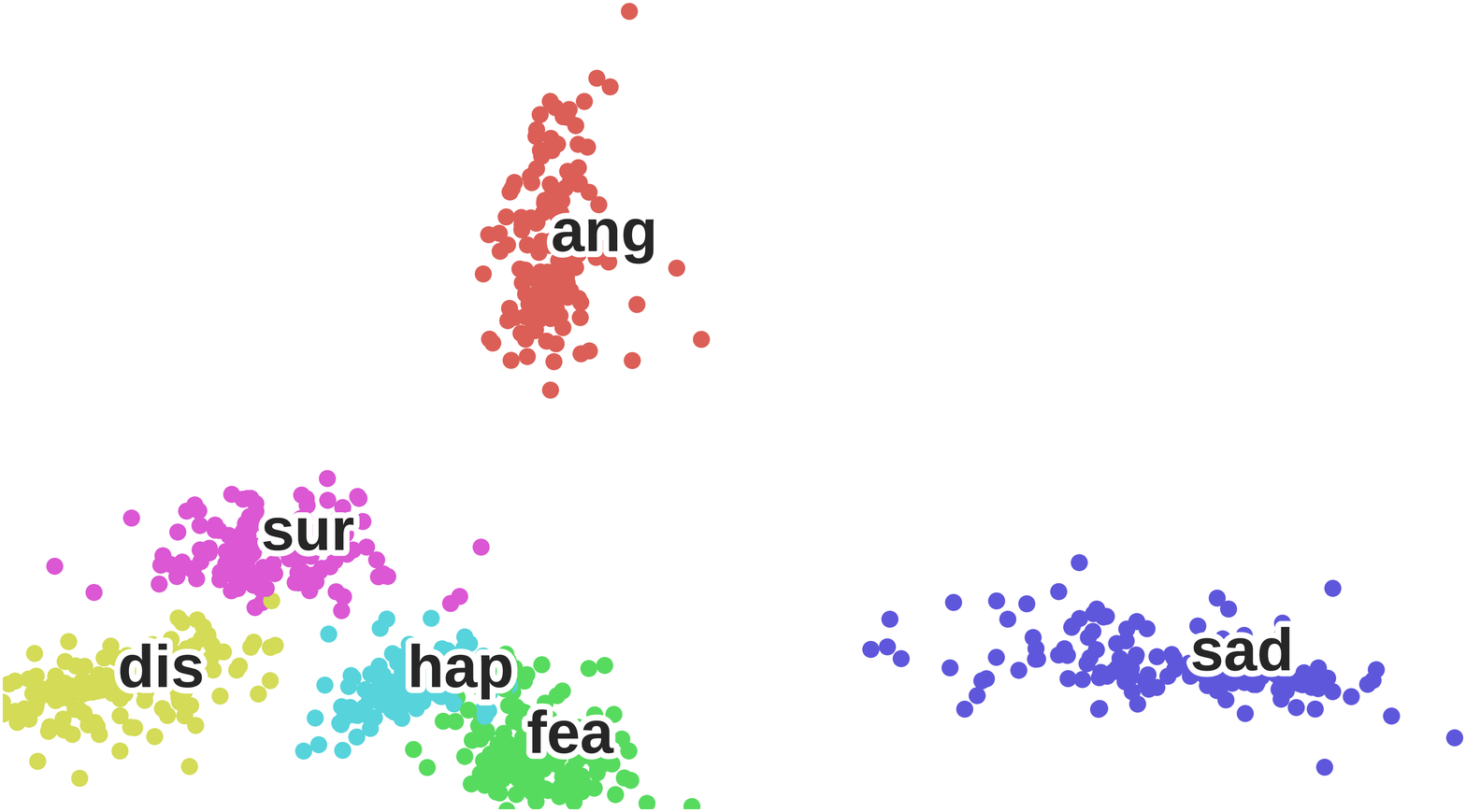}
    \caption{$0.317$}\label{fig:ftcldnn_logmels_lastfc_emo}
  \end{subfigure}\\%
  
  \begin{subfigure}[b]{0.33\linewidth}
    \centering
    \includegraphics[width=1\textwidth]{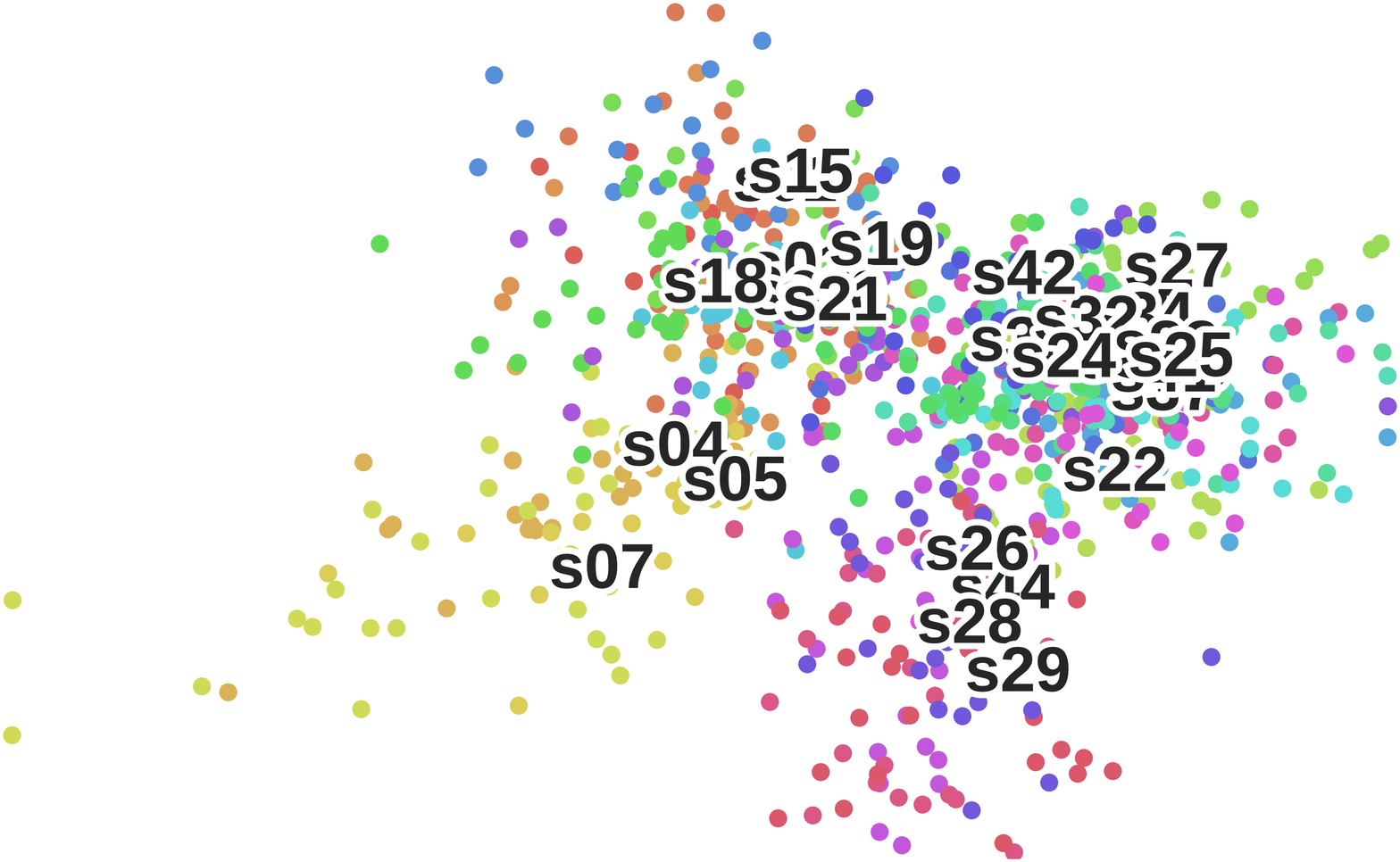}
    \caption{$1.298$}\label{fig:ftcldnn_logmels_act_spkr}
  \end{subfigure}%
  \begin{subfigure}[b]{.33\linewidth}
    \centering
    \includegraphics[width=1\textwidth]{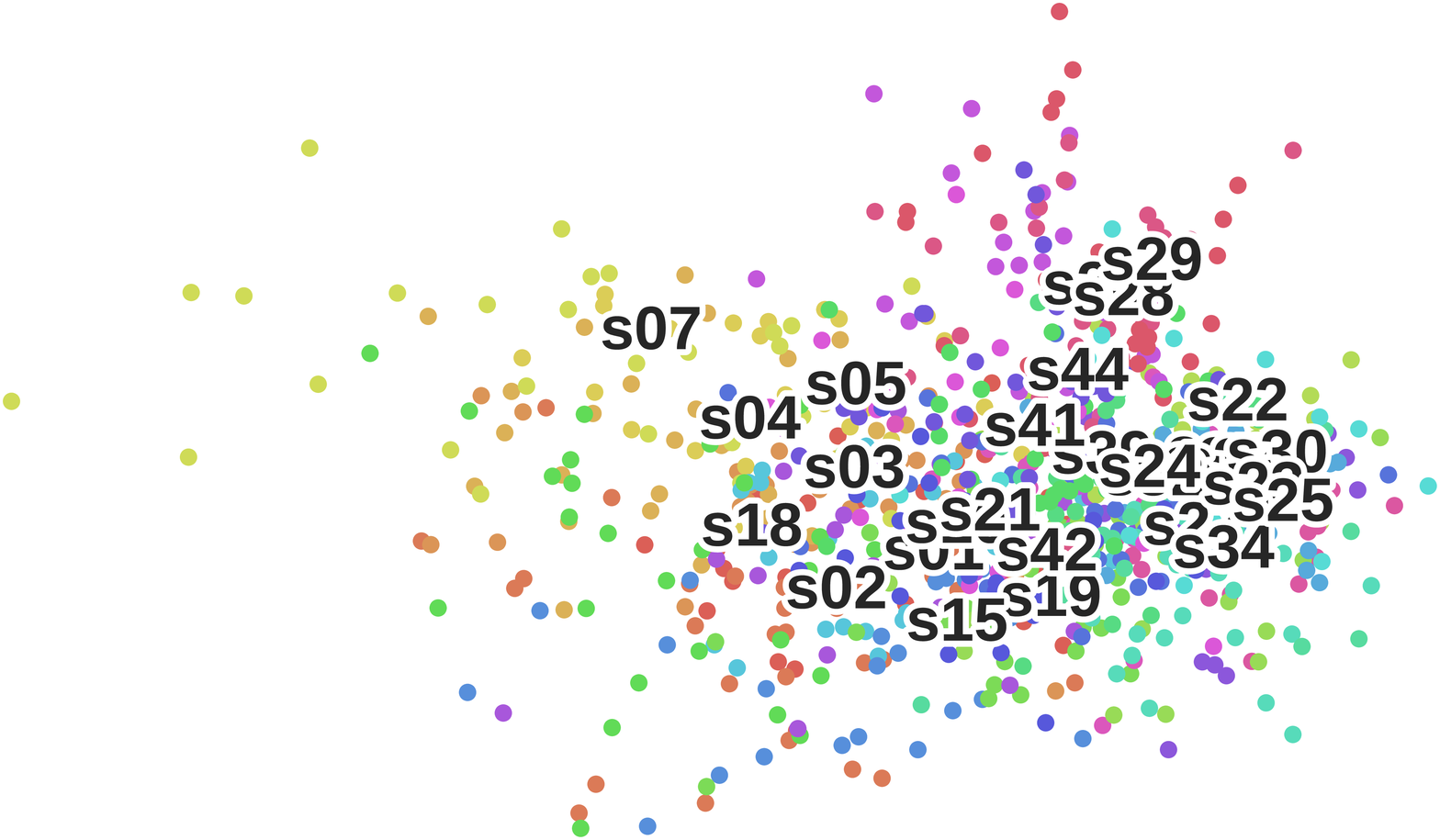}
    \caption{$2.615$}\label{fig:ftcldnn_logmels_tpool_spkr}
  \end{subfigure}%
  \begin{subfigure}[b]{.33\linewidth}
    \centering
    \includegraphics[width=1\textwidth]{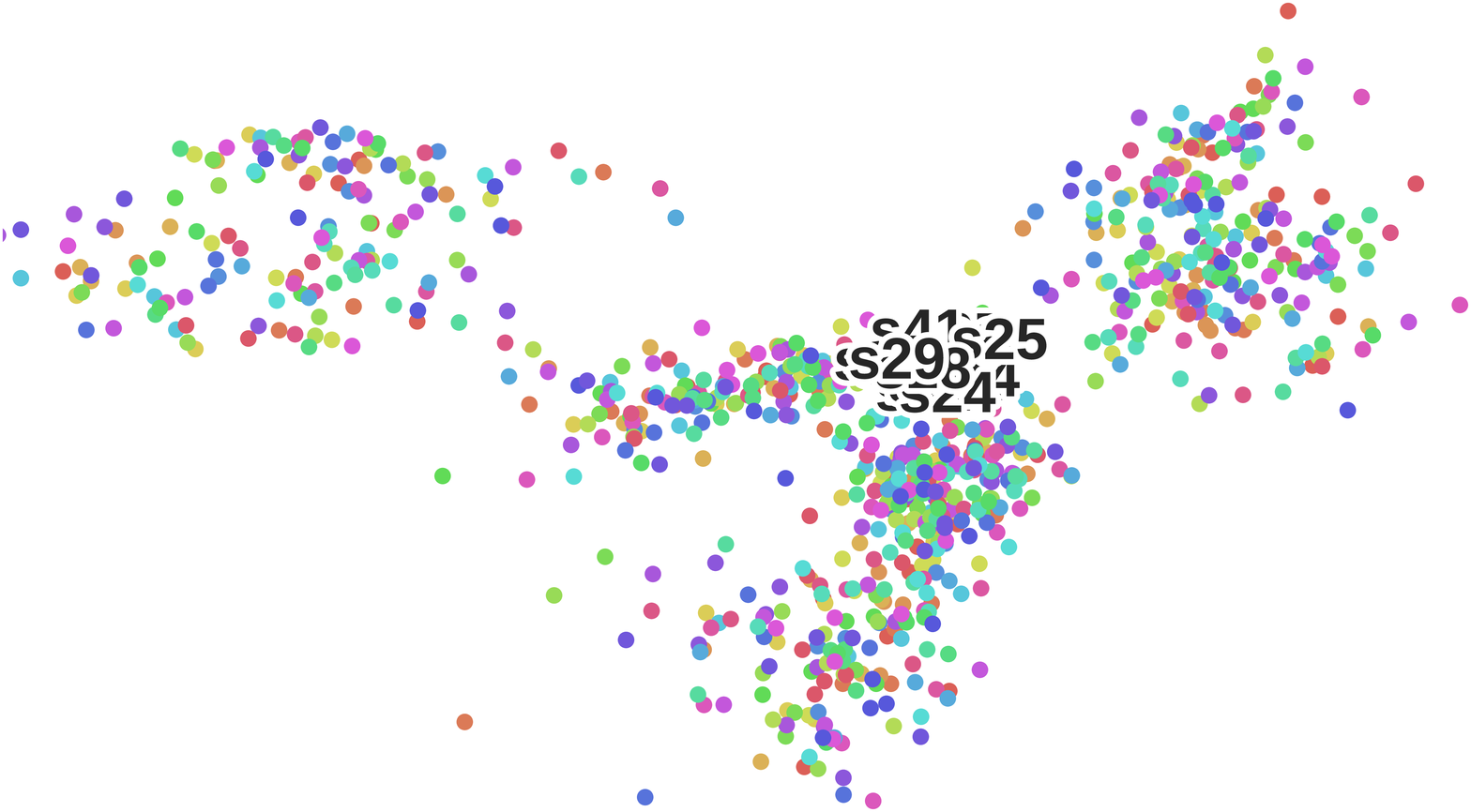}
    \caption{$7.747$}\label{fig:ftcldnn_logmels_lastfc_spkr}
  \end{subfigure}\\%
  
  \begin{subfigure}[b]{0.33\linewidth}
    \centering
    \includegraphics[width=1\textwidth]{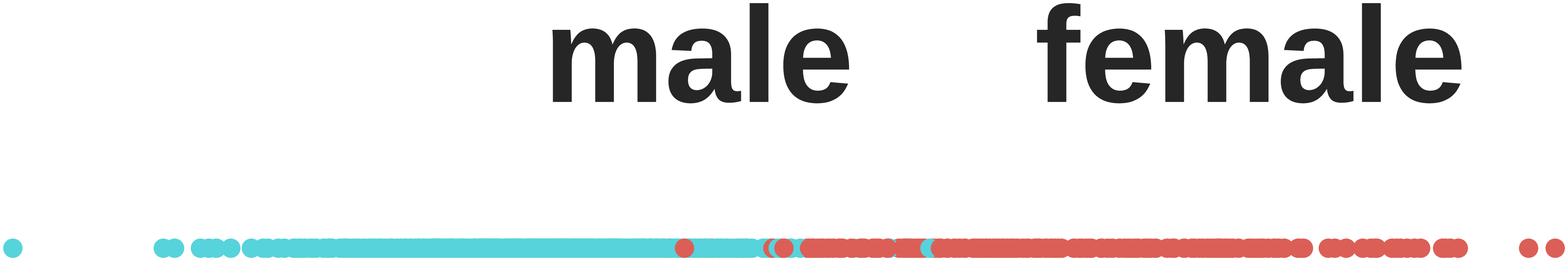}
    \caption{$2.015$}\label{fig:ftcldnn_logmels_act_gender}
  \end{subfigure}%
  \begin{subfigure}[b]{.33\linewidth}
    \centering
    \includegraphics[width=1\textwidth]{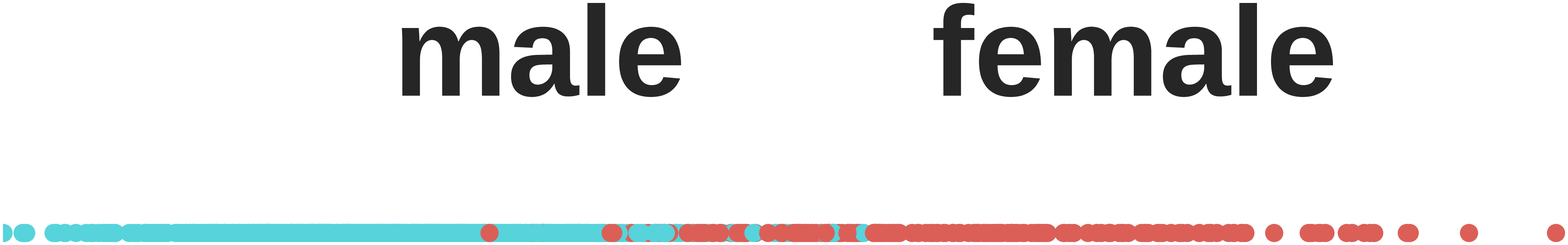}
    \caption{$4.821$}\label{fig:ftcldnn_logmels_tpool_gender}
  \end{subfigure}%
  \begin{subfigure}[b]{.33\linewidth}
    \centering
    \includegraphics[width=1\textwidth]{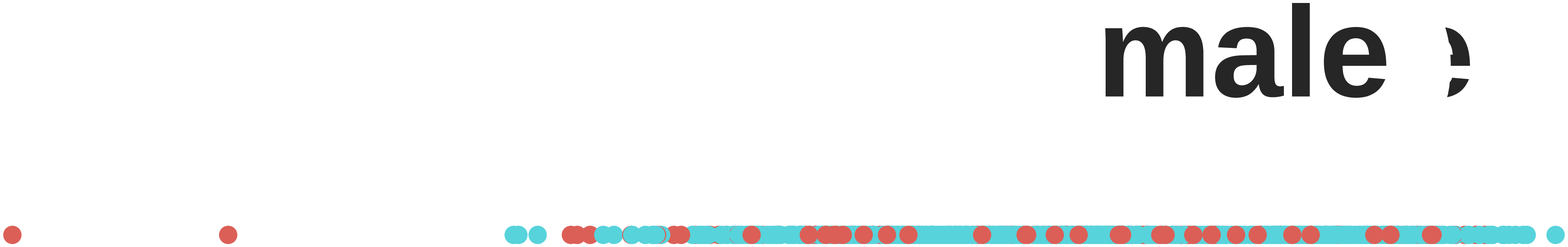}
    \caption{$23.587$}\label{fig:ftcldnn_logmels_lastfc_gender}
  \end{subfigure}\\%
  \caption{The visualization for the modules in the ST-CLDNN (log-Mels). The first, second and third rows correspond to the affective, speaker and gender information, while the first, second and third columns denote the output of the CNN, the BLSTM and the MLP modules, respectively. In each subplot, every dot indicates an utterance, where utterances within the same class are painted with the same color and their centers of classes are marked with according labels such as \textit{hap}, \textit{s07} and \textit{female}. The title of each subplot is the $\mathbf{\rho}$ value, i.e. the quality measure of a clustering, for the distributions in the subplot.}
  \label{fig:ftcldnn_logmels}
\end{figure*}

\begin{figure*}[hb]
\centering
  \begin{subfigure}[b]{0.33\linewidth}
    \centering
    \includegraphics[width=1\textwidth]{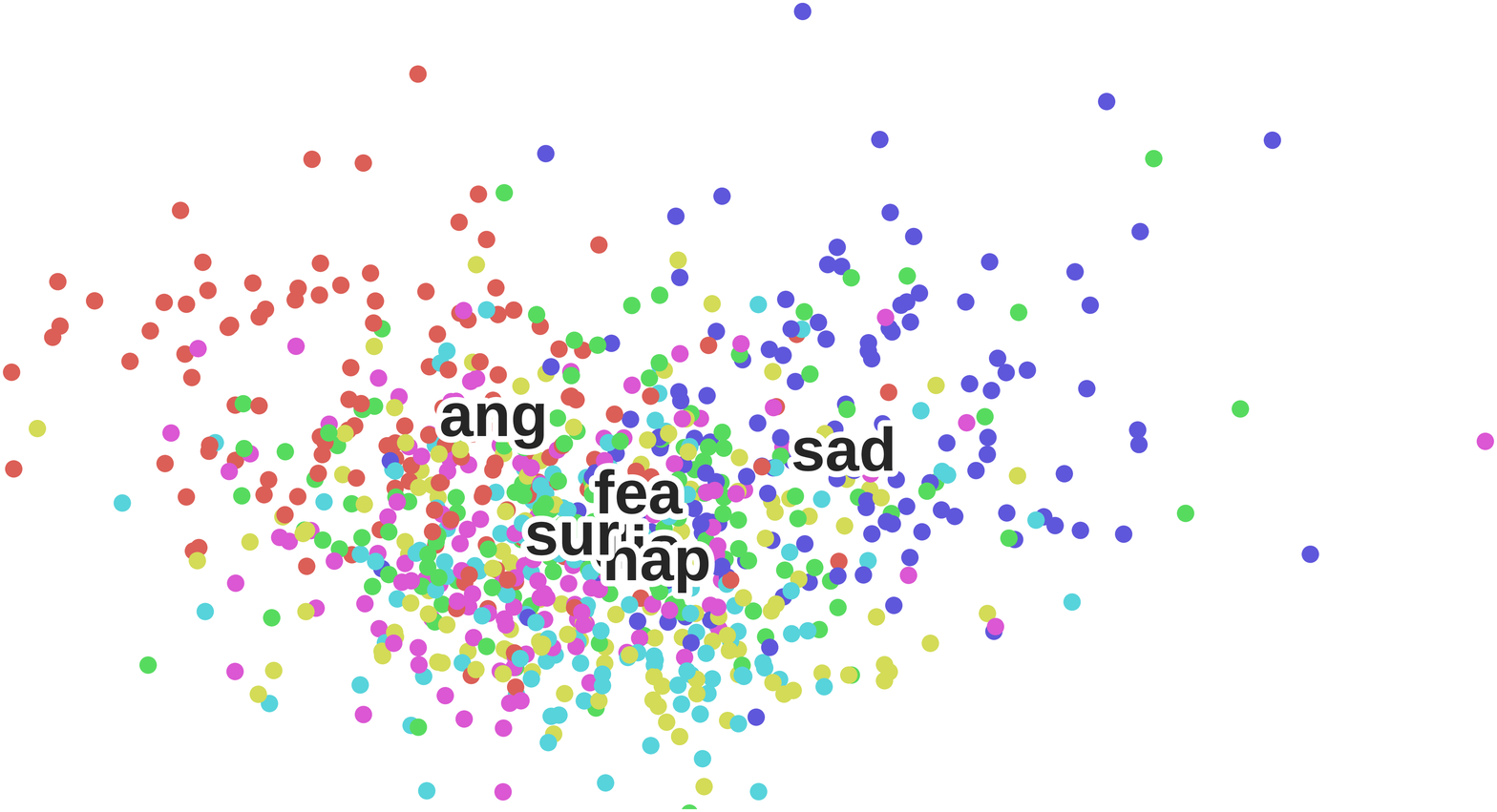}
    \caption{$2.670$}\label{fig:fstcldnn_logmels_act_emo}
  \end{subfigure}%
  \begin{subfigure}[b]{.33\linewidth}
    \centering
    \includegraphics[width=1\textwidth]{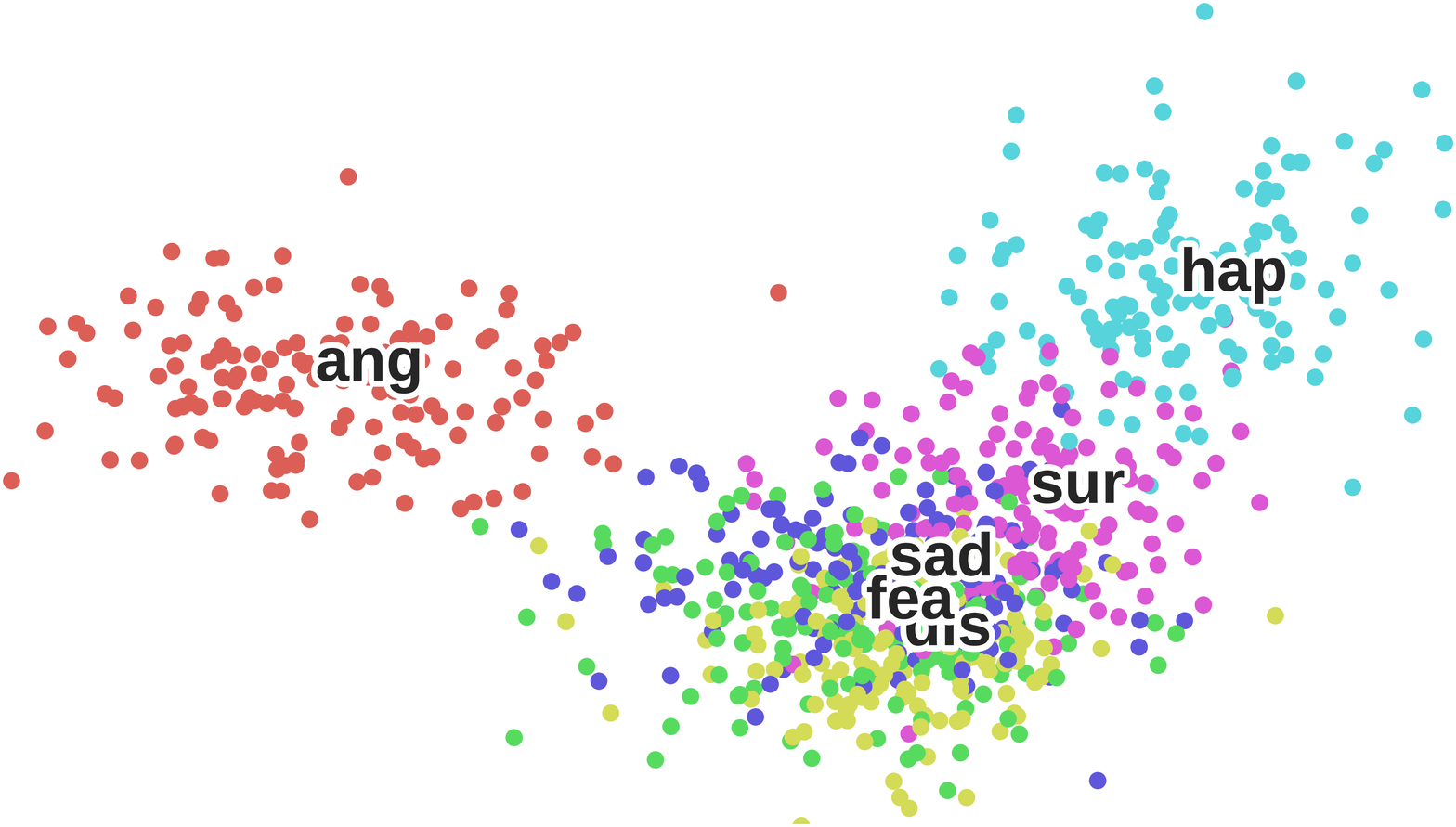}
    \caption{$1.731$}\label{fig:fstcldnn_logmels_tpool_emo}
  \end{subfigure}%
  \begin{subfigure}[b]{.33\linewidth}
    \centering
    \includegraphics[width=1\textwidth]{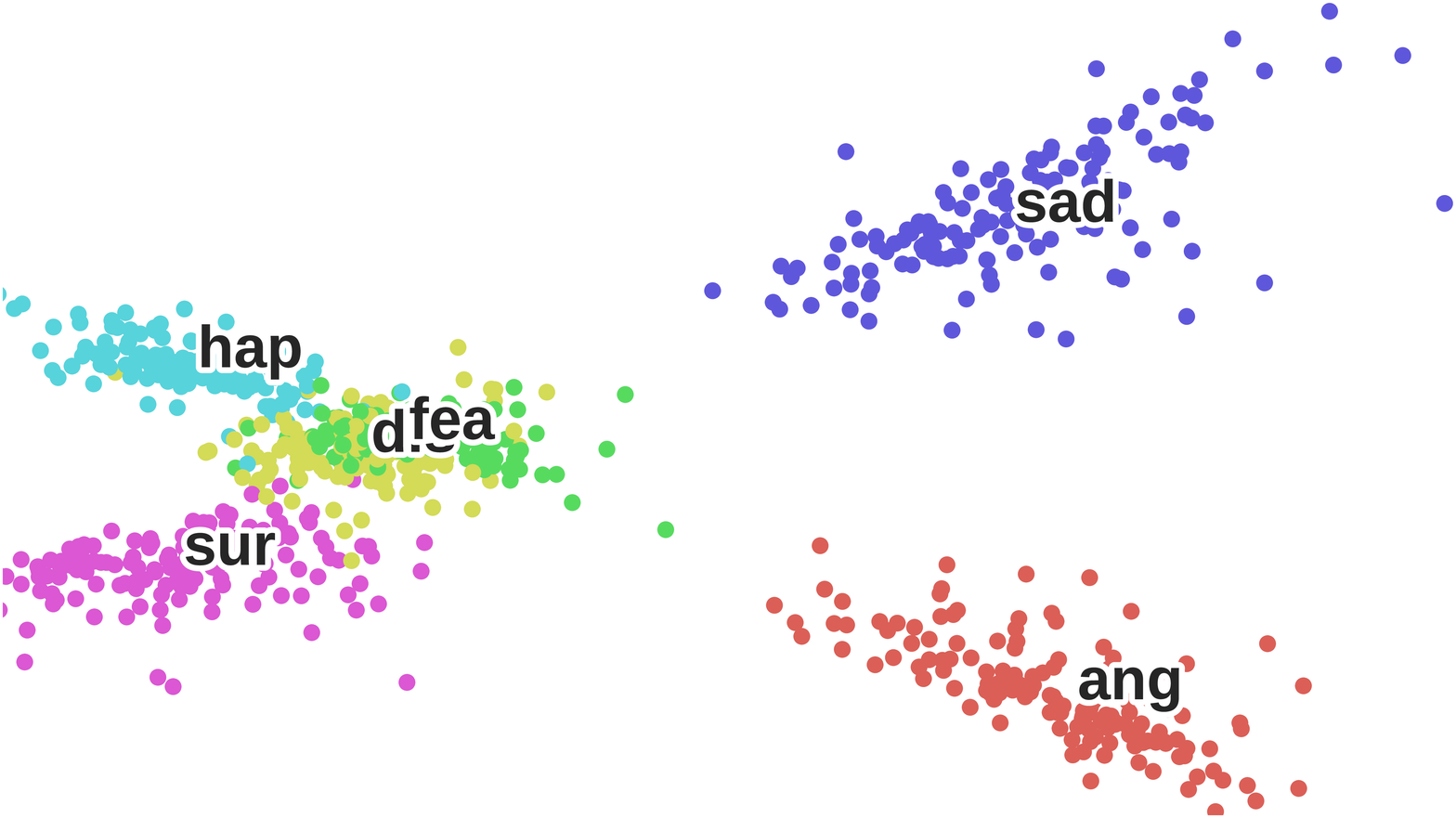}
    \caption{$0.361$}\label{fig:fstcldnn_logmels_lastfc_emo}
  \end{subfigure}\\%
  
  \begin{subfigure}[b]{0.33\linewidth}
    \centering
    \includegraphics[width=1\textwidth]{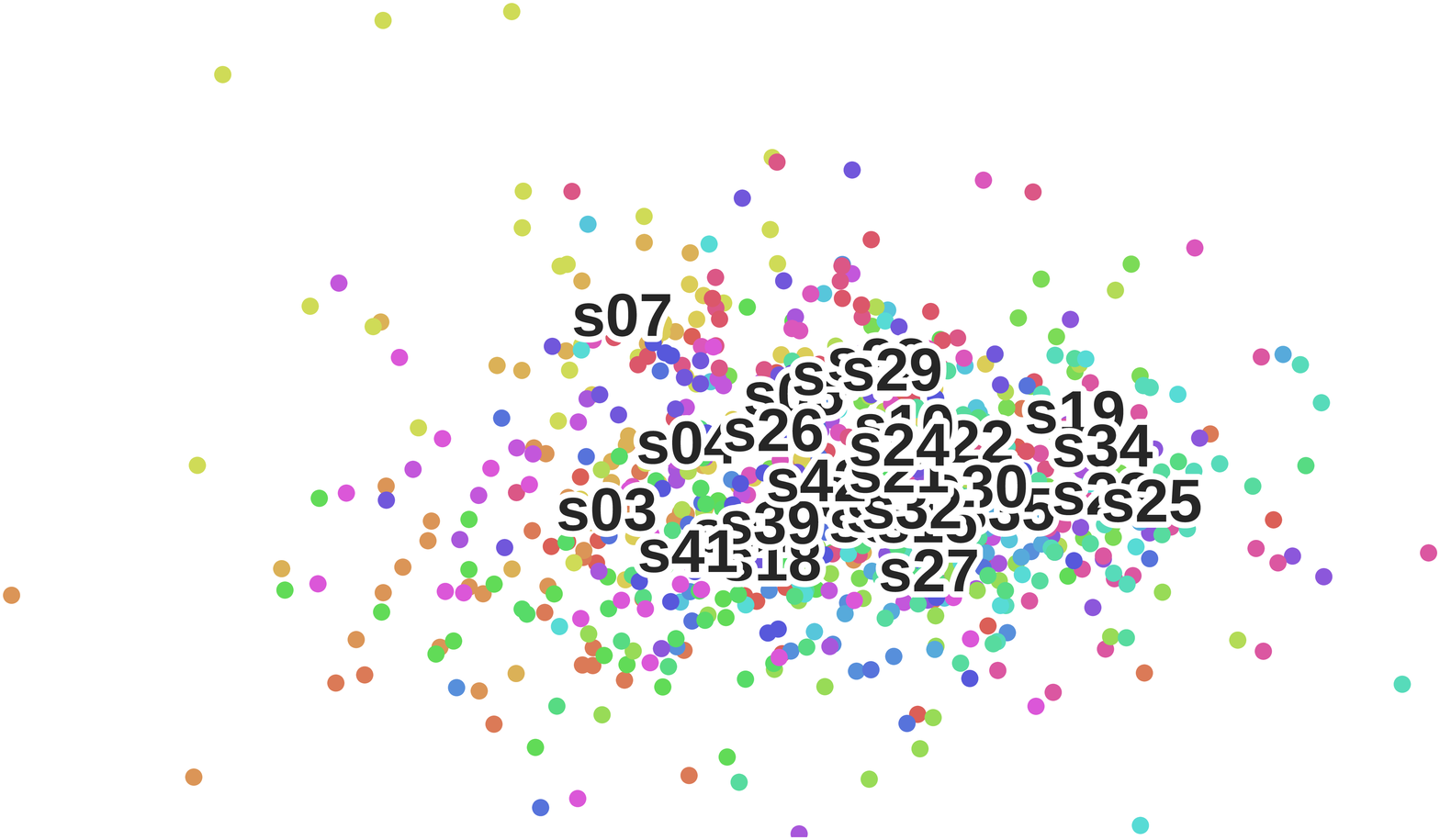}
    \caption{$1.391$}\label{fig:fstcldnn_logmels_act_spkr}
  \end{subfigure}%
  \begin{subfigure}[b]{.33\linewidth}
    \centering
    \includegraphics[width=1\textwidth]{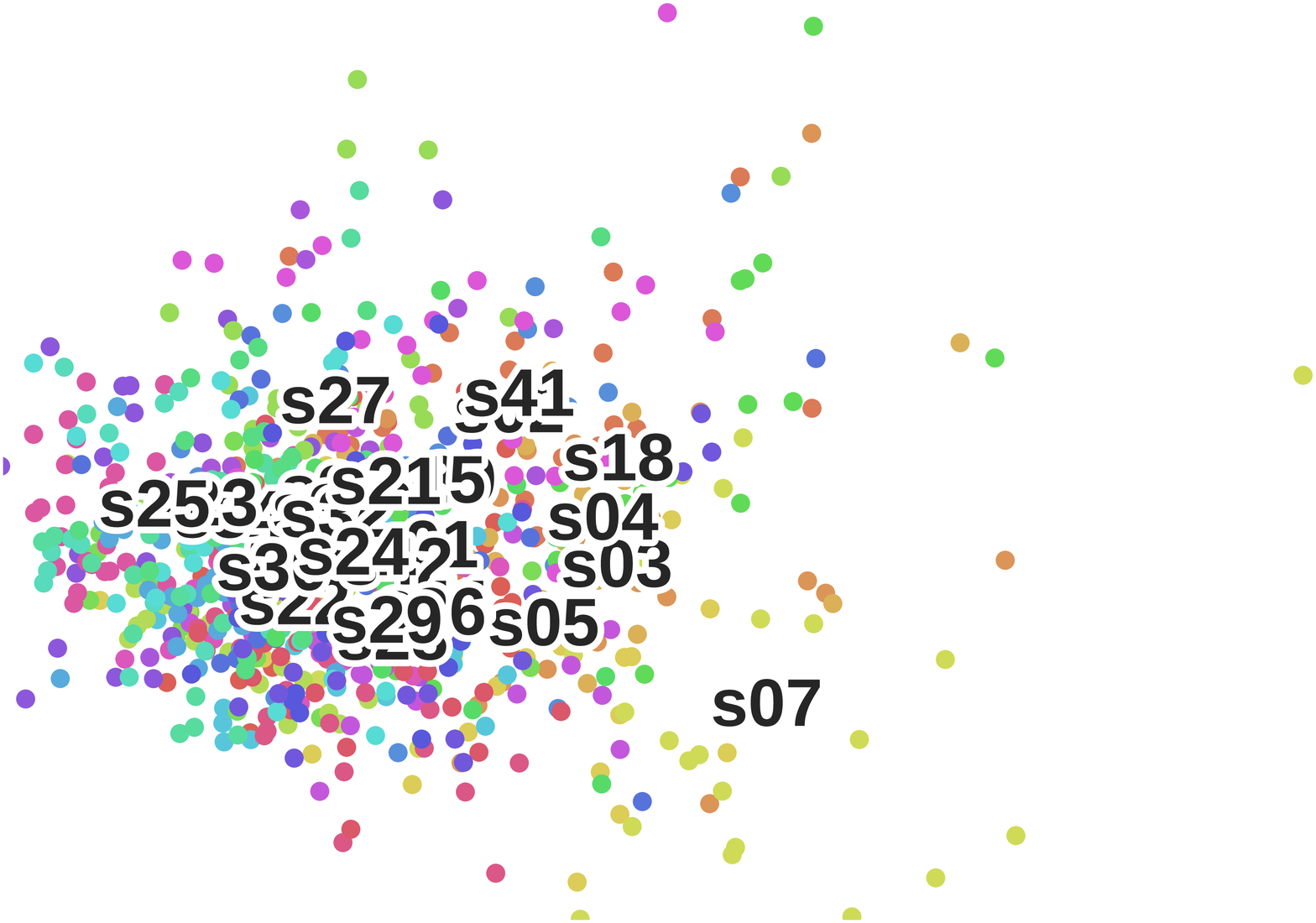}
    \caption{$2.507$}\label{fig:fstcldnn_logmels_tpool_spkr}
  \end{subfigure}%
  \begin{subfigure}[b]{.33\linewidth}
    \centering
    \includegraphics[width=1\textwidth]{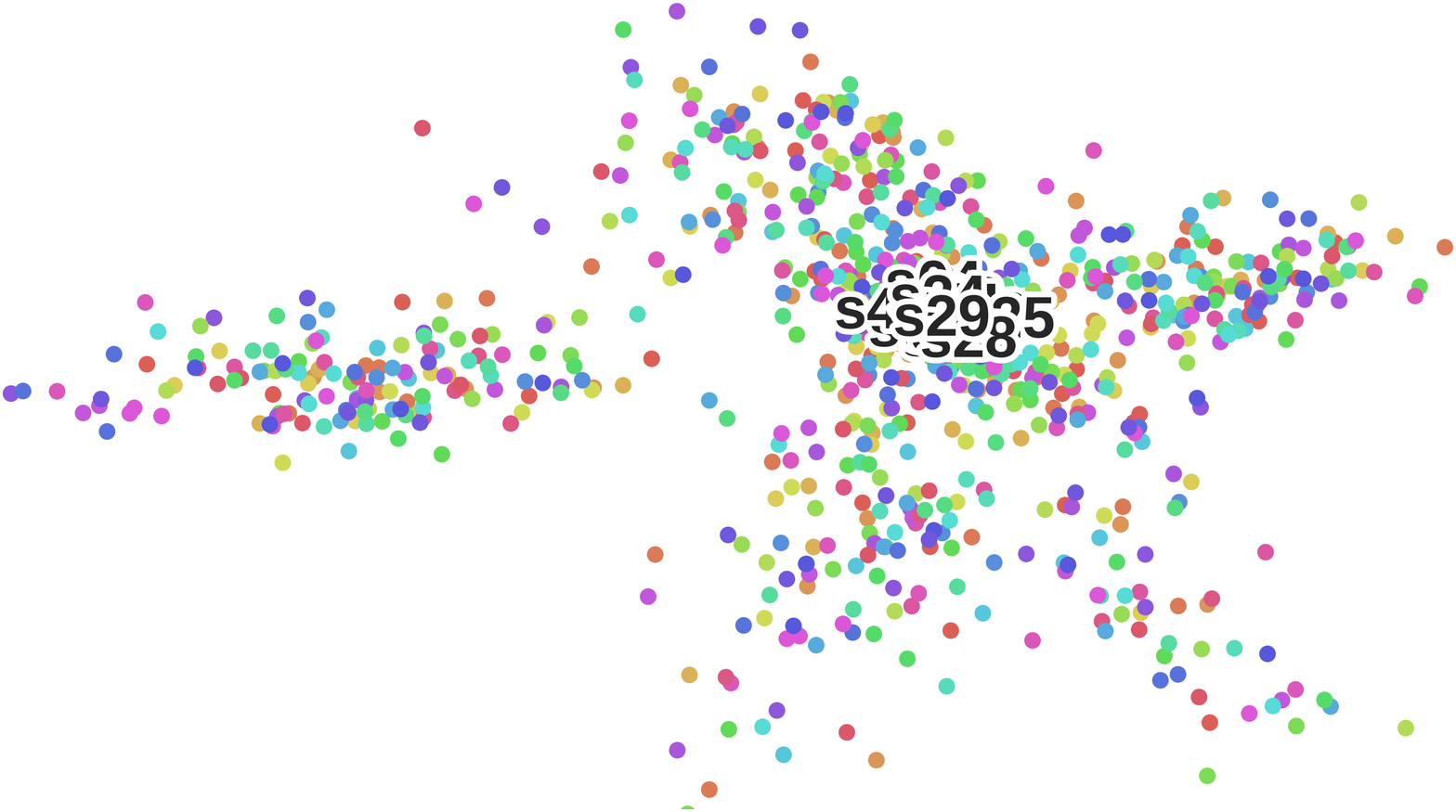}
    \caption{$7.057$}\label{fig:fstcldnn_logmels_lastfc_spkr}
  \end{subfigure}\\%
  
  \begin{subfigure}[b]{0.33\linewidth}
    \centering
    \includegraphics[width=1\textwidth]{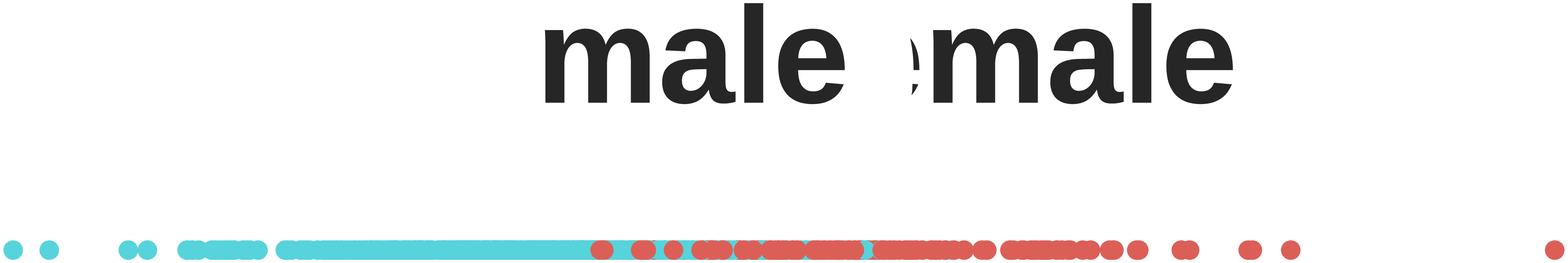}
    \caption{$3.338$}\label{fig:fstcldnn_logmels_act_gender}
  \end{subfigure}%
  \begin{subfigure}[b]{.33\linewidth}
    \centering
    \includegraphics[width=1\textwidth]{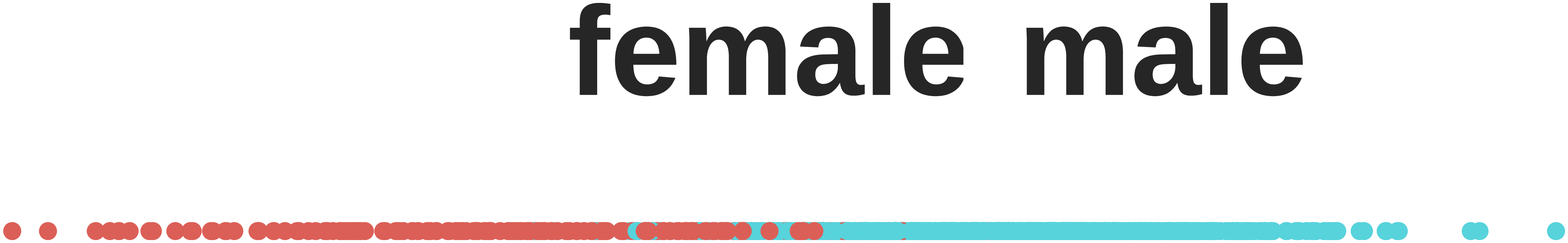}
    \caption{$5.656$}\label{fig:fstcldnn_logmels_tpool_gender}
  \end{subfigure}%
  \begin{subfigure}[b]{.33\linewidth}
    \centering
    \includegraphics[width=1\textwidth]{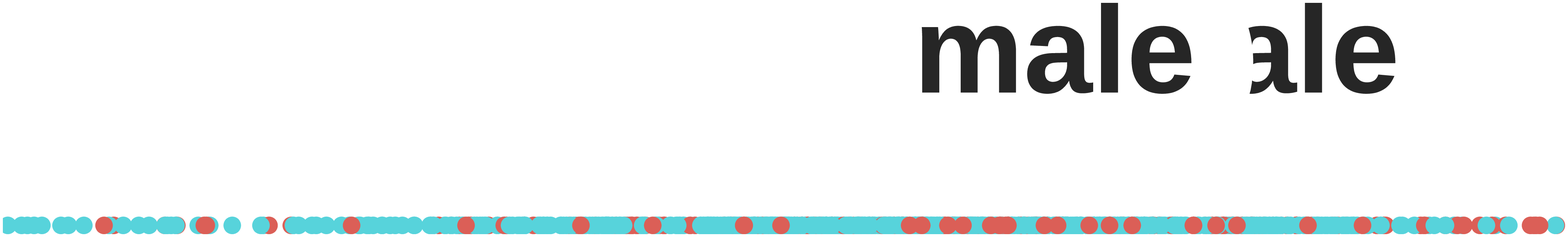}
    \caption{$16.753$}\label{fig:fstcldnn_logmels_lastfc_gender}
  \end{subfigure}\\%
  \caption{The visualization for the modules in the FST-CLDNN (log-Mels). The first, second and third rows correspond to the affective, speaker and gender information, while the first, second and third columns denote the output of the CNN, the BLSTM and the MLP modules, respectively. In each subplot, every dot indicates an utterance, where utterances within the same class are painted with the same color and their centers of classes are marked with according labels such as \textit{hap}, \textit{s07} and \textit{female}. The title of each subplot is the $\mathbf{\rho}$ value, i.e. the quality measure of a clustering, for the distributions in the subplot.}
  \label{fig:fstcldnn_logmels}
\end{figure*}

\begin{figure*}[hb]
\centering
  \begin{subfigure}[b]{0.33\linewidth}
    \centering
    \includegraphics[width=1\textwidth]{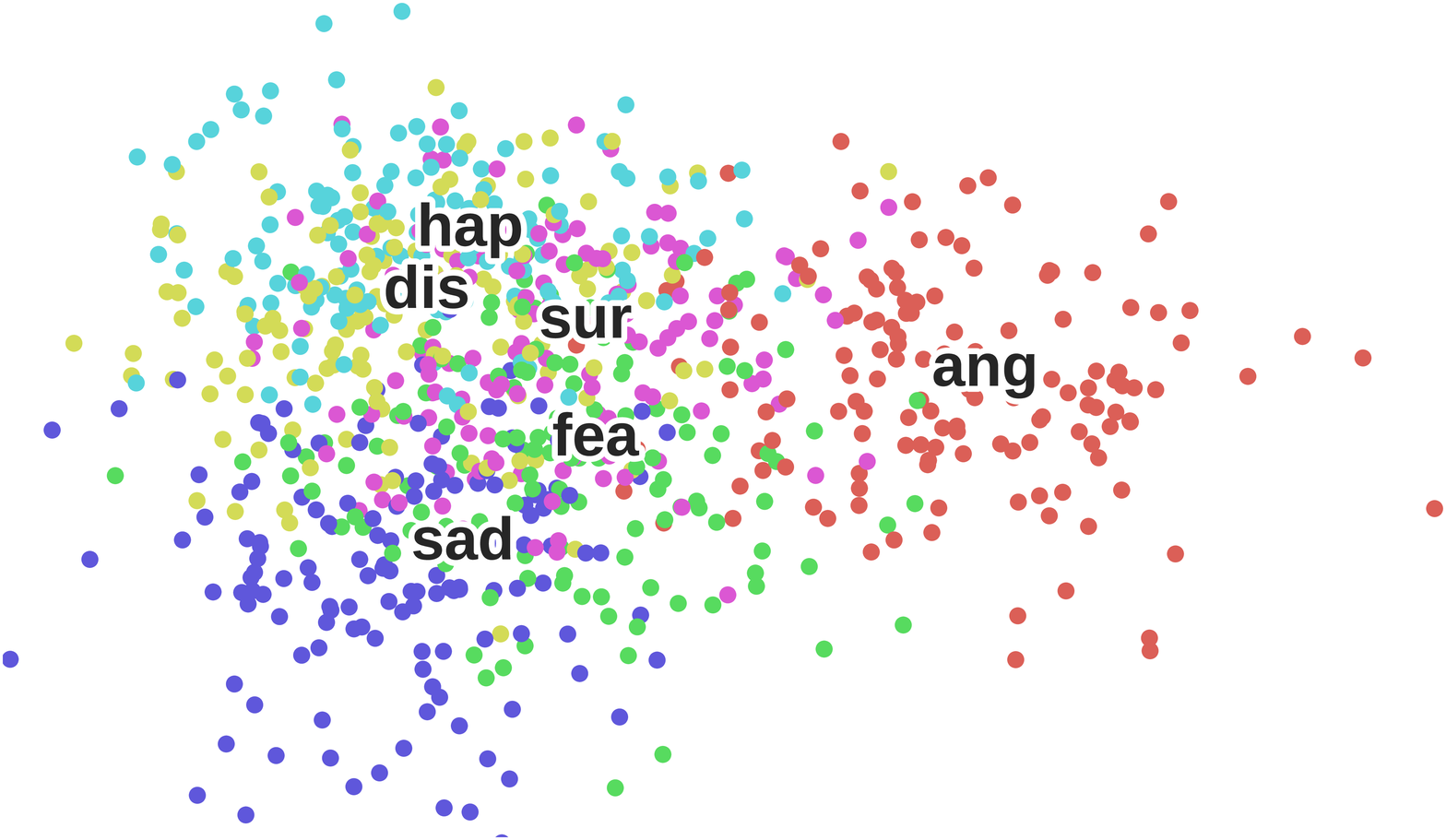}
    \caption{$2.850$}\label{fig:tcldnn_mfccs_act_emo}
  \end{subfigure}%
  \begin{subfigure}[b]{.33\linewidth}
    \centering
    \includegraphics[width=1\textwidth]{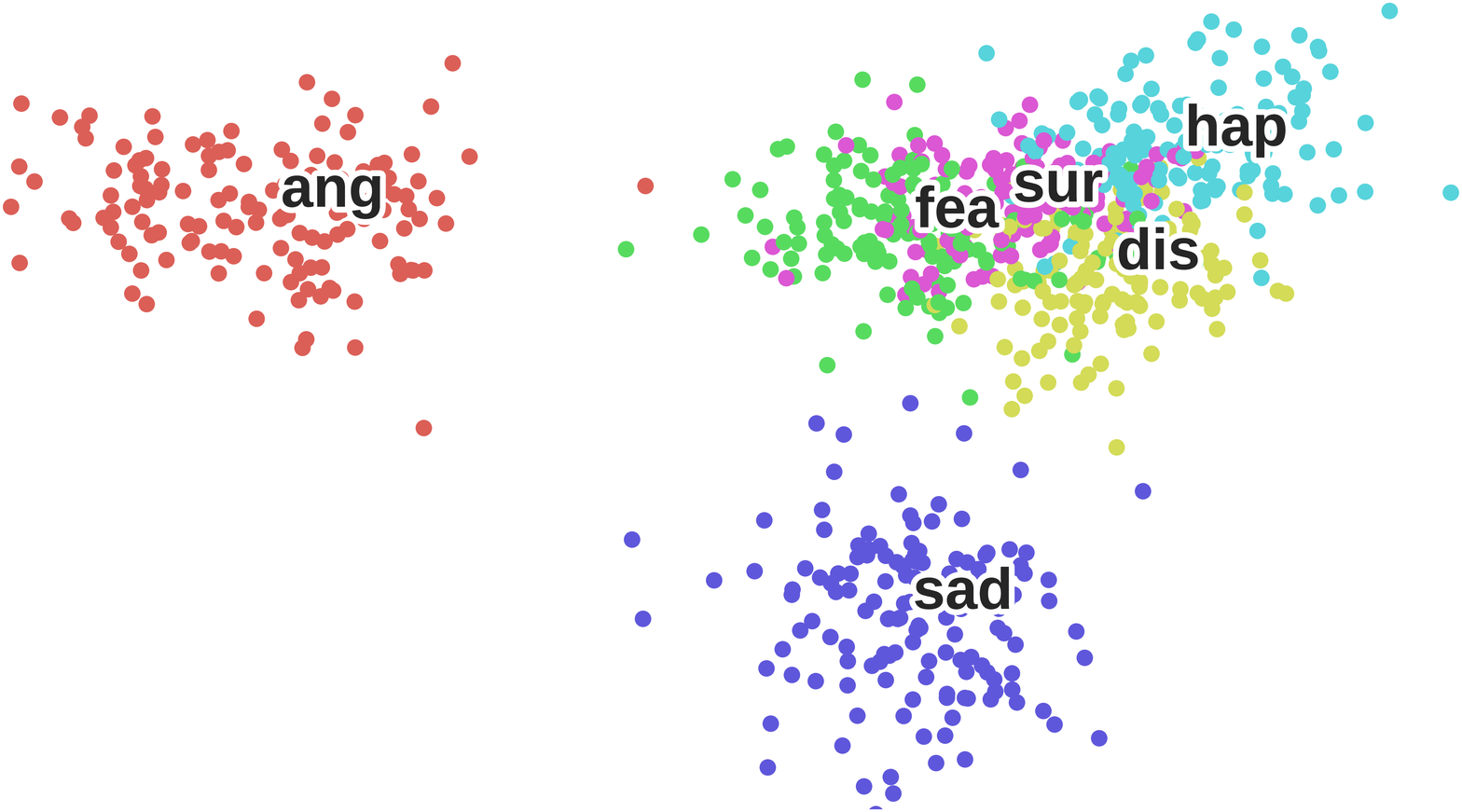}
    \caption{$1.023$}\label{fig:tcldnn_mfccs_tpool_emo}
  \end{subfigure}%
  \begin{subfigure}[b]{.33\linewidth}
    \centering
    \includegraphics[width=1\textwidth]{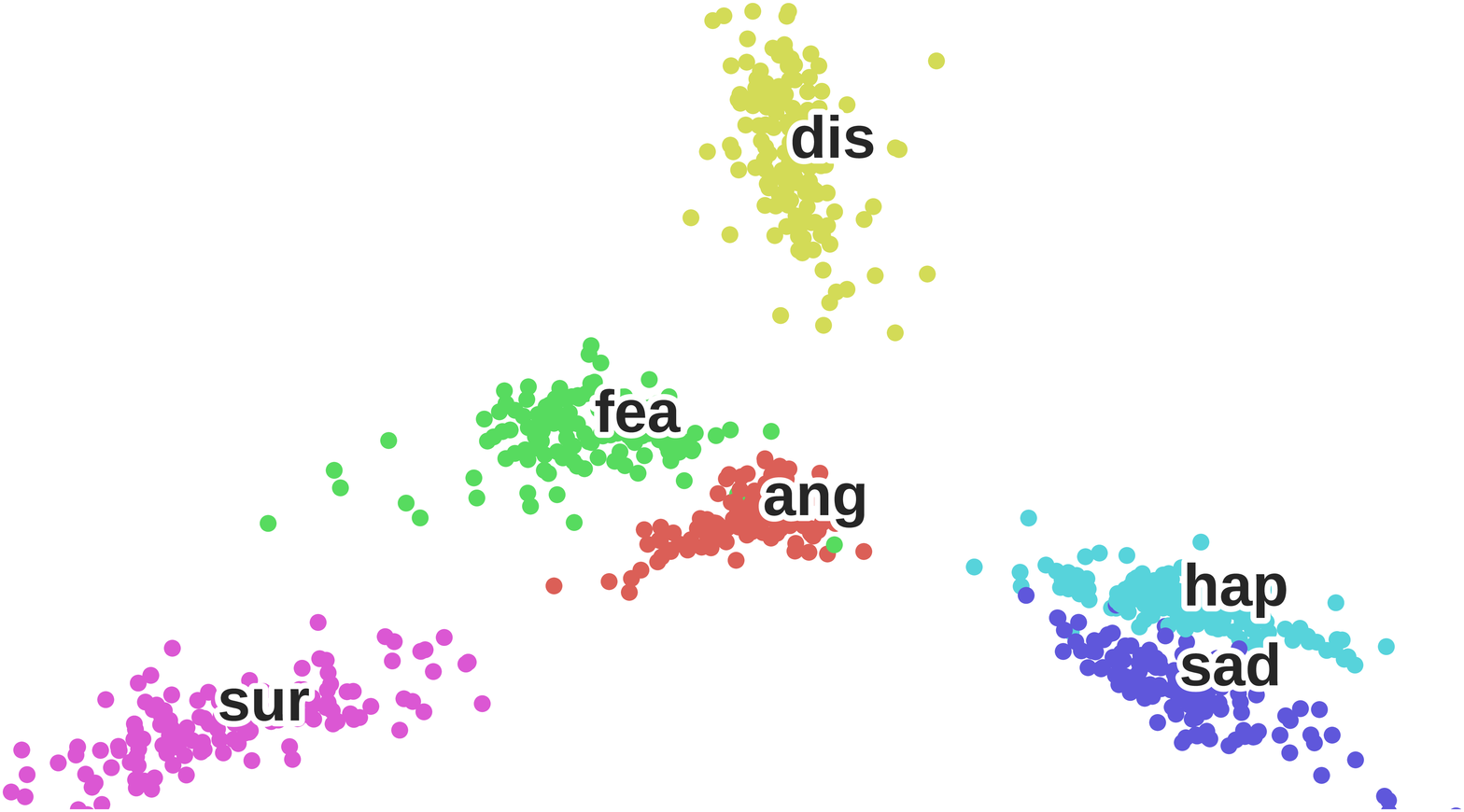}
    \caption{$0.315$}\label{fig:tcldnn_mfccs_lastfc_emo}
  \end{subfigure}\\%
  
  \begin{subfigure}[b]{0.33\linewidth}
    \centering
    \includegraphics[width=1\textwidth]{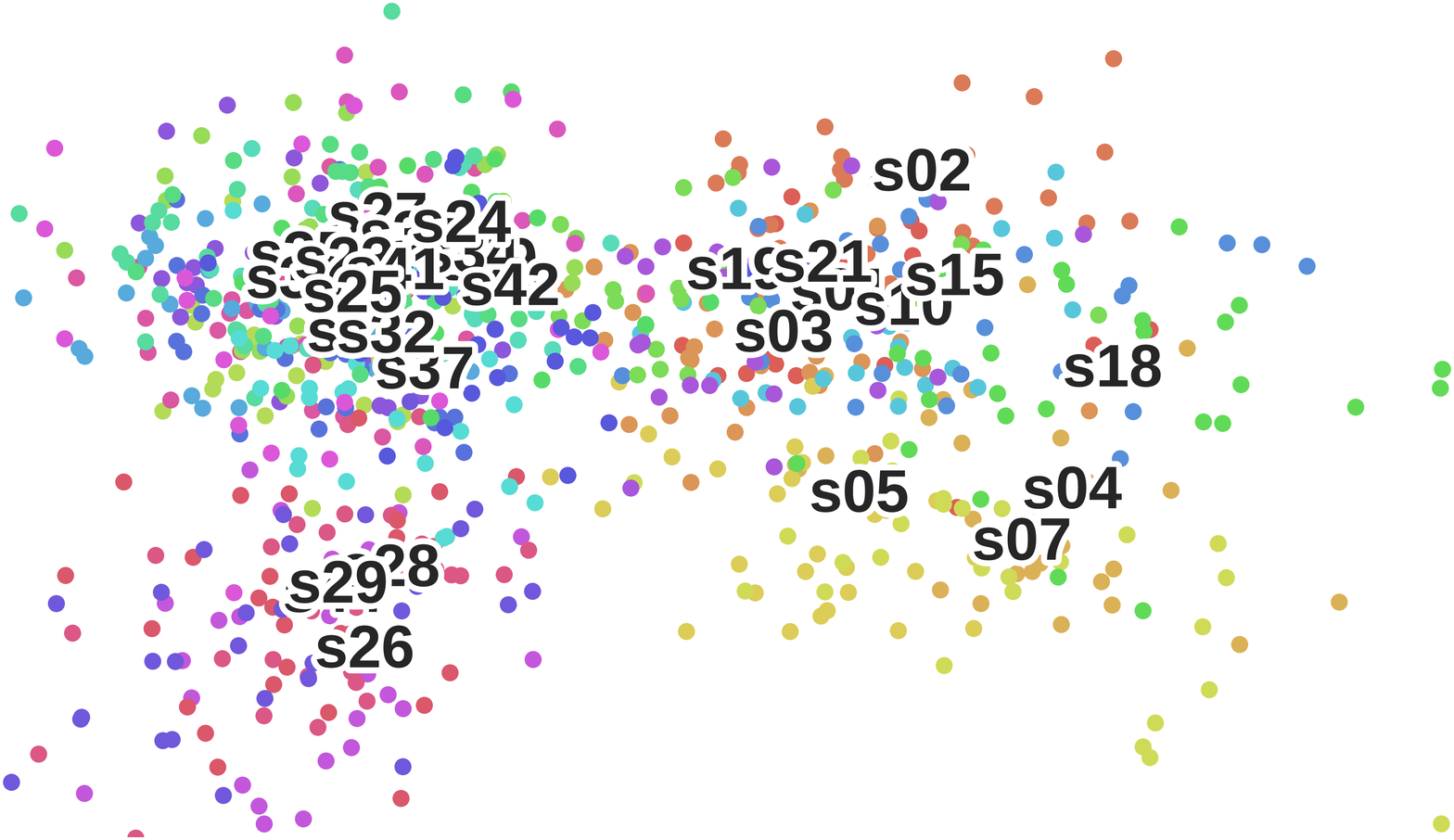}
    \caption{$1.326$}\label{fig:tcldnn_mfccs_act_spkr}
  \end{subfigure}%
  \begin{subfigure}[b]{.33\linewidth}
    \centering
    \includegraphics[width=1\textwidth]{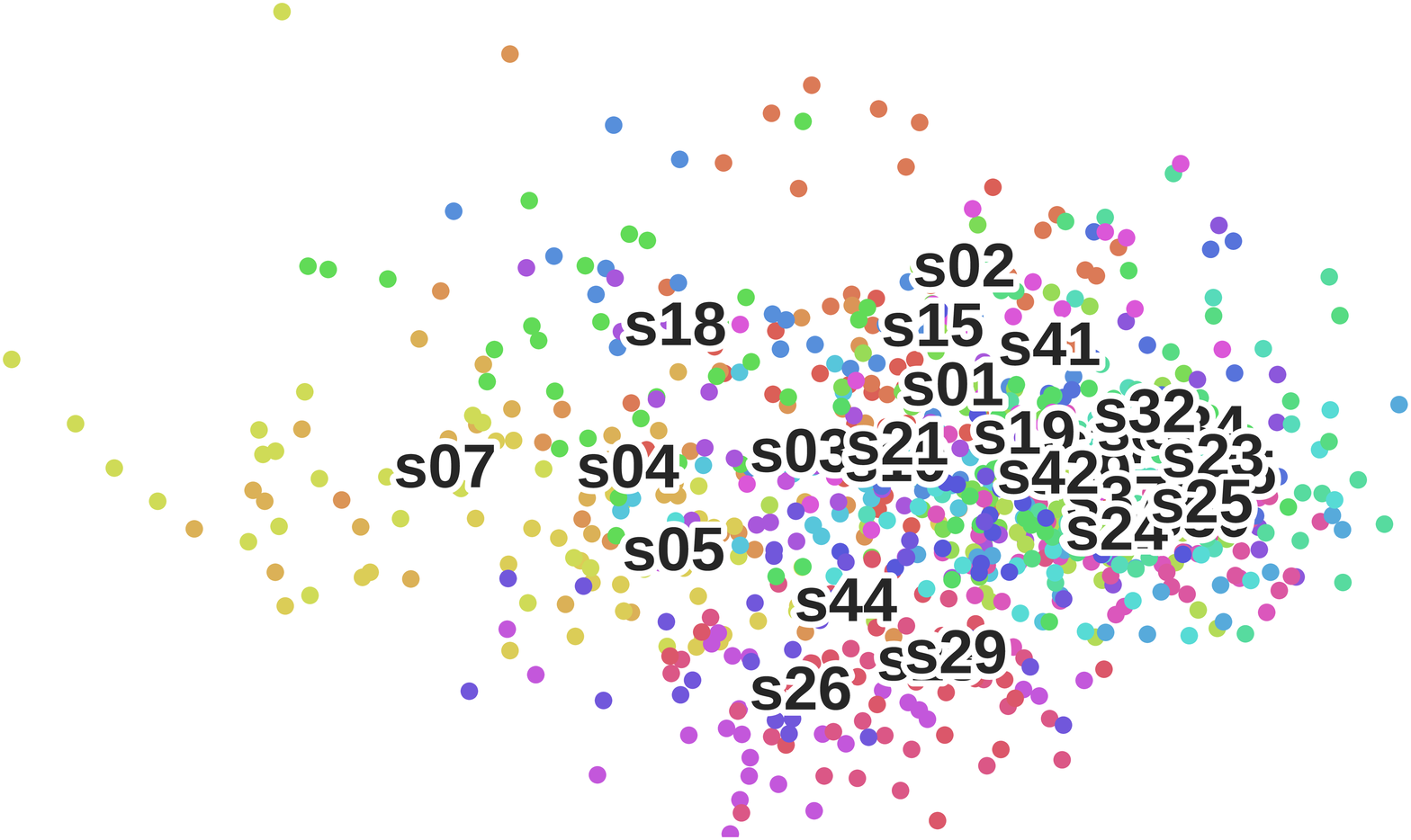}
    \caption{$2.436$}\label{fig:tcldnn_mfccs_tpool_spkr}
  \end{subfigure}%
  \begin{subfigure}[b]{.33\linewidth}
    \centering
    \includegraphics[width=1\textwidth]{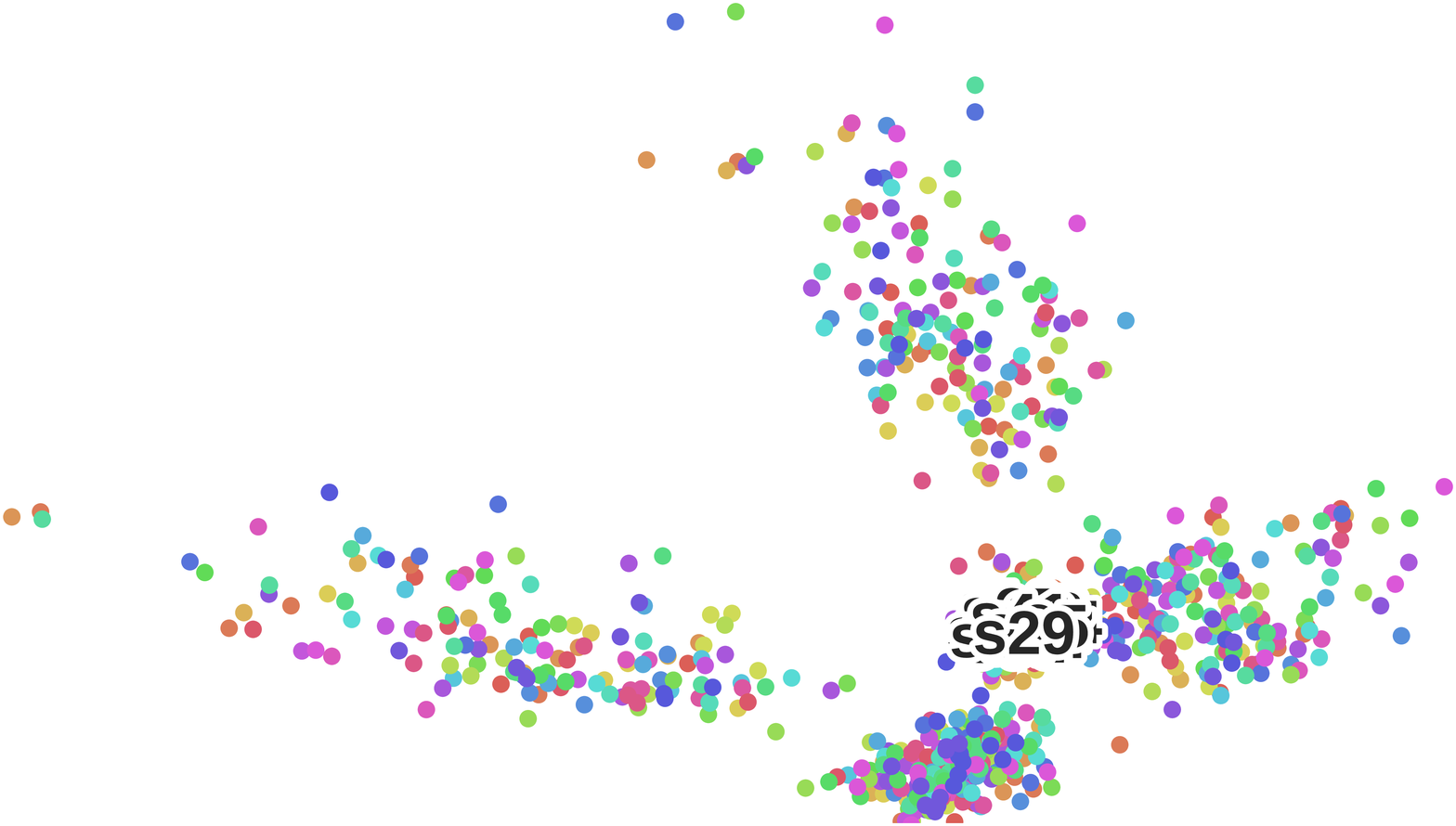}
    \caption{$7.273$}\label{fig:tcldnn_mfccs_lastfc_spkr}
  \end{subfigure}\\%
  
  \begin{subfigure}[b]{0.33\linewidth}
    \centering
    \includegraphics[width=1\textwidth]{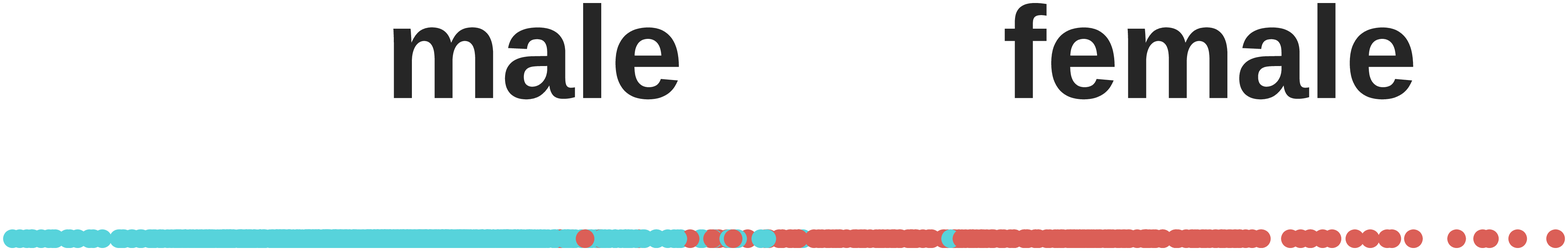}
    \caption{$2.441$}\label{fig:tcldnn_mfccs_act_gender}
  \end{subfigure}%
  \begin{subfigure}[b]{.33\linewidth}
    \centering
    \includegraphics[width=1\textwidth]{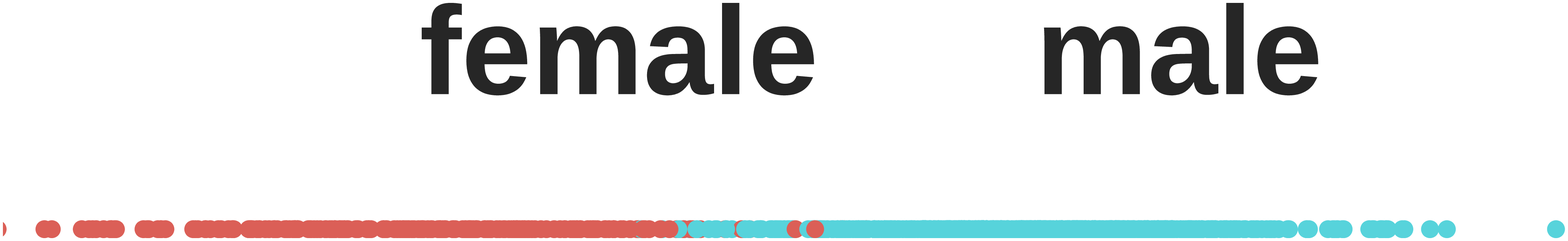}
    \caption{$4.318$}\label{fig:tcldnn_mfccs_tpool_gender}
  \end{subfigure}%
  \begin{subfigure}[b]{.33\linewidth}
    \centering
    \includegraphics[width=1\textwidth]{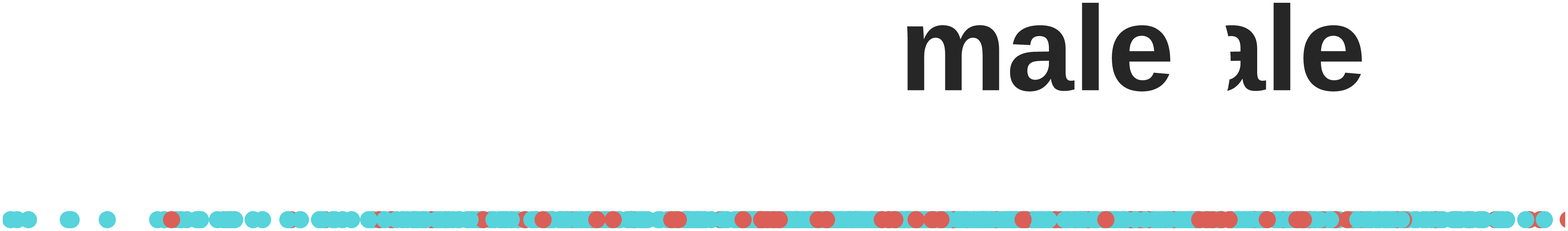}
    \caption{$13.363$}\label{fig:tcldnn_mfccs_lastfc_gender}
  \end{subfigure}\\%
  \caption{The visualization for the modules in the T-CLDNN (MFCCs). The first, second and third rows correspond to the affective, speaker and gender information, while the first, second and third columns denote the output of the CNN, the BLSTM and the MLP modules, respectively. In each subplot, every dot indicates an utterance, where utterances within the same class are painted with the same color and their centers of classes are marked with according labels such as \textit{hap}, \textit{s07} and \textit{female}. The title of each subplot is the $\mathbf{\rho}$ value, i.e. the quality measure of a clustering, for the distributions in the subplot.}
  \label{fig:tcldnn_mfccs}
\end{figure*}

\begin{figure*}[hb]
\centering
  \begin{subfigure}[b]{0.33\linewidth}
    \centering
    \includegraphics[width=1\textwidth]{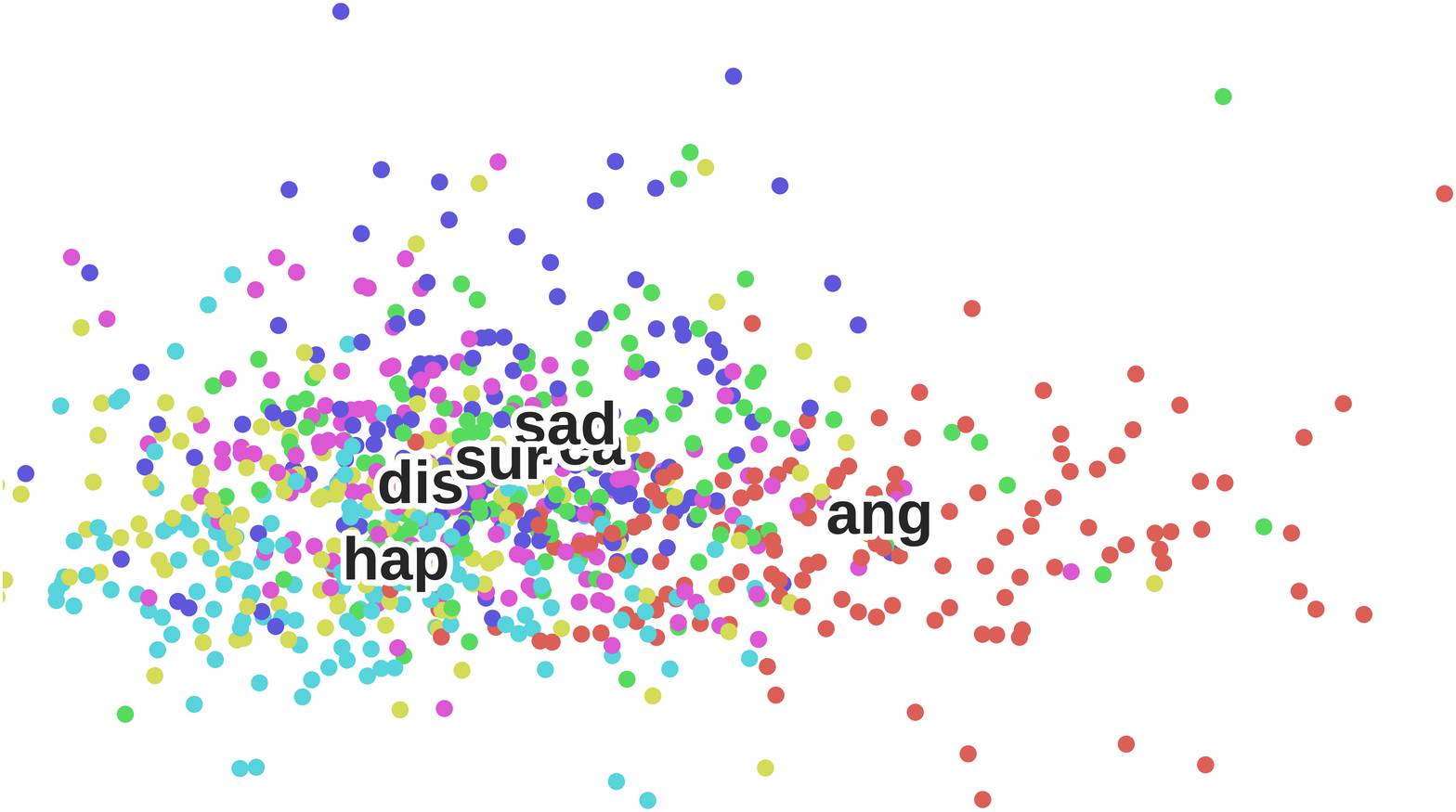}
    \caption{$2.495$}\label{fig:fstcldnn_mfccs_act_emo}
  \end{subfigure}%
  \begin{subfigure}[b]{.33\linewidth}
    \centering
    \includegraphics[width=1\textwidth]{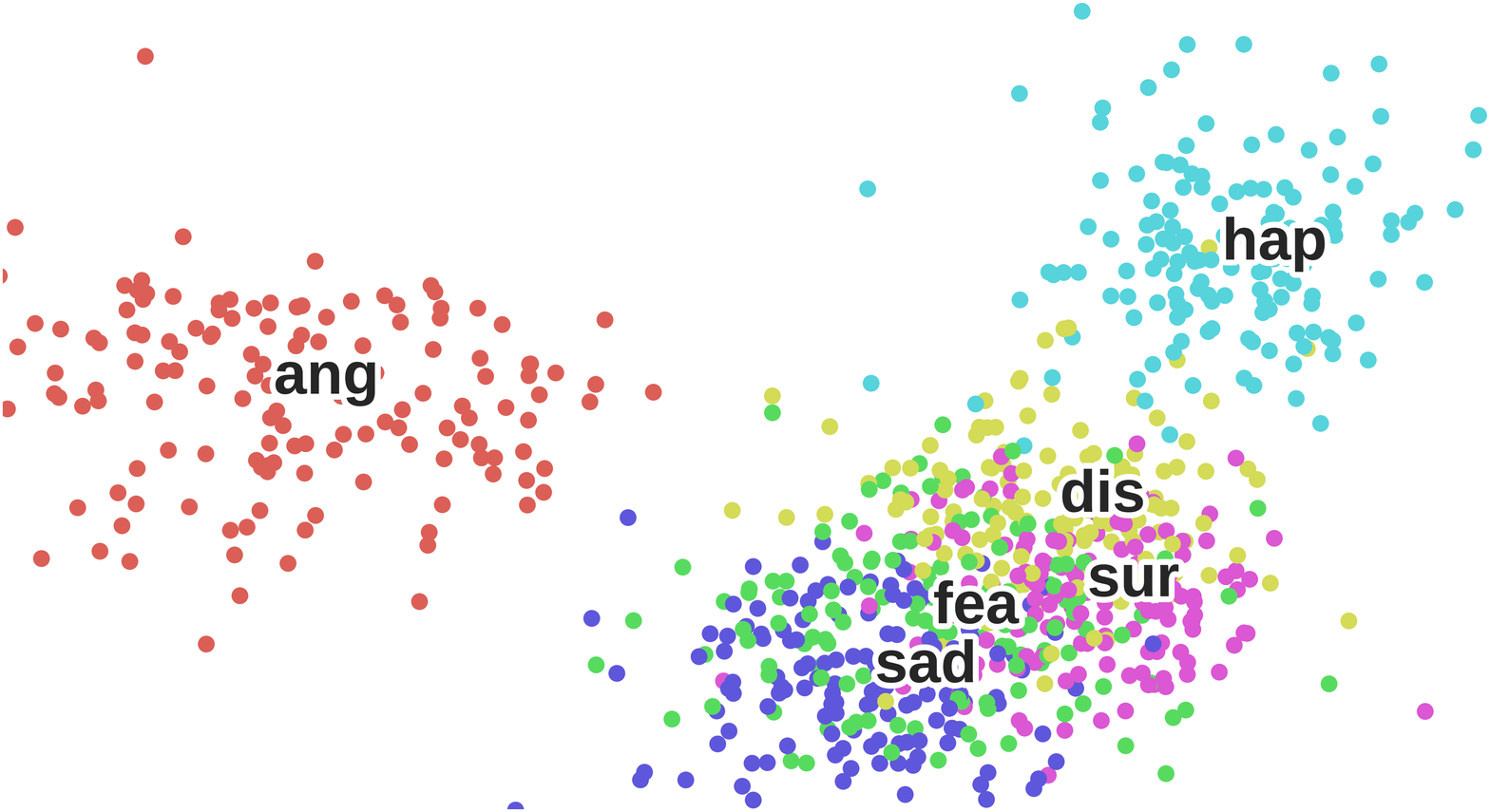}
    \caption{$1.593$}\label{fig:fstcldnn_mfccs_tpool_emo}
  \end{subfigure}%
  \begin{subfigure}[b]{.33\linewidth}
    \centering
    \includegraphics[width=1\textwidth]{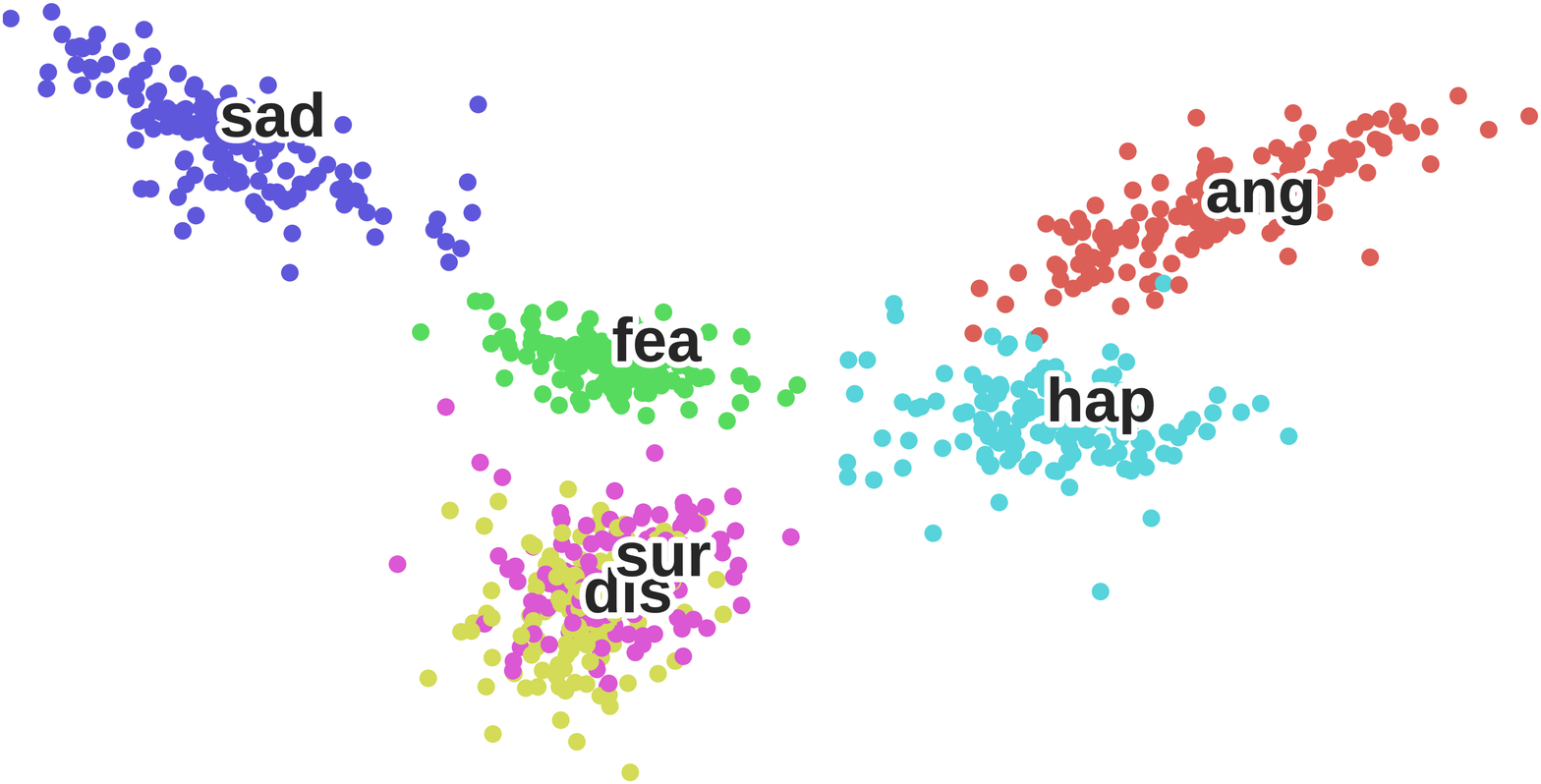}
    \caption{$0.301$}\label{fig:fstcldnn_mfccs_lastfc_emo}
  \end{subfigure}\\%
  
  \begin{subfigure}[b]{0.33\linewidth}
    \centering
    \includegraphics[width=1\textwidth]{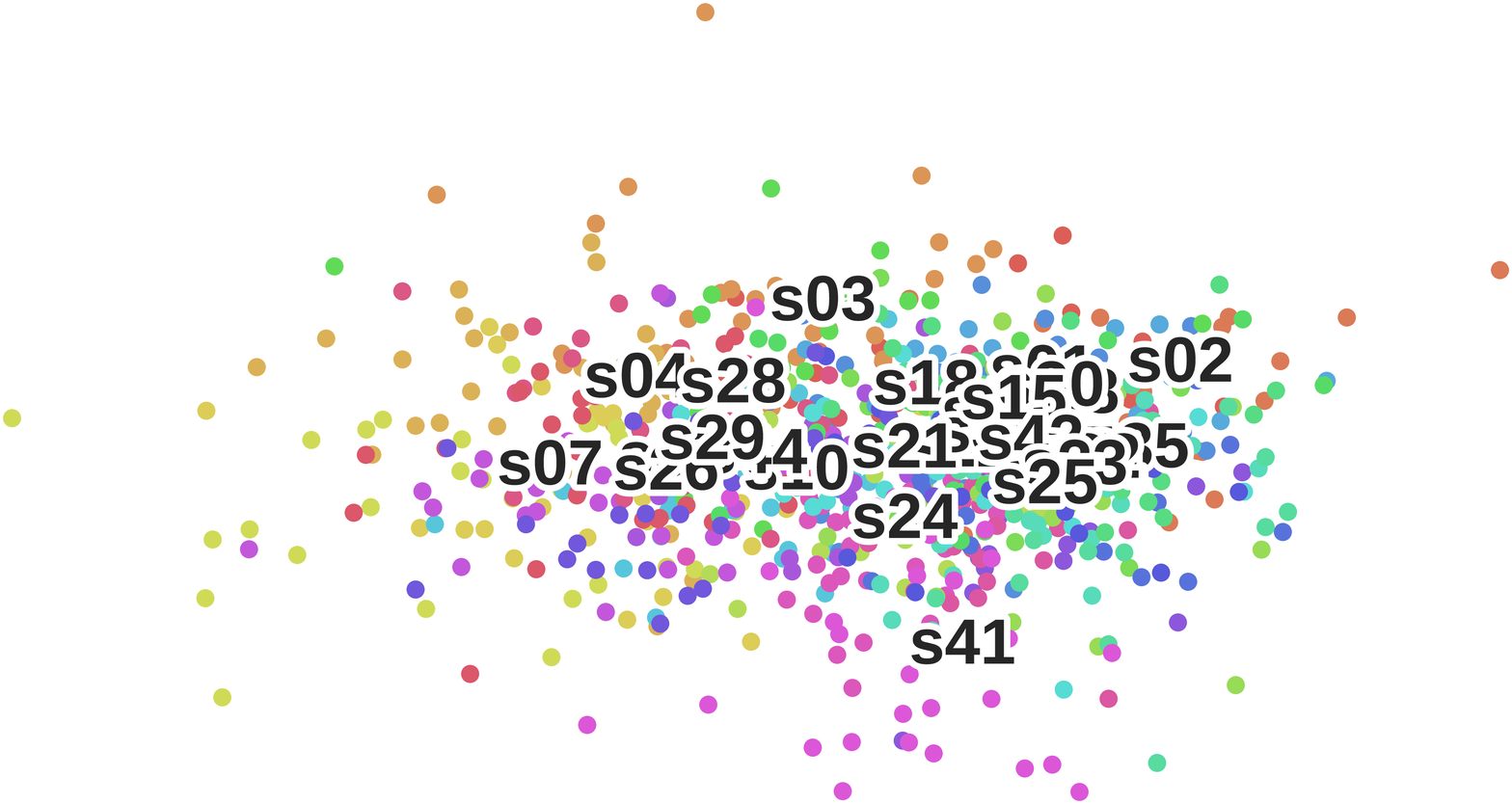}
    \caption{$1.486$}\label{fig:fstcldnn_mfccs_act_spkr}
  \end{subfigure}%
  \begin{subfigure}[b]{.33\linewidth}
    \centering
    \includegraphics[width=1\textwidth]{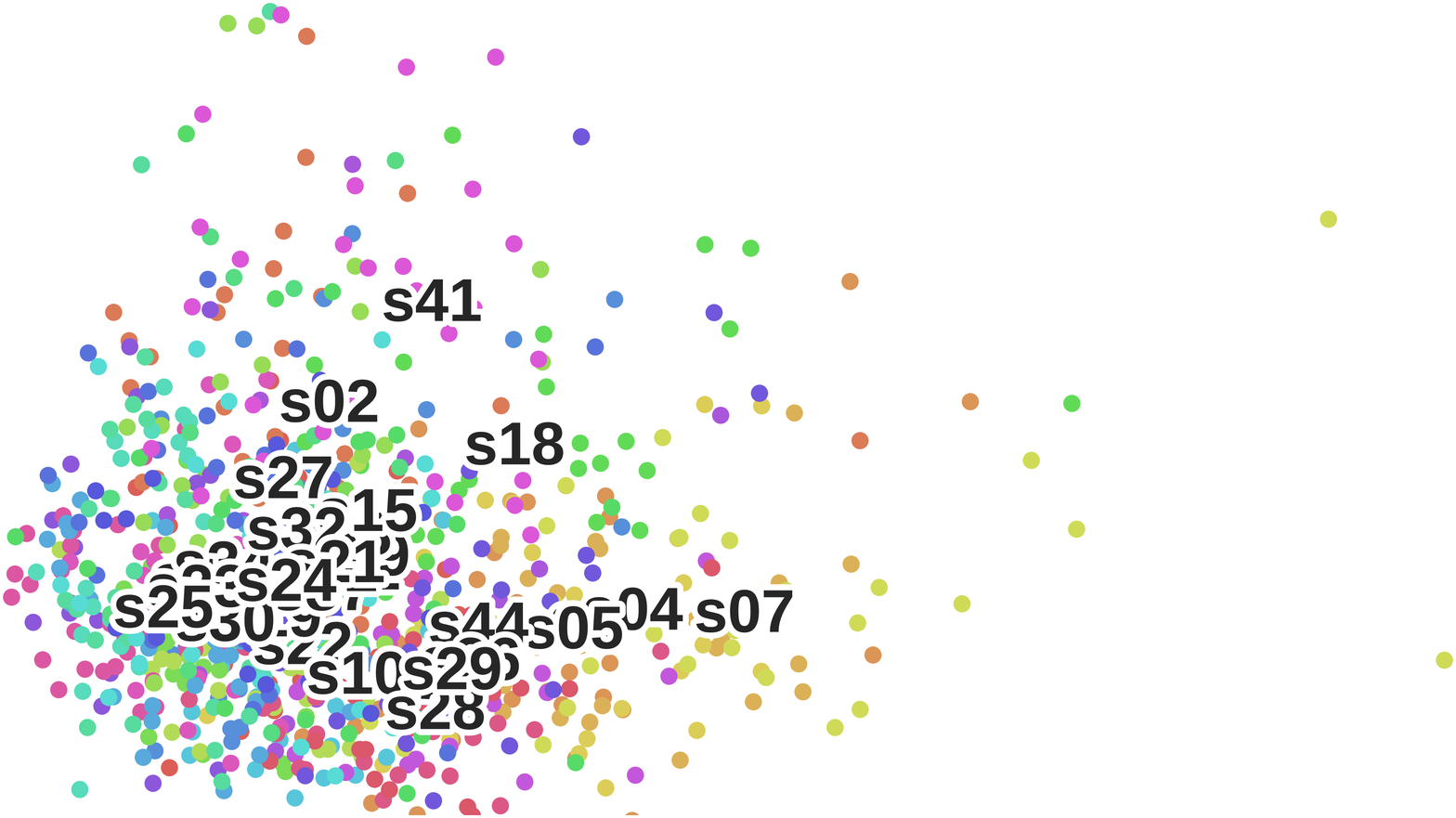}
    \caption{$2.563$}\label{fig:fstcldnn_mfccs_tpool_spkr}
  \end{subfigure}%
  \begin{subfigure}[b]{.33\linewidth}
    \centering
    \includegraphics[width=1\textwidth]{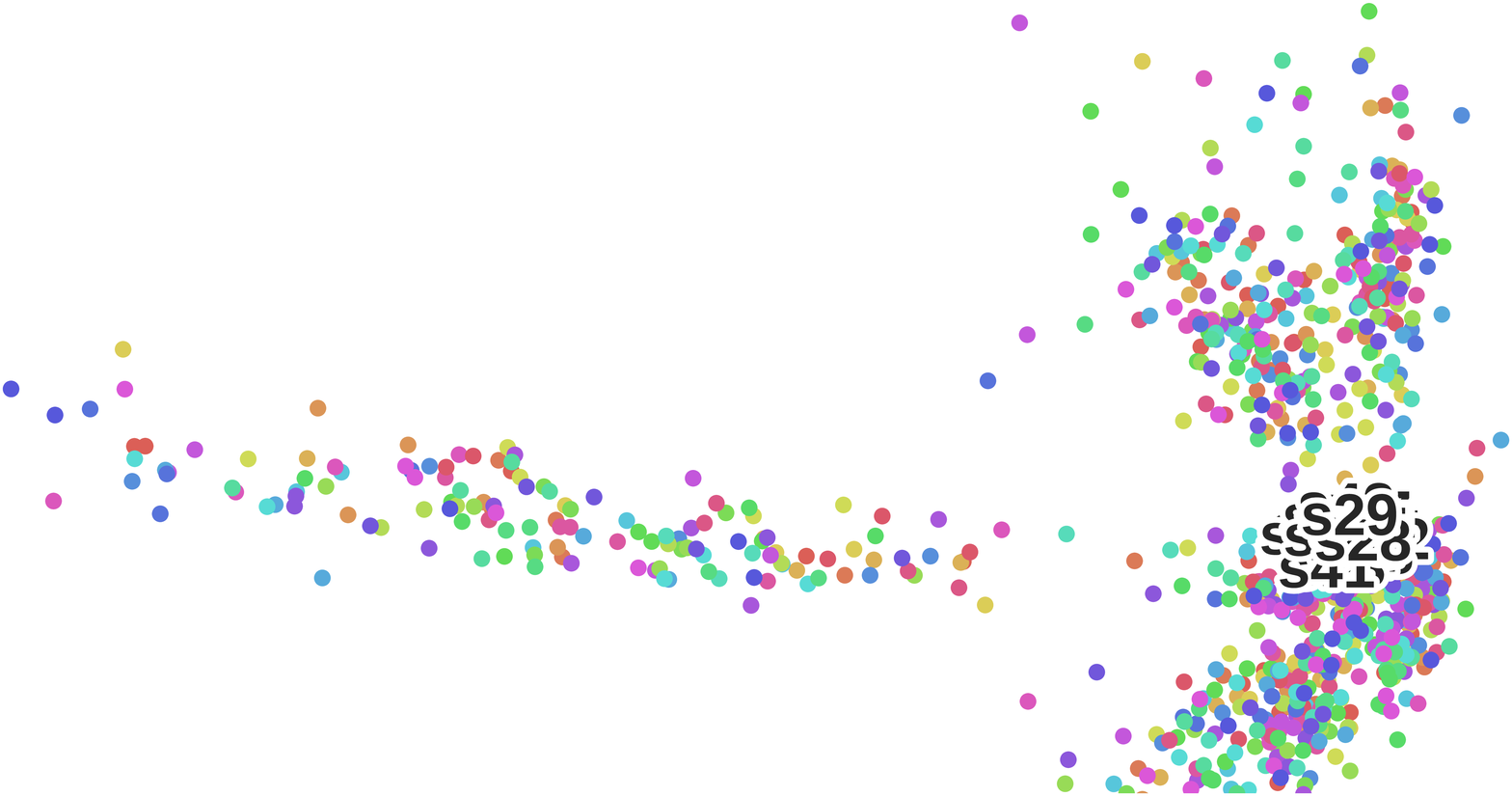}
    \caption{$6.749$}\label{fig:fstcldnn_mfccs_lastfc_spkr}
  \end{subfigure}\\%
  
  \begin{subfigure}[b]{0.33\linewidth}
    \centering
    \includegraphics[width=1\textwidth]{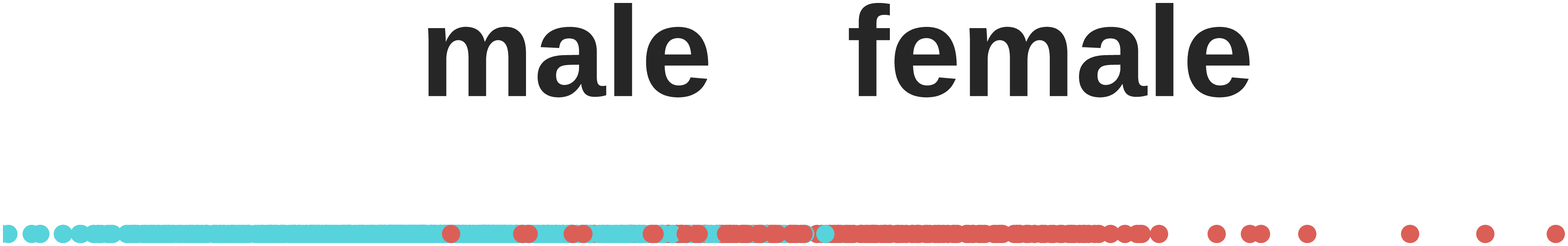}
    \caption{$2.502$}\label{fig:fstcldnn_mfccs_act_gender}
  \end{subfigure}%
  \begin{subfigure}[b]{.33\linewidth}
    \centering
    \includegraphics[width=1\textwidth]{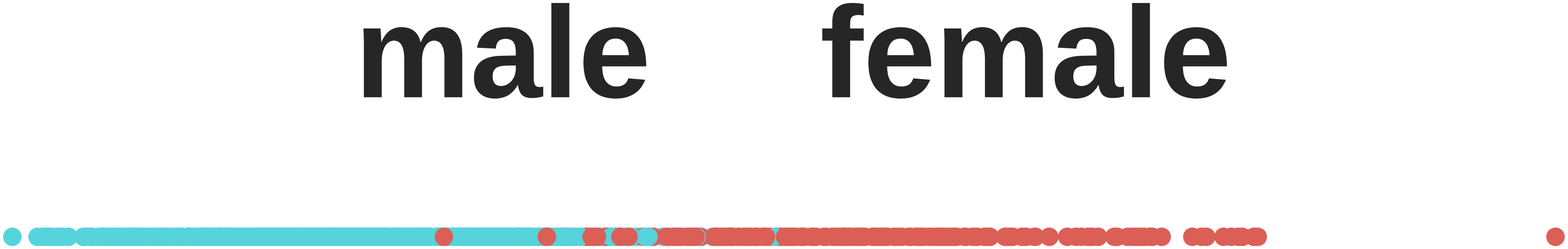}
    \caption{$4.185$}\label{fig:fstcldnn_mfccs_tpool_gender}
  \end{subfigure}%
  \begin{subfigure}[b]{.33\linewidth}
    \centering
    \includegraphics[width=1\textwidth]{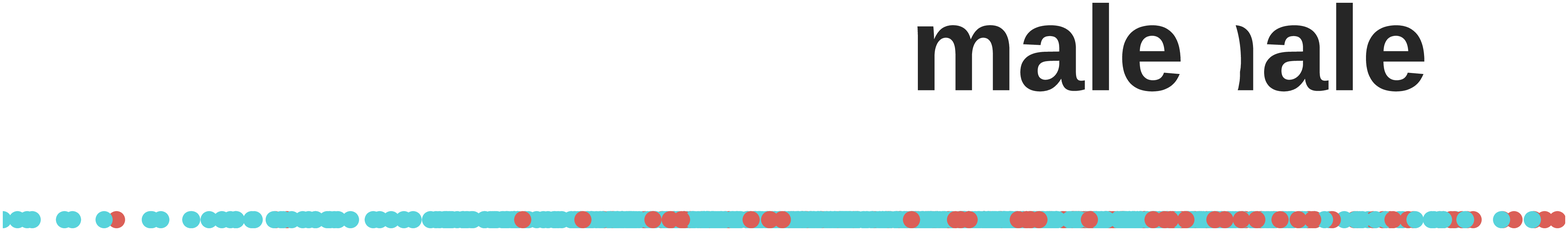}
    \caption{$12.230$}\label{fig:fstcldnn_mfccs_lastfc_gender}
  \end{subfigure}\\%
  \caption{The visualization for the modules in the FST-CLDNN (MFCCs). The first, second and third rows correspond to the affective, speaker and gender information, while the first, second and third columns denote the output of the CNN, the BLSTM and the MLP modules, respectively. In each subplot, every dot indicates an utterance, where utterances within the same class are painted with the same color and their centers of classes are marked with according labels such as \textit{hap}, \textit{s07} and \textit{female}. The title of each subplot is the $\mathbf{\rho}$ value, i.e. the quality measure of a clustering, for the distributions in the subplot.}
  \label{fig:fstcldnn_mfccs}
\end{figure*}

\ifCLASSOPTIONcaptionsoff
  \newpage
\fi

\end{document}